\documentclass{article}
\usepackage{PRIMEarxiv}
\usepackage[T1]{fontenc}
\usepackage{amsmath,amssymb,booktabs,array,graphicx,microtype}
\usepackage{xcolor}
\usepackage{enumitem}
\usepackage{float}
\usepackage{placeins}

\usepackage[round]{natbib}

\usepackage[hidelinks]{hyperref}
\newcommand{\tabrelibench}{\textsc{TabReliBench}}
\newcommand{\codeurl}{https://github.com/Tianqi-Zhao/TabReliBench}
\definecolor{datasetteal}{HTML}{16758C}
\definecolor{modelgold}{HTML}{D9AE53}
\definecolor{interactionorange}{HTML}{C36B46}
\definecolor{splitgreen}{HTML}{76A8A2}
\definecolor{residualgray}{HTML}{C4CBD1}
\definecolor{baselinegray}{HTML}{7A7E83}
\definecolor{repeatteal}{HTML}{8DC2BF}
\definecolor{medianochre}{HTML}{D9AE53}
\definecolor{graycross}{HTML}{737D85}
\definecolor{positiveheat}{HTML}{B2182B}
\DeclareRobustCommand{\colorblock}[1]{\begingroup\raisebox{.15ex}{\textcolor{#1}{\rule{1.0em}{.75ex}}}\endgroup}
\hypersetup{pdftitle={TabReliBench: Evaluating Tabular Foundation Models Beyond Predictive Accuracy},pdfauthor={Tianqi Zhao, Jia Guo, Yan Shuo Tan, Qiong Zhang}}
\title{TabReliBench: Evaluating Tabular Foundation Models Beyond Predictive Accuracy}
\author{
Tianqi Zhao\textsuperscript{1}, Jia Guo\textsuperscript{2},
Yan Shuo Tan\textsuperscript{3,*}, Qiong Zhang\textsuperscript{4,*}\\[0.5em]
\normalfont\textsuperscript{1}Renmin University of China\\
\normalfont\textsuperscript{2}Cerebras Systems, Inc.\\
\normalfont\textsuperscript{3}National University of Singapore\\
\normalfont\textsuperscript{4}Mohamed bin Zayed University of Artificial Intelligence
}
\date{}
\begin{document}
\maketitle
\begingroup
\renewcommand{\thefootnote}{\fnsymbol{footnote}}
\footnotetext[1]{\raggedright Equal supervision. Correspondence to: Yan Shuo Tan (\href{mailto:yanshuo@nus.edu.sg}{\texttt{yanshuo@nus.edu.sg}}) and Qiong Zhang (\href{mailto:qiong.zhang@mbzuai.ac.ae}{\texttt{qiong.zhang@mbzuai.ac.ae}}).}
\endgroup
\raggedbottom
\begin{abstract}
Tabular foundation models (TFMs) are increasingly popular for their strong predictive performance.
Many also provide posterior predictive distributions (PPDs), whose reliability remains less studied.
We assess reliability through point prediction, distributional quality, marginal and subgroup calibration, and interval efficiency.
We then introduce \tabrelibench, a large classification and regression benchmark for these comparisons and dataset-level analyses.
We find that TFMs' rankings change between TabArena and our benchmark and across reliability metrics.
Regression TFMs usually achieve near-nominal marginal coverage and retain narrow intervals, but this favorable marginal behavior does not imply valid conditional coverage.
Dataset-level analyses show that classification difficulty often accompanies poor calibration, whereas regression predictive performance is not a reliable indicator of calibration.
Ensembling improves predictive performance but does not consistently improve calibration of PPDs; TFMs also benefit little from conformal post-processing, which does not reliably resolve subgroup gaps.
Reliable TFM selection therefore requires multidimensional evaluation beyond aggregate rankings.
Code is available at this \href{\codeurl}{link}.
\end{abstract}

\section{Introduction}
Tabular foundation models (TFMs)~\citep{hollmann2025accurate,qu2025tabicl} are pretrained neural networks that make predictions through in-context learning and have become increasingly popular because of their strong predictive performance.
Recent models extend this approach to larger scales~\citep{qu2026tabiclv2,jager2026tabpfn35,zhang2026limix2,kong2026tabfm}.
Benchmarks such as TabArena~\citep{erickson2025tabarena} have primarily compared these models by aggregated predictive performance.
These comparisons establish the predictive strength of TFMs, but many TFMs also produce posterior predictive distributions (PPDs), whose properties are not characterized by TabArena.
Evaluating these PPDs therefore requires a broader view of reliability than predictive performance alone.

\emph{Reliability itself is multidimensional}.
In classification, accuracy measures whether a predicted label is correct, whereas calibration asks whether stated probabilities agree with observed frequencies.
In regression, a model can predict the response accurately while producing intervals that cover too few outcomes.
Proper scoring rules, including the Brier score and continuous ranked probability score (CRPS), combine calibration and sharpness \citep{gneiting2007strictly,gneiting2007probabilistic}, but a favorable score does not reveal how a model balances them.
Predictive accuracy, calibration, coverage, and interval efficiency therefore provide complementary rather than interchangeable evidence.

Recent work has begun to evaluate these properties directly.
\citet{landsgesell2026distributional} study the finite-sample behavior of TFM distributional regression under proper scoring objectives, while \citet{izbicki2026benchmarking} benchmark conditional density estimation on regression datasets and distinguish distributional accuracy from calibration.
Both studies focus on regression, leaving open whether conclusions about reliability extend to classification or vary across benchmark collections and individual datasets.

Understanding such variation requires asking whether some datasets are difficult for nearly all models or instead favor particular methods.
\citet{tschalzev2026lost} show that aggregate rankings can conceal dataset-specific model strengths.
\citet{herre2026explaining} use meta-features to predict model-family performance gaps on TabArena but find limited held-out predictability.
Separating a common dataset-level shift from a model-specific advantage requires treating the dataset as the unit of analysis and therefore demands more datasets than are typically available in tabular leaderboards.

Reliability may also be improved at inference or through post-processing.
TFMs commonly use ensembles, while conformal methods provide post-hoc interval corrections~\citep{romano2019cqr,guan2021localized,hore2023local,han2022slcp,plassier2025rectifying,zhang2024posterior}.
How these interventions affect prediction, marginal coverage, subgroup coverage, and interval width across current TFMs remains less well understood.
We address these needs through 4 contributions (appendix~\ref{sec:related-work} provides a fuller comparison with related work):
\begin{itemize}[leftmargin=*,topsep=0pt,parsep=0pt,itemsep=0pt]
\item \textbf{A large benchmark for tabular reliability.}
We evaluate modern TFMs and classical methods on 160 classification and 81 regression datasets and separate measures of prediction, calibration, coverage, and interval efficiency.

\item \textbf{Comparisons across benchmarks and criteria.}
We compare \tabrelibench{} with TabArena and show that the strongest TFMs remain competitive, while their ordering changes across dataset collections and metrics and their coverage--width profiles reveal differences hidden by aggregate ranks.

\item \textbf{Statistical analysis of dataset-level reliability.}
We show that most score variation reflects dataset differences and that meta-features predict average performance shifts for predictive metrics and identify interpretable warning signs for classification and regression reliability.

\item \textbf{Evidence on improving reliability.}
Matched ensemble and conformal studies show that small-to-moderate ensembles improve accuracy and proper scores, whereas conformal calibration offers limited benefit to TFMs because many already attain near-nominal marginal coverage.
\end{itemize}

\section{TabReliBench: A Reliability Benchmark}
\label{sec:protocol}
TabArena is a widely used benchmark for comparing the predictive performance of tabular models across datasets~\citep{erickson2025tabarena}.
Our objective is to provide a more comprehensive comparison that includes both aggregate performance summaries, as in TabArena, and dataset-level analyses.
The dataset-level analyses ask how strongly outcomes vary across datasets and whether differences among models depend on the dataset.
They also ask whether observed dataset characteristics predict when models generally perform better or worse and which models have relative advantages.
For these questions, datasets are the units of analysis, so the number of datasets determines the cross-dataset sample size.
TabArena contains only 51 datasets---38 classification and 13 regression---too few to fit dataset-level regression models with many dataset-level features.
Additional train--test splits can improve within-dataset precision but cannot increase this cross-dataset sample size.
We therefore construct \tabrelibench, a larger benchmark for tabular reliability, and describe its datasets, models for comparison, and metrics below.

\textbf{Benchmark curation.}
We draw candidate tasks from OpenML, which provides a broad and reproducible pool of tabular datasets~\citep{vanschoren2014openml}.
To make results comparable across datasets, we apply the same sampling design to every candidate task.
We evaluate each candidate dataset on five separately generated train--test splits, which we call repetitions.
The repetitions reduce sensitivity to any single partition while leaving the dataset as the unit of analysis.
In each repetition, we sample at most 10,000 observations without replacement to remain within the operating regime of all TFMs and to control computational cost.
We allocate approximately half of the sampled observations to training and half to testing.
Classification sampling is stratified so that the capped sample and both partitions retain every class, while regression sampling is uniform.
We retain only tasks for which these samples and partitions can be constructed.
Every model receives the same observations and partitions within a repetition, which makes model differences paired at the dataset and split levels.
After applying these criteria, the benchmark contains 160 classification and 81 regression datasets, whose OpenML identifiers are listed in Appendix~\ref{sec:dataset-catalog}.

\textbf{Models under evaluation.}
Using these datasets and splits, we evaluate modern TFMs and established non-foundation methods.
We include 11 TFMs: 4 TabPFN versions (TabPFNv2, TabPFNv2.5, TabPFNv3, and TabPFNv3.5)~\citep{hollmann2025accurate,grinsztajn2025tabpfn25,grinsztajn2026tabpfn3,jager2026tabpfn35}, 2 TabICL versions (TabICLv1 and TabICLv2)~\citep{qu2025tabicl,qu2026tabiclv2}, 2 TabDPT versions (TabDPT1.1 and TabDPT1.3)~\citep{ma2024tabdpt}, Causilo~\citep{cho2026causilo}, LimiX-2~\citep{zhang2026limix2}, and TabFM~\citep{kong2026tabfm}.
TabICLv1 and TabDPT1.1 support classification only, leaving 9 TFMs for regression point prediction.
LimiX-2 and TabFM do not provide regression predictive distributions, so 7 TFMs enter the distributional comparison.

We compare these TFMs with Bayesian additive regression trees (BART)~\citep{chipman2010bart} and fixed and hyperparameter-optimized configurations of RealMLP~\citep{holzmuller2024realmlp} and XGBoost~\citep{chen2016xgboost}.
BART provides a conventional probabilistic reference, while RealMLP and XGBoost represent neural-network and gradient-boosted-tree baselines.
Appendix~\ref{sec:models} specifies model versions, predictive outputs, preprocessing, inference settings, and tuning protocols.

\textbf{Evaluation metrics.}
Rather than reduce reliability to a single leaderboard score, we construct a profile for each dataset--model pair across five dimensions: point prediction, distributional quality, marginal calibration, classwise or subgroup calibration, and interval sharpness.
Table~\ref{tab:metrics} summarizes these metrics.

\begin{table}[!htbp]
\centering
\vspace{-10pt}
\caption{
Metrics grouped by the property they assess.
$\uparrow$ and $\downarrow$ indicate the preferred direction, while $\to$ nominal indicates proximity to the target coverage level.
Interval width is compared only at similar coverage.
}
\label{tab:metrics}
\resizebox{\columnwidth}{!}{
\begin{tabular}{lll}
\toprule
Property & Classification & Regression\\
\midrule
Point prediction & Accuracy $\uparrow$ & Predictive $R^2$ $\uparrow$\\
\addlinespace
Proper score & Brier score $\downarrow$ & CRPS $\downarrow$\\
\addlinespace
Marginal calibration & Confidence ECE $\downarrow$ & PIT--KS $\downarrow$, marginal coverage $\to$ nominal\\
\addlinespace
Classwise / subgroup calibration & Classwise ECE $\downarrow$ & Worst-slab coverage $\to$ nominal\\
\addlinespace
Sharpness & -- & Interval width $\downarrow$\\
\bottomrule
\end{tabular}}
\end{table}

Accuracy and predictive $R^2$ assess point prediction, whereas the Brier score and CRPS assess the full predictive distribution.
Because proper scores combine calibration and sharpness, we supplement them with direct calibration diagnostics~\citep{gneiting2007probabilistic}.
Confidence and classwise ECE diagnose classification calibration, while PIT--KS, marginal coverage, and worst-slab coverage (WSC) diagnose regression calibration.
Marginal coverage and WSC are summarized by their absolute deviations from the nominal level, whose sum is total coverage error.
Because coverage can be increased by widening an interval, normalized mean width is interpreted only at comparable coverage.
Appendix~\ref{sec:metric-definitions} provides the formal definitions, nominal coverage levels, subgroup construction, normalization, and per-repetition calculations.
Retaining the resulting profiles at the dataset level supports both aggregate comparisons of average performance and analyses of whether models improve or deteriorate together, whether their relative performance changes across datasets, and whether either pattern is predictable from dataset characteristics.
Each subsequent analysis specifies its estimand, unit of analysis, aggregation, weighting, and, where applicable, uncertainty procedure when the corresponding results are introduced.

\section{Reliability Beyond Aggregate Rankings}
\label{sec:profiles}

\subsection{Rankings Across Benchmarks and Metrics}
\begin{figure}[htbp]
\centering
\includegraphics[width=0.28501\columnwidth]{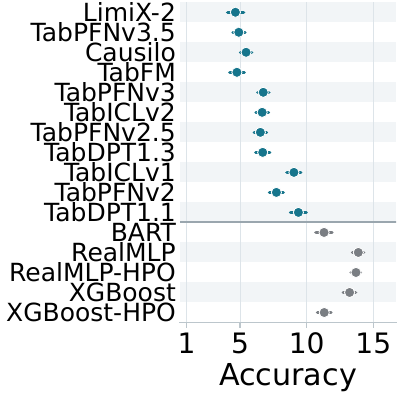}%
\includegraphics[width=0.17166\columnwidth]{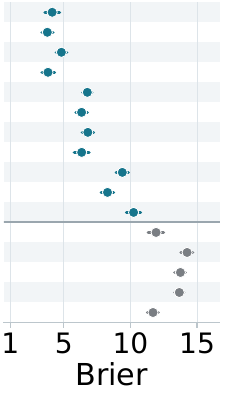}%
\includegraphics[width=0.17166\columnwidth]{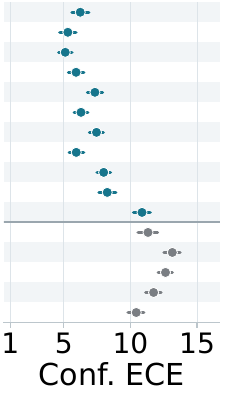}%
\includegraphics[width=0.17166\columnwidth]{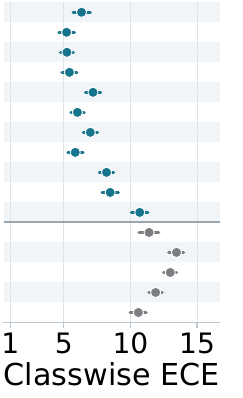}\\
\includegraphics[width=0.26721\columnwidth]{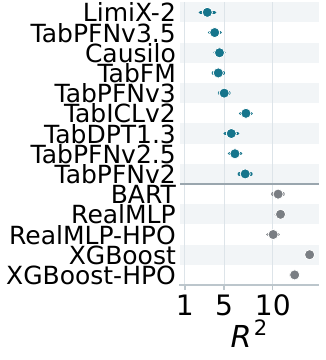}%
\includegraphics[width=0.13320\columnwidth]{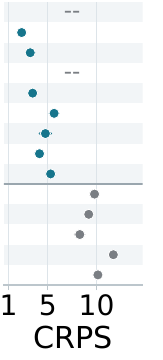}%
\includegraphics[width=0.13320\columnwidth]{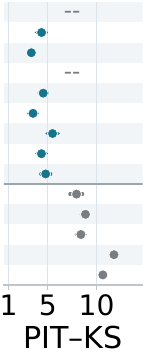}%
\includegraphics[width=0.13320\columnwidth]{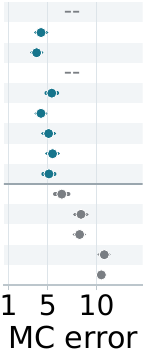}%
\includegraphics[width=0.13320\columnwidth]{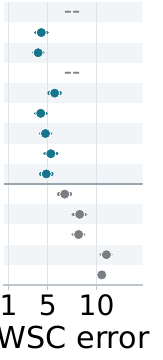}%
\vspace{-1ex}
\caption{
Mean ranks on our benchmark for classification (top) and regression (bottom); lower is better.
Lines show pointwise 95\% paired dataset-bootstrap intervals.
TFMs are ordered from best to worst on the $y$-axis (top to bottom) according to the task-specific TabArena classification and regression leaderboards recorded on October 5, 2026.
\colorblock{datasetteal} indicates TFMs, and \colorblock{baselinegray} indicates classical methods; dashes denote unavailable outputs.
MC and WSC errors are the absolute deviations of marginal coverage and worst-slab coverage, respectively, from the 90\% nominal level.
}
\label{fig:rank-overview}
\end{figure}

We ask whether the relative ordering of methods is stable across benchmarks and across measures of prediction and calibration.
For each metric, we first rank the configurations within each dataset and then average these ranks across datasets, giving every dataset equal weight.
To compare with TabArena~\citep{erickson2025tabarena}, we order the TFM rows by their positions on its leaderboard and report their metric-specific ranks on our benchmark in Figure~\ref{fig:rank-overview}.
\emph{The highest-ranked TFMs are largely the same across the two benchmarks, but several exchange positions}.
This instability may reflect benchmark-specific overfitting, although differences in the datasets and evaluation protocols provide alternative explanations.
Within our benchmark, \emph{the rankings also vary across metrics}.
In classification, LimiX-2 has the lowest mean rank for accuracy, TabPFNv3.5 for the Brier score and classwise ECE, and Causilo for confidence ECE.
In regression, LimiX-2 has the lowest mean rank for $R^2$, TabPFNv3.5 for CRPS, and Causilo for calibration metrics.

TFMs nevertheless rank ahead of the classical methods across most metrics.
In classification, all 11 TFMs rank ahead of every baseline on accuracy and the Brier score, and all except TabDPT1.1 do so on both ECE measures.
In regression, all 9 TFMs rank ahead of the baselines on $R^2$, and all 7 distributional TFMs do so on CRPS and the 4 calibration diagnostics.
Appendix~\ref{sec:support-absolute} reports the corresponding values and interquartile ranges.
Practitioners should therefore interpret TFM rankings cautiously, because the relative ordering of these models is sensitive to both the benchmark and the metrics.



\subsection{Marginal Coverage Masks Local Gaps}
\label{sec:intervals}
\begin{figure}[!htbp]
\centering
\vspace{-1ex}
\includegraphics[width=\linewidth]{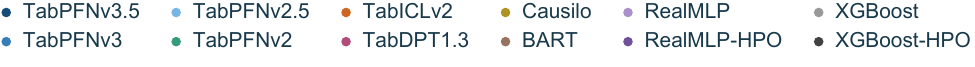}
\par\vspace{-1ex}
\includegraphics[width=0.4500\columnwidth]{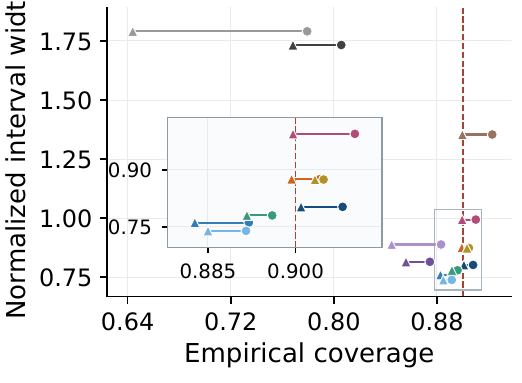}
\vspace{-3ex}
\caption{
Marginal coverage (circles) and held-out WSC (triangles) versus normalized interval width at 90\% nominal coverage over 81 regression datasets.
Coordinates are medians across datasets of five-split means; dashed lines mark nominal coverage, and the inset enlarges the TFM region.
}
\label{fig:base-intervals}
\end{figure}

Rank comparisons summarize relative performance, but they do not show how the underlying reliability measures differ.
The regression TFMs already attain marginal coverage close to the 90\% target without post-hoc calibration.
We therefore ask whether this favorable marginal behavior extends to less favorable regions of the feature space.
Figure~\ref{fig:base-intervals} compares marginal coverage and held-out WSC at the 90\% nominal level, together with the normalized width of the corresponding intervals.

\emph{The TFM intervals are generally close to nominal under both marginal coverage and WSC, although WSC is lower than marginal coverage for every configuration}.
Because WSC evaluates coverage within a selected feature slab, near-nominal WSC does not establish conditional coverage.
The difference is small for TabPFNv3.5, TabICLv2, and Causilo, which keep both coverage measures close to the target.
By contrast, BART achieves WSC near the target through marginal overcoverage and substantially wider intervals.
The RealMLP variants under-cover more strongly within the worst slab than marginally, while the XGBoost exhibits severe undercoverage under both criteria despite producing the widest intervals.

\FloatBarrier

These results provide encouraging evidence for the uncalibrated TFM intervals, while showing that marginal coverage alone does not determine worst-slab behavior or interval efficiency.
Marginal coverage, WSC, and interval width should therefore be considered jointly when comparing predictive intervals.
Appendix~\ref{sec:support-intervals} reports the corresponding comparisons at 95\%, 85\%, and 80\% nominal coverage.
The main conclusion persists at the other nominal levels.

\subsection{Decomposing Score Variation}
\label{sec:variation}
The comparisons above do not show whether datasets, models, or dataset--model interactions account for most of the observed score variation.
We therefore use a two-way analysis-of-variance decomposition to quantify these three components.
We analyze accuracy, Brier score, and both ECE measures for classification; and $R^2$, CRPS normalized by the training-response standard deviation, PIT--KS, and marginal, WSC, and total coverage errors for regression.
We first apply the decomposition to the TFMs.

\begin{figure}[t]
\centering
\begin{minipage}{0.68\linewidth}
\centering
\includegraphics[width=.49\columnwidth]{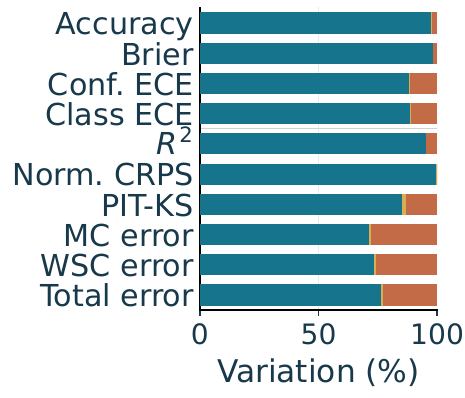}\hfill
\includegraphics[width=.49\columnwidth]{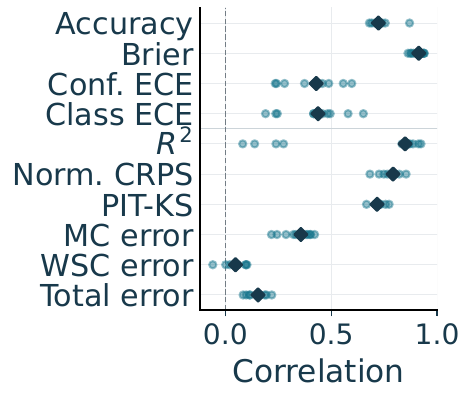}
\vspace{-1ex}
\end{minipage}
\caption{
TFM score variation across 160 classification and 81 regression datasets.
Left: proportions of the corrected total sum of squares from dataset (\colorblock{datasetteal}), model (\colorblock{modelgold}), and interaction (\colorblock{interactionorange}) terms.
Right: Pearson correlations between split-specific interaction arrays; dots show split pairs and diamonds their medians.
}
\label{fig:decomposition}
\end{figure}

For each metric, let $A_{dm}$ denote the 5-split mean score for dataset $d$ and model $m$ in a complete $D\times M$ array.
We write
\begin{equation}
A_{dm}=\mu+a_d+b_m+c_{dm},
\label{eq:decomposition}
\end{equation}
where $\mu$ is the grand mean, $a_d=\bar A_{d\cdot}-\mu$ is the dataset main effect, $b_m=\bar A_{\cdot m}-\mu$ is the model main effect, and $c_{dm}=A_{dm}-\mu-a_d-b_m$ is the dataset--model interaction.
Because every dataset is evaluated with every model, the centered terms are orthogonal and the corrected total sum of squares partitions exactly as
\[
\sum_{d,m}(A_{dm}-\mu)^2
=M\sum_d a_d^2+D\sum_m b_m^2+\sum_{d,m}c_{dm}^2.
\]
Letting $T$ denote the left-hand side, dividing each term by $T$ gives the dataset, model, and interaction proportions as shown in Figure~\ref{fig:decomposition} (left).

It is shown that \emph{most variation among TFMs comes from differences among datasets}.
For accuracy, Brier score, $R^2$, and normalized CRPS, TFMs tend to improve or deteriorate together on the same datasets, while their average differences are small.
Calibration scores, especially coverage errors, show larger dataset--model interactions, meaning that the relative calibration of TFMs varies more across datasets.
Because the decomposition depends on the methods being compared, we repeat it after adding the 5 classical methods.
The expanded comparison has larger model and interaction contributions in regression, indicating that the dominance of dataset differences is most pronounced among the more homogeneous TFMs (Appendix~\ref{sec:support-decomposition}).

The right plot shows that these dataset-specific differences are more repeatable for prediction and distributional scores than for coverage errors, although the $R^2$ correlations vary markedly across split pairs.
A split-level decomposition reinforces this result: prediction metrics remain primarily dataset-driven, whereas coverage diagnostics, especially WSC, contain substantial partition-related variation (Appendix Figure~\ref{fig:nested-five}).
These results establish that dataset differences account for much of the observed variation, but not whether performance on a new dataset can be anticipated from properties of its training data.

\section{Predicting Reliability}
\label{sec:meta}
We address this question using dataset meta-features, asking whether they predict the average performance shift across TFMs and, separately, their relative performance on a new dataset.
\citet{herre2026explaining} use meta-features to predict normalized predictive-error gaps between model families or selected model pairs on the much smaller TabArena datasets.
Our analysis instead considers prediction and calibration metrics and separates dataset-level from model-specific variation.

\subsection{Average \& Relative Performance}
Meta-features are widely used in meta-learning to represent a dataset by a fixed-dimensional vector of statistical and structural summaries~\citep{vanschoren2018meta,rivolli2018characterizing}.
Intuitively, they summarize dataset size and dimensionality, missingness and feature distributions, the distribution of the response, and associations between features and the response.
We represent the 160 classification and 81 regression datasets using 62 and 59 meta-features, respectively, all computed exclusively from training data.
To assess which sources of information improve prediction, we add meta-features cumulatively in four groups: covariate summaries ($X$), response summaries ($+Y$), direct covariate--response summaries ($+XY$), and auxiliary fitted-model summaries ($+\mathrm{Aux}$).
The auxiliary group contains cross-validated scores from simple linear and nonlinear predictors and summaries of fitted decision trees.
We compute each meta-feature separately on each training split and average it across the 5 splits; benchmark scores are averaged over the same splits, leaving one observation per dataset.
Appendix~\ref{sec:feature-definitions} provides the complete definitions and computation details.
\begin{figure}[!htbp]
\centering
\begin{minipage}{0.96\linewidth}
\centering
\includegraphics[width=.325\columnwidth]{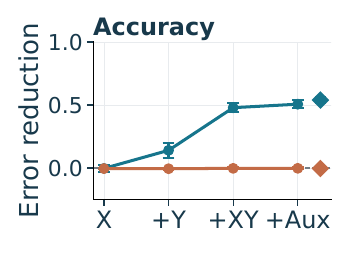}\hfill
\includegraphics[width=.325\columnwidth]{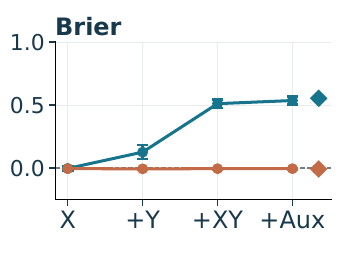}\hfill
\includegraphics[width=.325\columnwidth]{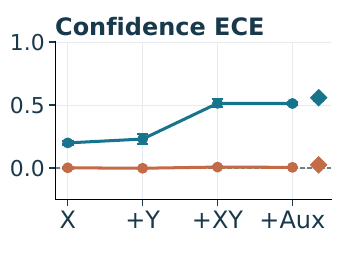}
\par\vspace{1ex}
\includegraphics[width=.24\columnwidth]{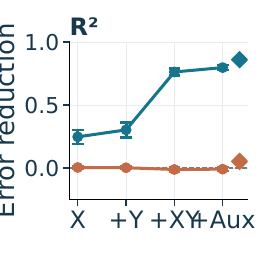}\hfill
\includegraphics[width=.24\columnwidth]{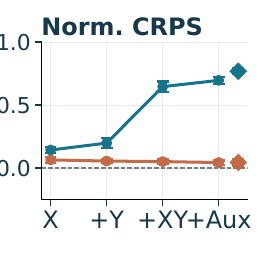}\hfill
\includegraphics[width=.24\columnwidth]{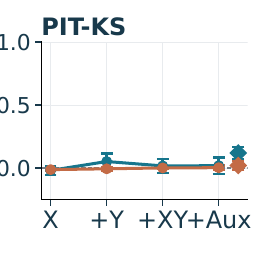}\hfill
\includegraphics[width=.24\columnwidth]{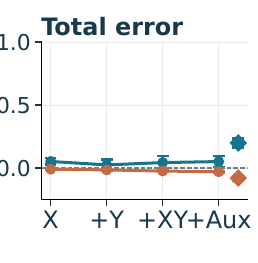}
\end{minipage}
\caption{
Held-out squared-error reduction for predicting the average score shift across TFMs (\colorblock{datasetteal}) and the relative-deviation component, with errors summed over TFMs (\colorblock{interactionorange}).
Lines show ridge regression as meta-feature groups are added, and diamonds show Extra-Trees using all meta-features.
Bars show one standard deviation across cross-validation repetitions.
}
\label{fig:components}
\end{figure}

To distinguish predictable dataset-level variation from predictable differences among TFMs, we define two held-out targets.
For a held-out dataset $d$, let $z_d$ denote its meta-feature vector and let $\widehat B_m$ be model $m$'s mean score over the training datasets.
Define
\begin{equation}
\label{eq:component-targets}
q_d = \frac{1}{M}\sum_m(A_{dm}-\widehat B_m),
h_{dm} = A_{dm}-\widehat B_m-q_d.
\end{equation}
Here $q_d$ is the average score shift across TFMs relative to their training-dataset baselines, while $h_{dm}$ is model $m$'s deviation from that average.
We fit one predictor $f(z_d)$ for $q_d$ and zero-sum model-specific predictors $g_m(z_d)$ for $h_{dm}$, giving $\widehat A_{dm}=\widehat B_m+\widehat f(z_d)+\widehat g_m(z_d)$.
Because the number of datasets is limited relative to the number of meta-features, we use regularized ridge regression as the primary meta-predictor and Extra-Trees as a nonlinear sensitivity analysis.
We estimate held-out performance using five repetitions of 5-fold cross-validation over datasets, with all baselines, targets, preprocessing, and tuning estimated within each training fold.
For both $q_d$ and $h_{dm}$, we report the reduction in held-out squared error relative to using only the model baseline $\widehat B_m$.
A value of 0 matches this baseline, 1 denotes exact prediction, and a negative value indicates worse prediction.
Appendix~\ref{sec:meta-prediction-details} gives the complete fitting and evaluation procedure.

\emph{For predictive metrics, meta-features predict the average performance shift across TFMs but provide little information about relative differences among TFMs} (Figure~\ref{fig:components}).
Direct feature--response meta-features provide the largest gains for predicting the average score shifts in accuracy, Brier score, $R^2$, and normalized CRPS, while auxiliary fitted-model summaries add comparatively little.
By contrast, average shifts in PIT--KS and coverage error remain difficult to predict, even with Extra-Trees.
The model-specific deviation component remains near the baseline for almost every metric, even when all meta-features or Extra-Trees are used.
Section~\ref{sec:variation} established that dataset-level variation is large; the present analysis shows that part of this variation can be anticipated from observable dataset properties for predictive metrics, whereas relative TFM performance remains difficult to predict.
This conclusion is robust to the evaluated models and meta-predictor (Appendix Figure~\ref{fig:component-sensitivity}).
Separate model-specific meta-regressions also predict absolute scores reasonably well, particularly for predictive metrics (Appendix Figure~\ref{fig:absolute-predictability}).
This absolute-score predictability can arise from average dataset-level shifts and is broadly consistent across TFMs, but less so across classical baselines.

\subsection{Meta-features Associated with Reliability}
\label{sec:interpretable-associations}
Having found that meta-features predict average score shifts more successfully than relative differences among TFMs, we next ask which observable dataset properties are marginally associated with typical performance and may therefore serve as practical warning signs.
For each dataset and metric, we take the median of the 5-split mean scores across all TFMs and classical baselines available for that metric, and compute its Spearman correlation with each meta-feature.
To focus on dataset characteristics rather than these auxiliary fitted-model summaries, we omit the $\mathrm{Aux}$ group, leaving 49 classification and 45 regression meta-features.
For each metric, we select the 5 largest absolute correlations and display their union without removing correlated meta-features or requiring statistical significance.
Appendix~\ref{sec:dataset-associations} gives the eligibility rules, multiple-testing correction, and individual-model results.
\begin{figure}[ht]
\centering
\begin{minipage}{0.72\linewidth}
\centering
\includegraphics[width=.85\columnwidth]{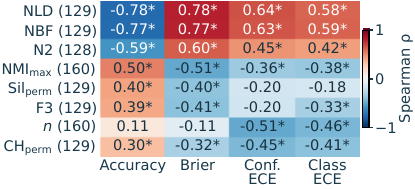}
\vspace{-1ex}
\end{minipage}
\caption{
Spearman correlations between meta-features and median classification scores across 16 methods.
Rows are the union of the five largest absolute correlations per metric; parentheses give dataset counts, and stars mark significance after controlling the false discovery rate at 5\%.
Positive values are favorable only for accuracy.
}
\vspace{-1ex}
\label{fig:classification-topk-associations}
\end{figure}

\emph{Class overlap signals both predictive and calibration difficulty} (Figure~\ref{fig:classification-topk-associations}).
Greater neighborhood disagreement (NLD), more frequent class boundaries (NBF), and weaker cluster separation ($\mathrm{Sil}_{\mathrm{perm}}$, $\mathrm{CH}_{\mathrm{perm}}$) are associated with lower accuracy, worse Brier scores, and larger calibration errors.
Neighborhood disagreement has the strongest associations, with $\rho=-0.78$ for accuracy and $\rho=0.64$ for confidence ECE.
Practitioners should therefore treat local class overlap as a warning that both predicted labels and probabilities require careful validation.
Training size ($n$) is associated with lower ECE but only weakly with accuracy and Brier score, so dataset size alone is not a reliable indicator of classification difficulty.

\begin{figure}[ht]
\centering
\begin{minipage}{0.86\linewidth}
\centering
\includegraphics[width=.9\columnwidth]{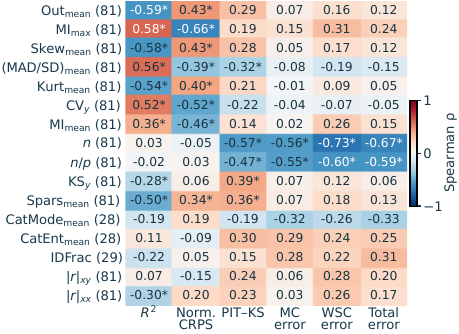}
\vspace{-1ex}
\end{minipage}
\caption{
Spearman correlations between meta-features and median regression scores across 14 methods for $R^2$ and 12 for distributional metrics.
Rows, parentheses, and stars follow Figure~\ref{fig:classification-topk-associations}; coverage errors use 90\% nominal intervals.
Positive values are favorable only for $R^2$.
}
\vspace{-1ex}
\label{fig:regression-topk-associations}
\end{figure}

\emph{Regression prediction and calibration require different diagnostics} (Figure~\ref{fig:regression-topk-associations}).
Stronger feature--target dependence ($\mathrm{MI}_{\max}$, $\mathrm{MI}_{\mathrm{mean}}$) is associated with higher $R^2$ and lower normalized CRPS, whereas feature skewness ($\mathrm{Skew}_{\mathrm{mean}}$) and outliers ($\mathrm{Out}_{\mathrm{mean}}$) accompany poorer predictive performance.
In contrast, coverage errors are most strongly associated with training size ($n$) and the number of training samples per feature ($n/p$), while target non-normality ($\mathrm{KS}_y$) is associated with PIT--KS.
Practitioners should therefore evaluate coverage and PIT calibration directly, especially when the number of training samples per feature is small, rather than infer uncertainty reliability from $R^2$ or normalized CRPS.

\section{Improving Reliability}
\label{sec:interventions}

\subsection{Ensemble Gains are Metric-dependent}
\label{sec:ensemble-size}
\begin{figure}[ht]
\centering
\includegraphics[width=\linewidth]{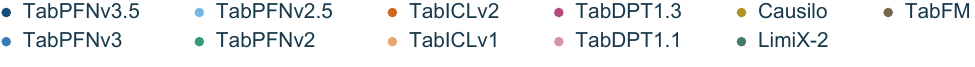}
\par\vspace{0.5ex}
\begin{minipage}{0.7\linewidth}
\centering
\vspace{-1ex}

\includegraphics[width=.49\columnwidth]{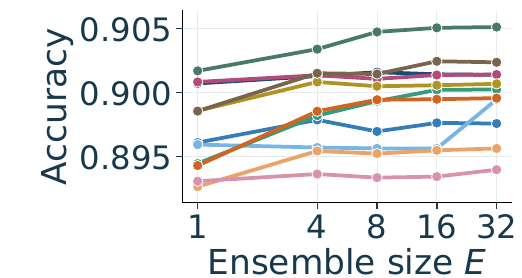}
\includegraphics[width=.49\columnwidth]{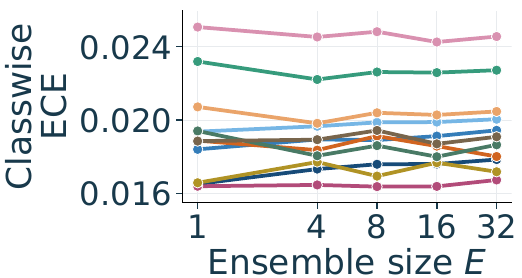}\\
\includegraphics[width=.49\columnwidth]{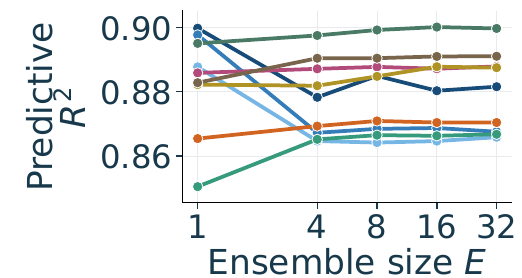}\hfill
\includegraphics[width=.49\columnwidth]{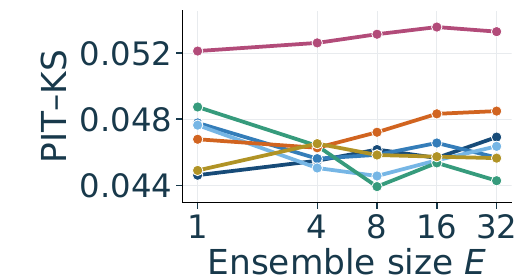}
\end{minipage}
\caption{
Model-specific median performance across ensemble sizes for classification (top) and regression (bottom).
Each point is the median across datasets after 5-seed averaging; higher values are better for predictive metrics and lower values for calibration metrics.
}
\label{fig:ensemble-size-summary}
\end{figure}

TFMs commonly average multiple ensemble members at inference, with each additional member increasing computational cost.
We therefore examine which aspects of predictive and calibration performance improve as the ensemble grows and whether these gains persist at larger ensemble sizes.
We vary the number of estimators over $E\in\{1,4,8,16,32\}$ for every TFM while holding the dataset splits and all other settings fixed.
Within each dataset and metric, we first average the five seeds and then report the median metric across the common set of datasets for every model--ensemble configuration.
Higher values are better for accuracy and $R^2$, whereas lower values are better for all other metrics.
Figure~\ref{fig:ensemble-size-summary} shows the model-specific trajectories for selected predictive and calibration metrics, while Appendix~\ref{sec:ensemble-appendix} reports every metric.

\emph{Predictive gains are mostly front-loaded, but not uniform across metrics.}
Median accuracy increases from $E=1$ to $E=32$ for all 11 classification TFMs, and 10 of 11 realize an initial gain by $E=4$.
Over the same range, the median Brier score improves for all 11 classification TFMs and normalized CRPS improves for all seven distributional regression TFMs (Appendix~\ref{sec:ensemble-median}).
Median $R^2$ is less uniform: 6 of 9 regression TFMs improve from $E=1$ to $E=32$, whereas TabPFNv3.5, TabPFNv3, and TabPFNv2.5 have lower medians at the larger ensemble size.
Across improving metrics, most gains occur at small to moderate ensemble sizes, with little change at larger sizes.

\emph{Calibration changes are smaller and model-specific.}
Classwise ECE and PIT--KS medians are comparatively stable or non-monotone as the ensemble grows, and their trajectories differ across TFMs.
From $E=1$ to $E=32$, classwise ECE improves for only 5 of 11 classification models and PIT--KS for only three of seven distributional regression models.
The coverage--width trajectories reinforce this pattern: increasing $E$ changes the operating point differently across TFMs, and marginal and worst-slab coverage need not move in the same direction (Appendix Figure~\ref{fig:ensemble-intervals-full}).
For several TFMs, movement toward 90\% coverage is accompanied by wider intervals, so better coverage alone does not establish a better coverage--width tradeoff.
Thus, larger ensembles broadly improve accuracy and proper scores, especially over the first few members, but do not reliably improve direct calibration diagnostics.
The preferred ensemble size therefore depends on the metric of interest rather than increasing uniformly with $E$.

\subsection{Conformal Gains Leave Subgroup Gaps}
\label{sec:conformal}
Because increasing ensemble size does not uniformly improve interval calibration, we next ask whether post-hoc conformal calibration can correct the coverage deficiencies in Section~\ref{sec:intervals}.
Our evaluation distinguishes marginal coverage from worst-slab coverage.
This distinction allows us to determine whether conformal calibration improves only average coverage or also reliability within heterogeneous subgroups.

We evaluate 9 post-processors.
Four global split-conformal variants use absolute-residual (SCP-ABS), normalized-residual (SCP-NORM), signed quantile (SCP-CQR; \citealp{romano2019cqr}), and probability-integral-transform (SCP-PIT) scores.
We also consider the localized methods CalLCP \citep{guan2021localized}, RLCP \citep{hore2023local}, and SLCP \citep{han2022slcp}, together with posterior conformal prediction (PCP; \citealp{zhang2024posterior}) and rectified conformal prediction (RCP; \citealp{plassier2025rectifying}).
The latter methods allow the conformal correction to vary with covariates through localized or learned adjustment, and therefore provide a stronger test of whether conformal calibration can address subgroup undercoverage.
Across the 7 distributional regression TFMs and BART, we apply each post-processor to frozen $E=8$ predictions and compare marginal coverage, worst-slab coverage, interval width, and degenerate outputs.
Appendix~\ref{sec:support-conformal} gives the data allocation, method-specific implementation, aggregation, uncertainty quantification, and complete results across nominal coverage levels.
We first use SCP-CQR as a global calibration baseline and then ask whether covariate-adaptive methods better address subgroup undercoverage.

\begin{figure}[!htbp]
\centering
\begin{minipage}{0.82\linewidth}
\centering
\vspace{-1.5ex}
\includegraphics[width=.28\columnwidth]{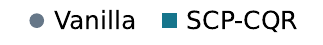}\\
\includegraphics[width=.32\columnwidth]{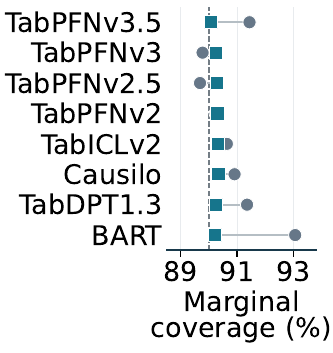}
\includegraphics[width=.32\columnwidth]{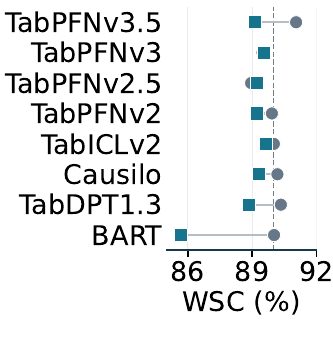}
\includegraphics[width=.32\columnwidth]{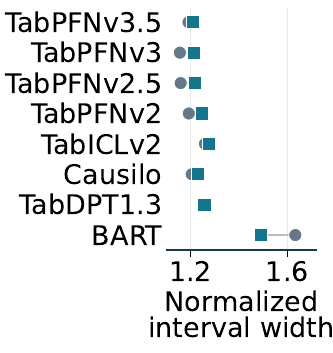}
\vspace{-1ex}
\end{minipage}
\caption{
Vanilla and SCP-CQR intervals at 90\% nominal coverage.
Connected markers show paired changes in marginal coverage (left), held-out WSC (center), and normalized interval width (right); dashed lines mark the coverage target.
}
\label{fig:conformal-intervals}
\end{figure}

\emph{Marginal calibration benefits BART more than TFMs.}
Figure~\ref{fig:conformal-intervals} compares the vanilla and SCP-CQR intervals at 90\% nominal coverage.
BART begins with the largest marginal overcoverage, and SCP-CQR brings it close to 90\% while considerably narrowing its intervals.
The uncalibrated TFMs already have marginal coverage close to the target, so SCP-CQR changes their marginal coverage and widths more modestly; TabPFNv3 and TabPFNv2.5 are widened because they initially undercover.
These results suggest that \emph{conformal calibration may be more useful for a classical predictor with substantial marginal miscalibration than for TFMs} whose intervals are already close to nominal coverage.
The marginal correction does not extend to worst-slab coverage, which remains heterogeneous and often decreases after calibration, most markedly for BART.
All 7 TFMs remain narrower than BART after calibration, so the base model continues to determine interval efficiency even after marginal coverage is aligned.
This contrast persists across nominal levels: marginal-error reductions are largest for BART at lower targets, while worst-slab effects remain heterogeneous and can be unfavorable for BART (Appendix Figure~\ref{fig:conformal-cqr}).

\begin{figure}[H]
\centering
\begin{minipage}{0.74\linewidth}
\centering
\includegraphics[width=\columnwidth]{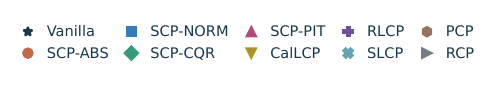}\\
\includegraphics[width=.45\columnwidth]{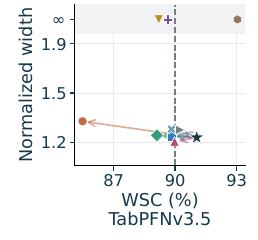}\includegraphics[width=.45\columnwidth]{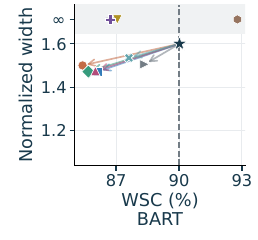}
\vspace{-1ex}
\end{minipage}
\caption{
Selected worst-slab coverage and normalized-width profiles under all post-processors at 90\% nominal coverage for TabPFNv3.5 (left) and BART (right).
Arrows begin at the vanilla intervals, the dashed line marks the coverage target, and the gray strip denotes infinite mean width.
}
\vspace{-1ex}
\label{fig:conformal-adaptive-wsc}
\end{figure}

\emph{Covariate-adaptive calibration does not resolve the subgroup gap.}
The failure of SCP-CQR to improve worst-slab coverage could reflect its use of a single global calibration threshold.
Figure~\ref{fig:conformal-adaptive-wsc} therefore compares all post-processors for TabPFNv3.5 and BART at the 90\% target.

For both base models, finite covariate-adaptive corrections do not consistently move worst-slab coverage toward 90\%; for BART, most move it farther below the target.
Across the full benchmark, some model--target combinations improve, but no post-processor consistently reduces worst-slab coverage error across models and nominal levels (Appendix Figures~\ref{fig:conformal-subgroup-levels} and \ref{fig:errors-90}--\ref{fig:errors-80}).
Worst-slab coverage is not equivalent to pointwise conditional coverage, but these results show that covariate-dependent conformal adaptation does not by itself provide reliable coverage in the poorly covered regions identified here.
PCP and, less frequently, CalLCP and RLCP produce more infinite intervals at higher target coverages, so their coverage gains can come at the cost of informativeness (Appendix Figure~\ref{fig:conformal-infinite-main}).

Overall, conformal post-processing can align marginal coverage, but neither global nor covariate-adaptive calibration consistently eliminates worst-slab undercoverage.
Reliable interval evaluation therefore requires examining marginal coverage, subgroup coverage, interval width, and degenerate outputs jointly.

\FloatBarrier
\section{Discussion \& Conclusion}
Our results show why TFM reliability cannot be reduced to a single leaderboard.
Rankings change across benchmarks and metrics, and models with similar predictive performance can differ in calibration and interval efficiency.
Most score variation is nevertheless associated with datasets rather than average differences among TFMs, so meta-features predict whether TFMs will generally perform well more readily than they predict which TFM will perform best.

This distinction has practical consequences.
Local class overlap is a warning sign for both classification accuracy and calibration, whereas regression prediction and calibration are associated with different dataset properties and should be assessed separately.
Larger ensembles improve accuracy and proper scores most consistently, but direct calibration measures change less uniformly.
Conformal calibration can correct substantial marginal miscalibration, particularly for BART, yet neither global nor covariate-adaptive methods consistently remove worst-slab undercoverage; interval width and degenerate outputs must therefore remain part of the comparison.

Our evidence is limited to the benchmark datasets, fitted configurations, and reliability criteria considered here, and worst-slab coverage does not guarantee pointwise conditional coverage.
Even so, the central conclusion is clear: predictive accuracy, marginal calibration, subgroup reliability, and interval efficiency answer different questions.
Reliable tabular prediction therefore requires jointly evaluating these properties for each target dataset and decision.

\section*{Acknowledgments}
Tianqi Zhao and Qiong Zhang are supported by the National Key R\&D Program of China Grant 2024YFA1015800.
Y.S. Tan was supported by NUS Startup Grant A-8000448-00-00 and the Singapore Ministry of Education (MOE) AcRF Tier 1 Grants A-8002498-00-00 and A-8004458-00-00.

This work was conducted while Qiong Zhang was at the Institute of Statistics and Big Data, Renmin University of China.
Qiong Zhang is now affiliated with Mohamed bin Zayed University of Artificial Intelligence.

\FloatBarrier
\bibliographystyle{apalike}
\bibliography{biblio}

\clearpage
\section*{Supplementary Materials}
\appendix
\section{Additional Related Work}
\label{sec:related-work}
\textbf{Tabular prediction and reliability.}
TFMs have become strong alternatives to dataset-specific learners \citep{hollmann2025accurate,qu2025tabicl,ma2024tabdpt}, with benchmarks such as TabArena organizing comparisons across tasks and model families \citep{erickson2025tabarena}.
Distributional evaluations complement these predictive comparisons.
\citet{landsgesell2026distributional} show that proper scoring objectives affect the finite-sample behavior of TFMs.
\citet{izbicki2026benchmarking} evaluate conditional density estimation on regression datasets and distinguish distributional accuracy from calibration.
We extend this perspective to classification and regression under a common protocol, and connect aggregate results to dataset-level variation.

\textbf{Dataset properties and model differences.}
Meta-learning uses properties of a dataset to anticipate algorithm behavior \citep{vanschoren2018meta,rivolli2018characterizing}.
Recent work shows that aggregate rankings can conceal models with distinctive strengths on individual datasets \citep{tschalzev2026lost}.
\citet{herre2026explaining} find that meta-features provide limited held-out evidence for predicting model-family performance gaps on TabArena.
We extend this question to multiple reliability outcomes by separately predicting when TFMs generally perform better or worse and which TFMs have relative advantages.

\textbf{Conformal calibration.}
Conformalized quantile regression adjusts predictive intervals using held-out conformity scores \citep{romano2019cqr}.
Local weighting, score rectification, and posterior mixtures extend this approach toward more adaptive coverage \citep{hore2023local,han2022slcp,plassier2025rectifying,zhang2024posterior}.
We evaluate these methods as post-processors of TFM and BART predictions, measuring marginal error, held-out subgroup error, width, and degenerate prediction sets on matched observations.

\section{Benchmark Datasets and Model Configurations}
\label{sec:models}

\subsection{Benchmark Datasets and Splits}

\subsubsection{Dataset Curation and Target Processing}
Starting from the frozen 51-dataset TabArena panel, we added active OpenML supervised-learning datasets with a single default target, at least 200 observations, at most 500 registered columns including the target, and a nonsparse representation. Classification datasets were additionally restricted to 2--10 classes. The sample-size floor supports repeated 50/50 train--test splits, while the column and class limits keep the additional tasks within the intended operating range of TabPFNv2. We manually excluded sequential or time-series tasks, survival or time-to-event outcomes, and other tasks incompatible with an i.i.d. tabular evaluation, and reconciled duplicate or alternative versions of the same source and prediction target. The resulting benchmark contains 160 classification and 81 regression datasets. Each task uses its designated default target. We discard observations with missing classification labels or nonfinite regression responses. Classification labels are deterministically mapped to consecutive integers, and the same mapping is used to align predicted probabilities with evaluation labels.

\subsubsection{Sampling and Train--Test Splits}
For each of five repetitions, we sample at most 10,000 observations without replacement and allocate $\lfloor N/2\rfloor$ to training and the remainder to testing. Regression observations are sampled uniformly. For classification, we use stratified sampling with class allocations as close as possible to proportional while preserving every class in the sampled data and both partitions; observations are sampled uniformly within each class.

Every model uses the same observations and outer partition within a repetition. TFMs receive the entire training partition as context, giving a maximum context size of 5,000, and BART fits on the same partition. RealMLP and XGBoost reserve an internal validation subset of the outer training partition as specified in Section~\ref{sec:baseline-parameters}. Meta-features use only the outer training partition.

\subsubsection{Complete Dataset Catalog}
\label{sec:dataset-catalog}
Table~\ref{tab:dataset-catalog} lists the OpenML dataset identifiers that define the benchmark.

\begin{table}[!ht]
\centering\small
\caption{OpenML dataset identifiers defining the benchmark.}
\label{tab:dataset-catalog}
\begin{tabular}{@{}p{.14\textwidth}p{.80\textwidth}@{}}
\toprule
Task & Dataset identifiers\\
\midrule
Classification & 3, 11, 20, 23, 24, 35, 38, 39, 43, 50, 56, 57, 60, 156, 333, 334, 335, 377, 446, 451, 455, 460, 464, 1049, 1050, 1056, 1063, 1065, 1067, 1068, 1069, 1071, 1100, 1112, 1114, 1120, 1460, 1462, 1475, 1479, 1480, 1495, 1496, 1498, 1499, 1507, 1510, 1511, 1512, 1523, 1547, 1548, 1551, 1553, 1554, 1555, 1565, 1567, 1590, 1600, 4329, 23381, 40497, 40646, 40647, 40648, 40649, 40650, 40668, 40677, 40682, 40693, 40706, 40707, 40708, 40711, 40713, 40945, 40994, 41027, 41150, 41162, 41919, 42395, 42585, 42742, 43786, 43896, 43897, 43920, 43921, 43938, 44095, 44155, 44227, 45547, 45548, 45550, 45563, 45568, 45570, 45660, 46298, 46331, 46334, 46483, 46752, 46758, 46801, 46831, 46839, 46846, 46856, 46857, 46859, 46866, 46868, 46869, 46905, 46906, 46908, 46910, 46911, 46912, 46913, 46915, 46916, 46918, 46919, 46920, 46921, 46922, 46924, 46927, 46929, 46930, 46932, 46933, 46935, 46937, 46938, 46939, 46940, 46941, 46944, 46947, 46950, 46951, 46952, 46955, 46956, 46958, 46959, 46960, 46962, 46963, 46969, 46979, 46980, 46999\\
Regression & 8, 196, 202, 204, 215, 225, 229, 230, 308, 344, 509, 511, 525, 529, 534, 538, 558, 560, 561, 564, 572, 583, 586, 588, 589, 590, 592, 593, 595, 598, 599, 606, 607, 608, 609, 612, 618, 620, 622, 623, 628, 699, 1097, 41265, 42224, 42352, 42563, 42571, 42723, 42724, 43141, 43714, 44223, 44960, 44967, 44980, 44981, 44984, 44993, 45042, 45702, 46295, 46299, 46612, 46613, 46904, 46907, 46914, 46917, 46923, 46928, 46931, 46934, 46942, 46946, 46949, 46953, 46954, 46961, 46964, 46976\\
\bottomrule
\end{tabular}

\end{table}

\subsection{Model Configurations}

\subsubsection{Model Overview}
Table~\ref{tab:model-configs} identifies the evaluated checkpoints and implementations. The comparison comprises 11 classification TFMs and nine regression TFMs, together with five conventional configurations: BART and fixed and hyperparameter-optimized variants of RealMLP and XGBoost. Seven TFMs provide regression predictive distributions; LimiX-2 and TabFM enter only the regression point-prediction comparison. In the table, C denotes class probabilities, R regression point predictions, and D regression predictive distributions.

\begin{table}[!ht]
\centering\scriptsize
\caption{Evaluated model versions, public implementations, and prediction types. A check mark indicates that the model enters the corresponding classification-probability (C), regression point-prediction (R), or regression distributional (D) comparison. TabPFN v2--v3 use the task-specific default checkpoint of the stated version.}
\label{tab:model-configs}
\begin{tabular}{@{}>{\raggedright\arraybackslash}p{.14\textwidth}>{\raggedright\arraybackslash}p{.38\textwidth}p{.07\textwidth}ccc >{\raggedright\arraybackslash}p{.14\textwidth}@{}}
\toprule
Model & Checkpoint / implementation & Source & C & R & D & Reference\\
\midrule
\multicolumn{7}{@{}l}{\textit{Tabular foundation models}}\\
TabPFNv2 & v2 default; \texttt{tabpfn} 8.0.2 & \href{https://github.com/PriorLabs/TabPFN}{code} & $\checkmark$ & $\checkmark$ & $\checkmark$ & \citep{hollmann2025accurate}\\
TabPFNv2.5 & v2.5 default; \texttt{tabpfn} 8.0.2 & \href{https://github.com/PriorLabs/TabPFN}{code} & $\checkmark$ & $\checkmark$ & $\checkmark$ & \citep{grinsztajn2025tabpfn25}\\
TabPFNv3 & v3 default; \texttt{tabpfn} 8.0.2 & \href{https://github.com/PriorLabs/TabPFN}{code} & $\checkmark$ & $\checkmark$ & $\checkmark$ & \citep{grinsztajn2026tabpfn3}\\
TabPFNv3.5 & 20260909 checkpoint; \texttt{tabpfn} 9.1.0 & \href{https://github.com/PriorLabs/TabPFN}{code} & $\checkmark$ & $\checkmark$ & $\checkmark$ & \citep{jager2026tabpfn35}\\
TabICLv1 & \texttt{classifier-v1.1-20250506}; \texttt{tabicl} 2.1.0 & \href{https://github.com/soda-inria/tabicl}{code} & $\checkmark$ & -- & -- & \citep{qu2025tabicl}\\
TabICLv2 & v2 classifier/regressor, 20260212; \texttt{tabicl} 2.1.0 & \href{https://github.com/soda-inria/tabicl}{code} & $\checkmark$ & $\checkmark$ & $\checkmark$ & \citep{qu2026tabiclv2}\\
TabDPT1.1 & \texttt{tabdpt1\_1.safetensors}; \texttt{tabdpt} 1.1.13 & \href{https://github.com/layer6ai-labs/TabDPT-inference}{code} & $\checkmark$ & -- & -- & \citep{ma2024tabdpt}\\
TabDPT1.3 & \texttt{tabdpt1\_3.safetensors}; \texttt{tabdpt} 1.3.1 & \href{https://github.com/layer6ai-labs/TabDPT-inference}{code} & $\checkmark$ & $\checkmark$ & $\checkmark$ & \citep{ma2024tabdpt}\\
Causilo & Task-specific released weights; package 1.0.3 & \href{https://github.com/nums-ai/causilo}{code} & $\checkmark$ & $\checkmark$ & $\checkmark$ & \citep{cho2026causilo}\\
LimiX-2 & \texttt{LimiX-2.ckpt}; native inference pipelines & \href{https://github.com/limix-ldm-ai/LimiX}{code} & $\checkmark$ & $\checkmark$ & -- & \citep{zhang2026limix2}\\
TabFM & v1.0.0 PyTorch classifier/regressor; package 1.0.1 & \href{https://github.com/google-research/tabfm}{code} & $\checkmark$ & $\checkmark$ & -- & \citep{kong2026tabfm}\\
\addlinespace
\multicolumn{7}{@{}l}{\textit{Conventional configurations}}\\
BART & \texttt{bartz} 0.12.1 & \href{https://github.com/bartz-org/bartz}{code} & $\checkmark$ & $\checkmark$ & $\checkmark$ & \citep{chipman2010bart}\\
RealMLP & PyTabKit 1.7.3; TD configuration & \href{https://github.com/dholzmueller/pytabkit}{code} & $\checkmark$ & $\checkmark$ & $\checkmark$ & \citep{holzmuller2024realmlp}\\
RealMLP-HPO & PyTabKit 1.7.3; default HPO space & \href{https://github.com/dholzmueller/pytabkit}{code} & $\checkmark$ & $\checkmark$ & $\checkmark$ & \citep{holzmuller2024realmlp}\\
XGBoost & XGBoost 3.2.0; fixed histogram-tree configuration & \href{https://github.com/dmlc/xgboost}{code} & $\checkmark$ & $\checkmark$ & $\checkmark$ & \citep{chen2016xgboost}\\
XGBoost-HPO & XGBoost 3.2.0; TabArena-based HPO space & \href{https://github.com/dmlc/xgboost}{code} & $\checkmark$ & $\checkmark$ & $\checkmark$ & \citep{chen2016xgboost}\\
\bottomrule
\end{tabular}
\end{table}

To hold the inference-time ensemble budget fixed across TFMs, the main benchmark uses eight members of each model's native inference ensemble ($E=8$). We use fixed pretrained weights and the full supplied training partition, subject to the model's native internal context handling. Experiments that vary ensemble size are identified separately. We apply no supervised fine-tuning or dataset-specific hyperparameter search to the TFMs.

To evaluate heterogeneous distributional outputs on a common support, we represent them on a grid of 199 quantile levels, $\tau_j=0.005j$, $j=1,\ldots,199$. We replace each quantile with the cumulative maximum along the grid to obtain a valid nondecreasing quantile function and use linear interpolation between adjacent levels.

\subsubsection{Tabular Foundation Models}
Unless specified below, each TFM uses the default preprocessing and inference recipe of the checkpoint and software version listed in Table~\ref{tab:model-configs}. We report only benchmark-level overrides and implementation choices not determined by these defaults. We retain native regression point predictions so that the point-prediction comparison reflects each model's intended estimator rather than a post-hoc summary of the common quantile grid.

\textbf{TabPFN family.}
TabPFNv2, v2.5, and v3 receive numeric columns and categorical-typed nonnumeric columns so that their native encoder can process the mixed-type OpenML inputs. The category vocabulary includes the observed context and query feature values, preventing query-only levels from being conflated with missing values, but uses no query labels. We disable the advisory pretraining-size checks to evaluate the full benchmark suite; this bypasses input guards without changing the model weights or inference recipe. All other settings retain the \texttt{tabpfn} 8.0.2 defaults. TabPFNv3.5 receives the original feature frame and uses the default recipe associated with its checkpoint and software version.

\textbf{TabICL family.}
TabICLv1 and TabICLv2 use the distinct pretrained checkpoints listed in Table~\ref{tab:model-configs}, but both are run with the \texttt{tabicl} 2.1.0 inference package and its default settings. To ensure compatibility with columns that are entirely missing in the supplied context, we convert such columns to numeric form, using zero to represent missing values, before TabICL's native encoding.

\textbf{TabDPT family.}
Because the TabDPT interfaces require numeric inputs, categorical columns are ordinal-encoded using only the context vocabulary, with unseen query categories mapped to $-1$ to avoid query-informed encoding. TabDPT1.3 receives the resulting numeric array and v1.1 the corresponding numeric frame. Both versions otherwise use their package defaults.

\textbf{Causilo.}
Causilo receives the original feature frame and uses the task-specific released weights and native preprocessing and inference recipe.

\textbf{LimiX-2.}
LimiX-2 receives the original feature frame. To match the common $E=8$ budget without dataset-specific pipeline selection, we choose eight pipelines independently of the data while maximizing preprocessing diversity. From the released no-retrieval classification recipe, we greedily select the pipeline adding the most unseen normalization, categorical-encoding, and feature-permutation settings, breaking ties by original position, then preserve the selected pipelines' original order. The resulting zero-based indices are $0,1,2,3,4,5,21,29$. The released regression recipe already contains eight pipelines: four quantile-uniform transformations with ordinal encoding and SVD, followed by four power transformations with one-hot encoding, each with a feature shuffle. We use the no-retrieval recipes so that inference depends only on the supplied benchmark context and query features. Table~\ref{tab:limix-config} specifies the classifier transformations.

\begin{table}[!ht]
\centering\small
\caption{LimiX-2 classification pipelines. Transform names follow the released recipe. ``Original'' retains original numeric features alongside the transformed representation.}
\label{tab:limix-config}
\begin{tabular}{@{}lp{.22\textwidth}p{.22\textwidth}lp{.26\textwidth}@{}}
\toprule
Index & Numeric transform & Categorical encoding & Permutation & Additional features\\
\midrule
0 & Quantile-normal, all data & Strict shuffled ordinal & Rotation & Fingerprint\\
1 & Quantile-uniform, 10 & Numeric & Shuffle & Original, SVD (64)\\
2 & Quantile-normal, 5; discrete & Shuffled ordinal & Shuffle & Fingerprint\\
3 & KDI-uniform & One-hot & Rotation & 10 interactions, fingerprint\\
4 & Quantile-normal, 5 & Shuffled ordinal & Shuffle & ---\\
5 & Robust & Strict shuffled ordinal & Rotation & SVD (32), fingerprint\\
21 & Power & None & None & 50 interactions\\
29 & Robust & Ordinal & None & Original, SVD (32), 100 interactions\\
\bottomrule
\end{tabular}
\end{table}

\textbf{TabFM.}
TabFM receives the original feature frame and otherwise uses its native preprocessing and inference defaults.

\subsubsection{Conventional Baselines}
\label{sec:baseline-parameters}

\textbf{BART.}
We use the high-level \texttt{bartz.Bart} interface in \texttt{bartz} 0.12.1 and retain its default model, prior, and MCMC settings. We keep 1,000 post-burn-in draws in total, balanced across the package's default four chains, to use a fixed Monte Carlo sample size for every dataset and task.

The remaining choices adapt the interface to heterogeneous tabular inputs and benchmark outputs. We median-impute numeric features and one-hot encode nominal features because \texttt{bartz} uses ordered threshold splits; unseen categories yield an all-zero indicator block. To prevent a categorical feature with many levels from receiving greater prior split probability, we divide each original feature's prior mass uniformly among its one-hot columns. Binary classification uses the package's probit outcome. For multiclass tasks, we fit one binary probit outcome per class and normalize the probabilities within each posterior draw. Regression uses posterior predictive outcome draws, including observation noise, so that empirical quantiles describe the predictive rather than latent-mean distribution; the point prediction is the draw mean.

\textbf{RealMLP/XGBoost validation and model selection.}
RealMLP and XGBoost share one 80/20 internal training/validation partition for each benchmark repetition. Classification validation allocation preserves classes in the fitting subset and stratifies where feasible; regression uses a random holdout. Numeric missing values are replaced by context medians (zero for an entirely missing column), and categoricals are ordinal-encoded with a separate missing level and $-1$ for unseen categories. RealMLP receives categorical column indicators for its embedding layers; XGBoost treats the encoded values as numeric features. This outer preprocessing is fitted on the benchmark training partition.

Both fixed and HPO variants select their epoch or boosting round by validation loss. Classification uses cross-entropy, and regression uses mean pinball loss over all 199 quantiles. To make the nominal search budgets comparable, both HPO variants evaluate 50 configurations and cap each candidate at 256 epochs or boosting rounds. These limits standardize the number of candidates and maximum training stages, but do not equate wall-clock time or computational cost across model families. The best validation checkpoint of the selected configuration supplies the test prediction, with no full-training refit or ensemble over HPO candidates. Both model families predict the common quantile grid directly. Their regression point prediction is the integral of the piecewise-linear quantile function over $[0,1]$, with constant extension beyond the first and last grid levels.

\textbf{RealMLP and RealMLP-HPO.}
We use the RealMLP-TD and RealMLP-HPO estimators provided by PyTabKit 1.7.3. The fixed variant follows the package's published TD configuration, while the HPO variant uses PyTabKit's unmodified \texttt{default} search space. We specify only the common experimental controls described above. For classification, the fixed variant disables label smoothing, whereas the HPO space retains the package's label-smoothing choices. For regression, both variants replace the standard point-prediction objective with mean pinball loss on the common quantile grid and integrate the resulting quantile function for point prediction.

\textbf{XGBoost and XGBoost-HPO.}
The fixed variant uses XGBoost 3.2.0 with histogram tree construction, otherwise standard tree defaults, and at most 256 boosting rounds. Classification uses the binary or multiclass logistic objective; regression uses \texttt{reg:quantileerror} with one output tree per quantile. XGBoost-HPO uses a prespecified search space based on the TabArena protocol and held fixed across datasets. We raise the learning-rate lower bound from 0.005 to 0.02 to avoid systematic undertraining under the common 256-round cap. Table~\ref{tab:xgb-search} reports the resulting distributions.

\begin{table}[!ht]
\centering\small
\caption{Prespecified TabArena-based search space for XGBoost-HPO. $\mathrm U$ is continuous uniform sampling; $\mathrm{LogU}_{\mathbb Z}$ is rounded log-uniform sampling. Column sampling by tree remains one.}
\label{tab:xgb-search}
\begin{tabular}{@{}p{.35\textwidth}p{.59\textwidth}@{}}
\toprule
Parameter & Distribution\\
\midrule
Learning rate & $\mathrm{LogU}(0.02,0.1)$\\
Maximum depth & $\mathrm{LogU}_{\mathbb Z}(4,10)$\\
Minimum child weight & $\mathrm{LogU}(10^{-3},5)$\\
Row subsampling & $\mathrm U(0.6,1)$\\
Column sampling by level / node & Independently $\mathrm U(0.6,1)$\\
$L_1$ / $L_2$ penalties & Independently $\mathrm U(10^{-4},5)$\\
Growth policy & Depthwise or lossguide, equal probability\\
Maximum leaves & $\mathrm{LogU}_{\mathbb Z}(8,1024)$\\
\bottomrule
\end{tabular}
\end{table}

\subsection{Computational Environment}
\label{app:compute}

The main benchmark evaluations ran on NVIDIA A100 (40/80\,GB) GPUs, and the conformal prediction experiments ran on an NVIDIA GeForce RTX 5090 (32\,GB) GPU. The ensemble-size experiments used a compute cluster with AMD EPYC 7532 CPU nodes and NVIDIA A100 (40\,GB) GPU nodes. Software dependencies and model-specific installation instructions are provided in the code repository.

\section{Metric Definitions and Aggregation}
\label{sec:metric-definitions}
The following definitions apply to one dataset--model--split. Scores are averaged across splits before taking dataset-level ranks, means, or medians.

\subsection{Classification}
For observation $i$, let $p_{ik}$ be the predicted probability of class $k$, $t_{ik}=\mathbf{1}\{y_i=k\}$, $\widehat y_i=\arg\max_kp_{ik}$, $c_i=\max_kp_{ik}$, and $h_i=\mathbf{1}\{\widehat y_i=y_i\}$. Accuracy and the multiclass Brier score are
\[
\mathrm{Accuracy}=\frac1n\sum_i h_i,
\qquad
\mathrm{Brier}=\frac1n\sum_i\sum_{k=1}^{K}(p_{ik}-t_{ik})^2.
\]
The Brier score sums over classes. For scores $s_i$, binary outcomes $t_i$, and 15 approximately equal-mass bins $\mathcal B_b$, define
\[
\mathrm{ECE}(s,t)=\sum_{b:|\mathcal B_b|>0}\frac{|\mathcal B_b|}{n}
\left|\frac1{|\mathcal B_b|}\sum_{i\in\mathcal B_b}(t_i-s_i)\right|.
\]
Confidence ECE is $\mathrm{ECE}(c,h)$; classwise ECE is $K^{-1}\sum_k\mathrm{ECE}(p_{\cdot k},t_{\cdot k})$. Classwise binning is performed separately for each probability column, and each class receives equal weight. Empirical quantile boundaries define the bins; ties can produce unequal bin sizes. The reported discrepancies use the empirical bin frequencies directly.

\subsection{Regression Prediction and Distributional Calibration}
Given point predictions $\widehat y_i$, predictive CDFs $F_i$, and quantile functions $Q_i$, we evaluate
\[
R^2=1-\frac{\sum_i(y_i-\widehat y_i)^2}{\sum_i(y_i-\bar y)^2},
\qquad
\mathrm{CRPS}(F_i,y_i)=2\int_0^1\rho_\tau(y_i-Q_i(\tau))\,d\tau,
\]
where $\rho_\tau(u)=u(\tau-\mathbf{1}\{u<0\})$. CRPS is averaged over test observations. Its numerical approximation is twice the trapezoidal integral of pinball loss on the 199-point grid from 0.005 to 0.995; no endpoint terms are added to this score. It retains response units, so the main predictive comparison ranks it within each dataset.

For cross-dataset decomposition and meta-prediction, we additionally use a training-scale-normalized version. For dataset $d$ and split $s$, let $s_{y,ds}$ be the population standard deviation of the training responses, computed before any model fitting. We define
\[
\widetilde{\mathrm{CRPS}}_{dms}=\frac{\mathrm{CRPS}_{dms}}{\max(s_{y,ds},10^{-10})}.
\]
The normalized score is computed separately for every dataset--model--split and only then averaged over the five splits. This uses no test-response scale and preserves the within-split ordering of CRPS while placing datasets on a common response-scale denominator. Raw CRPS remains the reported proper-score rank in the benchmark tables; normalized CRPS is used for cross-dataset squared-variation and held-out meta-prediction analyses.

For PIT values $u_i=F_i(y_i)$ and empirical CDF $\widehat H_n(t)=n^{-1}\sum_i\mathbf{1}\{u_i\le t\}$,
\[
\mathrm{PIT\mbox{--}KS}=\sup_{t\in[0,1]}|\widehat H_n(t)-t|
=\max_j\left\{j/n-u_{(j)},\ u_{(j)}-(j-1)/n\right\}.
\]
The predictive CDF is reconstructed by interpolating the monotone quantile grid. Repeated quantiles use the midpoint of their probability levels, and observations outside the grid use its endpoint levels. $R^2$, CRPS, and PIT--KS use the full test set.

\subsection{Coverage, Worst-Slab Coverage, and Width}
At nominal coverage $1-\alpha$, define $[L_i,U_i]=[Q_i(\alpha/2),Q_i(1-\alpha/2)]$ and $I_i=\mathbf{1}\{L_i\le y_i\le U_i\}$. The first $\lfloor n/2\rfloor$ observations in the randomized test order form the search set $\mathcal S$; the remainder form the evaluation set $\mathcal E$. This partition is shared by all models and nominal levels. We use $\alpha\in\{0.05,0.10,0.15,0.20\}$. Marginal coverage and normalized width are
\[
C=\frac1{|\mathcal E|}\sum_{i\in\mathcal E} I_i,
\qquad
W=\frac{|\mathcal E|^{-1}\sum_{i\in\mathcal E}(U_i-L_i)}{s_{\mathcal E}},
\]
where $s_{\mathcal E}$ is the population standard deviation of evaluation responses. The metric export floors the denominator at $10^{-10}$; the conformal mean-width summaries use trials with $s_{\mathcal E}>10^{-10}$ for every method.

For WSC, candidates use one feature at a time and require at least $m=\max\{1,\lfloor|\mathcal S|/5\rfloor\}$ search observations. Features with fewer than two observed values or fewer than two distinct values are skipped. Numeric features start from the five empirical quantile bins, with duplicate boundaries removed. We repeatedly merge the smallest undersized bin with its smaller adjacent bin, breaking ties to the left, until every remaining bin has support at least $m$. Bins are left-open and right-closed, with the outermost boundaries extended to include the observed extrema. Categorical features with more distinct levels than half their observed search sample are skipped. For the others, each level starts as a group, and the two smallest groups are repeatedly merged until all groups have support at least $m$. Missing feature values belong to no candidate. Let $\mathcal G$ denote the supported candidates. Select
\[
\widehat G=\arg\min_{G\in\mathcal G}
\frac{\sum_{i\in\mathcal S}\mathbf{1}\{x_i\in G\}I_i}
{\sum_{i\in\mathcal S}\mathbf{1}\{x_i\in G\}},
\qquad
C_{\mathrm{WSC}}=
\frac{\sum_{i\in\mathcal E}\mathbf{1}\{x_i\in\widehat G\}I_i}
{\sum_{i\in\mathcal E}\mathbf{1}\{x_i\in\widehat G\}}.
\]
Candidates are traversed in feature order and ties retain the first minimizer. The selected boundaries or category groups are applied unchanged to the evaluation observations. A selected subgroup with no evaluation observations gives an undefined WSC. The absolute errors are $|C-(1-\alpha)|$ and $|C_{\mathrm{WSC}}-(1-\alpha)|$; their sum is the total coverage error. Each absolute deviation is computed before averaging splits or datasets. WSC evaluates the selected feature slab, and the associated width remains the overall evaluation-set width.

\textbf{Conformal prediction sets.}
Crossed endpoints define an empty set with zero coverage and width. Infinite intervals retain infinite width. Empty-set and infinite-interval rates are averaged within datasets and then across datasets, giving each dataset equal weight.

\subsection{Aggregation, Ranks, and Uncertainty Intervals}
\label{sec:aggregation}
Within each dataset and configuration, we average each metric over its available repetitions. Ranks are then computed within datasets, with average ranks for ties; accuracy and $R^2$ are ranked in descending order and all losses in ascending order. Mean ranks give equal weight to each dataset. Joint point/distribution profiles use their common set of models. Their absolute coordinates, raw calibration profiles, and vanilla coverage--width profiles use the median of the dataset-level averages; regression CRPS is first normalized by its split-specific training-response standard deviation. Conformal profiles and paired coverage-error changes use their mean; relative width changes against Vanilla use the median of dataset-level ratios.

All bootstrap intervals use 10,000 resamples of dataset identifiers with replacement and the 2.5th and 97.5th percentiles. A resampled dataset carries all its model and method results, preserving pairing. For an error change, we first subtract the vanilla error on the same repetition, average within datasets, and then bootstrap the dataset means. These are pointwise 95\% intervals, without adjustment for multiple comparisons or grouping by dataset family. They reflect variation over datasets, not over new training splits. Model differences are assessed using paired differences rather than overlap between individual intervals. Variation across meta-prediction repetitions is instead reported as one sample standard deviation, as stated in the corresponding captions.

\section{Additional Reliability-Profile Results}
\label{sec:support-ranks}
\subsection{Complete Rank Comparisons}
Tables~\ref{tab:cls-ranks}--\ref{tab:reg-coverage-ranks-80} give the complete numerical rank comparisons.
The classification, regression-prediction, and 90\%-coverage entries underlie Figure~\ref{fig:rank-overview}; the remaining coverage tables repeat the same analysis at the other three nominal levels.
Within each dataset, we average the five split-level scores and then rank the configurations.
We average these ranks across datasets, giving every dataset equal weight.
For each bootstrap replicate, we sample datasets with replacement and use the same sampled datasets for all configurations and metrics within a task, thereby preserving the paired comparisons.
The resulting 95\% intervals are pointwise and therefore apply separately to each configuration--metric combination rather than simultaneously across the full figure.
They describe variation under resampling of the observed dataset identifiers rather than variation over new train--test splits.
The procedure treats dataset identifiers as exchangeable and does not account for dependence among related dataset families.
The comparisons are between individual configurations under the protocol in Section~\ref{sec:protocol}.
We do not pool configurations into a single TFM--baseline effect because doing so would give greater weight to model families represented by more configurations.
Lower ranks are better; dashes indicate unavailable distributional outputs.

\begin{table}[tbp]
\centering\small
\setlength{\tabcolsep}{4pt}
\caption{Classification mean ranks and 95\% intervals over 160 datasets and 16 configurations.}
\label{tab:cls-ranks}
\begin{tabular}{@{}lcccc@{}}
\toprule
Model & Accuracy & Brier & Conf. ECE & Classwise ECE\\
\midrule
LimiX-2 & 4.66 [4.07, 5.28] & 4.13 [3.59, 4.69] & 6.24 [5.61, 6.88] & 6.33 [5.71, 6.99]\\
TabPFNv3.5 & 4.91 [4.47, 5.38] & 3.77 [3.38, 4.19] & 5.31 [4.71, 5.91] & 5.21 [4.64, 5.80]\\
Causilo & 5.46 [5.04, 5.88] & 4.83 [4.44, 5.22] & 5.11 [4.62, 5.61] & 5.24 [4.76, 5.72]\\
TabFM & 4.77 [4.23, 5.32] & 3.82 [3.38, 4.31] & 5.92 [5.35, 6.51] & 5.42 [4.88, 5.98]\\
TabPFNv3 & 6.74 [6.34, 7.17] & 6.77 [6.42, 7.11] & 7.34 [6.78, 7.91] & 7.19 [6.64, 7.76]\\
TabICLv2 & 6.66 [6.20, 7.13] & 6.34 [5.92, 6.77] & 6.29 [5.78, 6.81] & 6.03 [5.54, 6.53]\\
TabPFNv2.5 & 6.53 [6.07, 6.99] & 6.81 [6.39, 7.23] & 7.46 [6.93, 7.97] & 6.99 [6.47, 7.53]\\
TabDPT1.3 & 6.70 [6.18, 7.23] & 6.34 [5.79, 6.90] & 5.94 [5.41, 6.49] & 5.86 [5.31, 6.45]\\
TabICLv1 & 9.04 [8.52, 9.56] & 9.39 [8.94, 9.84] & 7.99 [7.50, 8.50] & 8.21 [7.68, 8.74]\\
TabPFNv2 & 7.73 [7.21, 8.25] & 8.28 [7.81, 8.74] & 8.26 [7.64, 8.91] & 8.48 [7.88, 9.09]\\
TabDPT1.1 & 9.38 [8.79, 9.97] & 10.24 [9.71, 10.76] & 10.87 [10.24, 11.48] & 10.70 [10.07, 11.31]\\
\midrule
BART & 11.30 [10.69, 11.90] & 11.93 [11.34, 12.49] & 11.32 [10.57, 12.03] & 11.40 [10.65, 12.12]\\
RealMLP & 13.88 [13.43, 14.28] & 14.25 [13.79, 14.67] & 13.14 [12.52, 13.72] & 13.46 [12.87, 13.99]\\
RealMLP-HPO & 13.70 [13.33, 14.04] & 13.75 [13.33, 14.11] & 12.63 [12.09, 13.14] & 12.99 [12.50, 13.45]\\
XGBoost & 13.22 [12.74, 13.67] & 13.67 [13.33, 13.99] & 11.74 [11.15, 12.29] & 11.89 [11.39, 12.38]\\
XGBoost-HPO & 11.32 [10.81, 11.83] & 11.69 [11.28, 12.08] & 10.44 [9.82, 11.01] & 10.59 [9.98, 11.18]\\
\bottomrule
\end{tabular}

\end{table}

\begin{table}[tbp]
\centering\small
\caption{Regression mean ranks and 95\% intervals over 81 datasets. $R^2$ uses 14 configurations; CRPS and PIT--KS use the 12 configurations with distributional outputs.}
\label{tab:reg-prediction-ranks}
\begin{tabular}{@{}lccc@{}}
\toprule
Model & $R^2$ & CRPS & PIT--KS\\
\midrule
LimiX-2 & 3.30 [2.58, 4.05] & -- & --\\
TabPFNv3.5 & 4.05 [3.52, 4.62] & 2.33 [1.98, 2.72] & 4.37 [3.86, 4.88]\\
Causilo & 4.54 [4.09, 5.00] & 3.21 [2.85, 3.58] & 3.32 [2.94, 3.72]\\
TabFM & 4.42 [3.91, 4.94] & -- & --\\
TabPFNv3 & 5.02 [4.54, 5.51] & 3.46 [3.11, 3.81] & 4.54 [4.14, 4.95]\\
TabICLv2 & 7.25 [6.69, 7.79] & 5.65 [5.26, 6.04] & 3.49 [3.02, 3.96]\\
TabDPT1.3 & 5.75 [5.10, 6.41] & 4.77 [4.23, 5.28] & 5.49 [4.94, 6.06]\\
TabPFNv2.5 & 6.12 [5.56, 6.72] & 4.16 [3.75, 4.56] & 4.36 [3.89, 4.84]\\
TabPFNv2 & 7.17 [6.57, 7.77] & 5.30 [4.93, 5.67] & 4.79 [4.30, 5.30]\\
\midrule
BART & 10.54 [10.00, 11.05] & 9.79 [9.42, 10.12] & 7.95 [7.27, 8.57]\\
RealMLP & 10.79 [10.35, 11.20] & 9.20 [8.88, 9.47] & 8.86 [8.44, 9.22]\\
RealMLP-HPO & 10.04 [9.49, 10.56] & 8.28 [7.84, 8.69] & 8.40 [7.91, 8.83]\\
XGBoost & 13.77 [13.62, 13.89] & 11.72 [11.54, 11.86] & 11.78 [11.58, 11.91]\\
XGBoost-HPO & 12.23 [11.86, 12.56] & 10.14 [9.88, 10.38] & 10.64 [10.44, 10.83]\\
\bottomrule
\end{tabular}

\end{table}

\begin{table}[tbp]
\centering\small
\caption{Regression coverage-error mean ranks and 95\% intervals over 81 datasets and 12 distributional configurations at 95\% nominal coverage ($\alpha=.05$). MC and WSC denote marginal and worst-slab coverage; total error is the sum of their absolute deviations from nominal coverage.}
\label{tab:reg-coverage-ranks-95}
\begin{tabular}{@{}lccc@{}}
\toprule
Model & MC error & WSC error & Total error\\
\midrule
LimiX-2 & -- & -- & --\\
TabPFNv3.5 & 4.67 [4.14, 5.21] & 4.27 [3.79, 4.76] & 4.40 [3.98, 4.83]\\
Causilo & 3.90 [3.44, 4.36] & 3.85 [3.33, 4.40] & 3.67 [3.19, 4.19]\\
TabFM & -- & -- & --\\
TabPFNv3 & 5.35 [4.70, 6.00] & 5.46 [4.89, 6.07] & 5.20 [4.64, 5.78]\\
TabICLv2 & 4.31 [3.77, 4.88] & 4.60 [4.09, 5.14] & 4.06 [3.58, 4.57]\\
TabDPT1.3 & 5.86 [5.30, 6.43] & 5.10 [4.54, 5.65] & 5.40 [4.81, 5.98]\\
TabPFNv2.5 & 5.78 [5.19, 6.38] & 5.32 [4.64, 6.01] & 5.41 [4.73, 6.06]\\
TabPFNv2 & 4.92 [4.31, 5.53] & 5.04 [4.46, 5.62] & 5.01 [4.43, 5.59]\\
\midrule
BART & 6.06 [5.38, 6.72] & 7.11 [6.46, 7.74] & 7.06 [6.41, 7.69]\\
RealMLP & 8.42 [7.68, 9.11] & 8.12 [7.46, 8.75] & 8.35 [7.69, 8.95]\\
RealMLP-HPO & 8.27 [7.68, 8.85] & 7.76 [7.11, 8.36] & 8.15 [7.53, 8.72]\\
XGBoost & 10.36 [9.74, 10.93] & 10.82 [10.30, 11.27] & 10.80 [10.30, 11.25]\\
XGBoost-HPO & 10.10 [9.56, 10.59] & 10.53 [10.17, 10.86] & 10.51 [10.09, 10.86]\\
\bottomrule
\end{tabular}

\end{table}

\begin{table}[tbp]
\centering\small
\caption{Regression coverage-error mean ranks and 95\% intervals at 90\% nominal coverage ($\alpha=.10$), using the same 81 datasets, 12 distributional configurations, and metric definitions as the 95\% table.}
\label{tab:reg-coverage-ranks-90}
\begin{tabular}{@{}lccc@{}}
\toprule
Model & MC error & WSC error & Total error\\
\midrule
LimiX-2 & -- & -- & --\\
TabPFNv3.5 & 4.31 [3.78, 4.89] & 4.34 [3.76, 4.94] & 4.10 [3.58, 4.65]\\
Causilo & 3.86 [3.36, 4.39] & 4.02 [3.54, 4.52] & 3.80 [3.32, 4.31]\\
TabFM & -- & -- & --\\
TabPFNv3 & 5.41 [4.84, 5.99] & 5.71 [5.14, 6.27] & 5.44 [4.87, 5.99]\\
TabICLv2 & 4.31 [3.81, 4.80] & 4.28 [3.74, 4.85] & 4.01 [3.52, 4.53]\\
TabDPT1.3 & 5.09 [4.50, 5.69] & 4.77 [4.27, 5.30] & 4.84 [4.27, 5.42]\\
TabPFNv2.5 & 5.49 [4.91, 6.07] & 5.31 [4.70, 5.95] & 5.33 [4.75, 5.91]\\
TabPFNv2 & 5.12 [4.52, 5.72] & 4.85 [4.29, 5.43] & 4.80 [4.25, 5.35]\\
\midrule
BART & 6.44 [5.72, 7.15] & 6.74 [6.12, 7.33] & 7.26 [6.67, 7.83]\\
RealMLP & 8.41 [7.76, 9.01] & 8.27 [7.59, 8.88] & 8.43 [7.80, 9.01]\\
RealMLP-HPO & 8.27 [7.72, 8.75] & 8.16 [7.59, 8.69] & 8.36 [7.81, 8.85]\\
XGBoost & 10.79 [10.26, 11.26] & 11.01 [10.46, 11.48] & 10.99 [10.48, 11.42]\\
XGBoost-HPO & 10.49 [10.12, 10.80] & 10.53 [10.20, 10.83] & 10.64 [10.37, 10.89]\\
\bottomrule
\end{tabular}

\end{table}

\begin{table}[tbp]
\centering\small
\caption{Regression coverage-error mean ranks and 95\% intervals at 85\% nominal coverage ($\alpha=.15$), using the same 81 datasets, 12 distributional configurations, and metric definitions as the 95\% table.}
\label{tab:reg-coverage-ranks-85}
\begin{tabular}{@{}lccc@{}}
\toprule
Model & MC error & WSC error & Total error\\
\midrule
LimiX-2 & -- & -- & --\\
TabPFNv3.5 & 4.57 [4.07, 5.10] & 4.64 [4.08, 5.22] & 4.45 [3.90, 5.01]\\
Causilo & 3.96 [3.46, 4.49] & 4.33 [3.77, 4.91] & 3.85 [3.31, 4.42]\\
TabFM & -- & -- & --\\
TabPFNv3 & 5.62 [5.05, 6.20] & 5.64 [5.02, 6.24] & 5.49 [4.94, 6.04]\\
TabICLv2 & 4.10 [3.58, 4.64] & 4.07 [3.56, 4.60] & 3.75 [3.33, 4.20]\\
TabDPT1.3 & 4.83 [4.22, 5.44] & 5.20 [4.60, 5.83] & 4.95 [4.37, 5.56]\\
TabPFNv2.5 & 5.43 [4.91, 5.96] & 5.18 [4.68, 5.69] & 5.20 [4.69, 5.70]\\
TabPFNv2 & 5.25 [4.64, 5.87] & 5.19 [4.56, 5.82] & 5.22 [4.63, 5.81]\\
\midrule
BART & 6.99 [6.23, 7.73] & 6.67 [6.00, 7.30] & 7.44 [6.79, 8.05]\\
RealMLP & 8.20 [7.60, 8.78] & 7.65 [6.96, 8.31] & 7.97 [7.33, 8.58]\\
RealMLP-HPO & 7.82 [7.23, 8.38] & 7.91 [7.27, 8.52] & 8.09 [7.48, 8.64]\\
XGBoost & 11.06 [10.57, 11.49] & 11.20 [10.69, 11.63] & 11.17 [10.67, 11.57]\\
XGBoost-HPO & 10.15 [9.65, 10.59] & 10.31 [9.95, 10.64] & 10.42 [10.02, 10.77]\\
\bottomrule
\end{tabular}

\end{table}

\begin{table}[tbp]
\centering\small
\caption{Regression coverage-error mean ranks and 95\% intervals at 80\% nominal coverage ($\alpha=.20$), using the same 81 datasets, 12 distributional configurations, and metric definitions as the 95\% table.}
\label{tab:reg-coverage-ranks-80}
\begin{tabular}{@{}lccc@{}}
\toprule
Model & MC error & WSC error & Total error\\
\midrule
LimiX-2 & -- & -- & --\\
TabPFNv3.5 & 4.83 [4.29, 5.39] & 4.60 [4.01, 5.20] & 4.52 [3.98, 5.09]\\
Causilo & 4.15 [3.61, 4.70] & 3.83 [3.33, 4.35] & 3.43 [2.96, 3.94]\\
TabFM & -- & -- & --\\
TabPFNv3 & 5.61 [5.07, 6.16] & 5.62 [5.05, 6.19] & 5.61 [5.06, 6.15]\\
TabICLv2 & 3.63 [3.19, 4.08] & 4.51 [3.94, 5.07] & 3.99 [3.49, 4.49]\\
TabDPT1.3 & 4.57 [3.97, 5.19] & 4.70 [4.09, 5.33] & 4.48 [3.88, 5.11]\\
TabPFNv2.5 & 5.40 [4.85, 5.95] & 5.15 [4.61, 5.72] & 5.23 [4.70, 5.75]\\
TabPFNv2 & 5.31 [4.68, 5.94] & 5.94 [5.33, 6.56] & 5.81 [5.21, 6.42]\\
\midrule
BART & 7.59 [6.86, 8.25] & 6.54 [5.85, 7.22] & 7.41 [6.74, 8.05]\\
RealMLP & 8.22 [7.62, 8.78] & 7.63 [7.00, 8.23] & 8.07 [7.48, 8.63]\\
RealMLP-HPO & 7.76 [7.18, 8.31] & 7.89 [7.30, 8.42] & 7.88 [7.32, 8.40]\\
XGBoost & 10.95 [10.40, 11.44] & 11.15 [10.60, 11.63] & 11.17 [10.66, 11.60]\\
XGBoost-HPO & 9.98 [9.43, 10.48] & 10.43 [10.09, 10.74] & 10.40 [9.99, 10.75]\\
\bottomrule
\end{tabular}

\end{table}

\subsection{Absolute Prediction and Calibration Profiles}
\label{sec:support-absolute}
\begin{figure}[tbp]
\centering
\includegraphics[width=\textwidth]{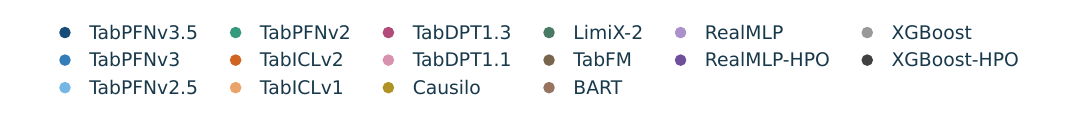}
\includegraphics[width=0.24\textwidth]{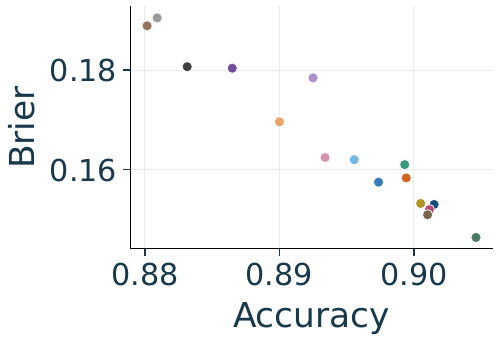}\hfill
\includegraphics[width=0.24\textwidth]{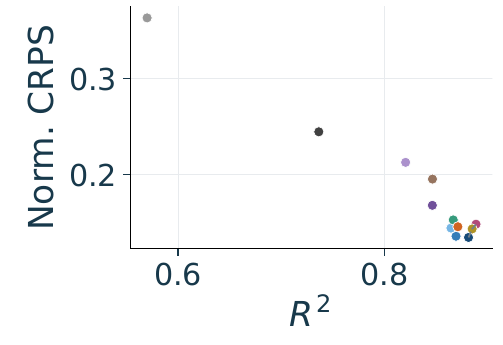}
\includegraphics[width=0.24\textwidth]{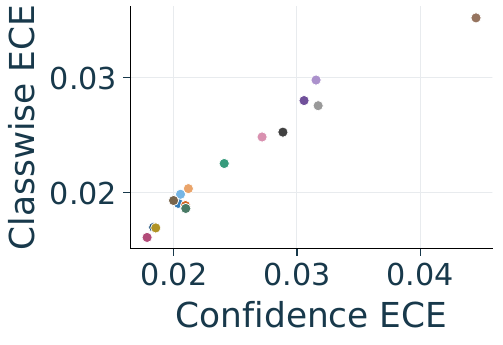}\hfill
\includegraphics[width=0.24\textwidth]{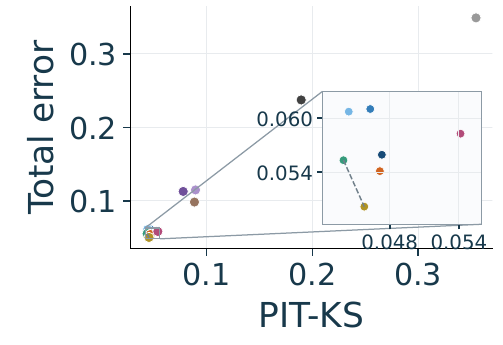}
\caption{
Absolute prediction and calibration profiles for classification and regression (reading order).
Coordinates are medians across datasets of five-split means; lower is better except for accuracy and $R^2$.
Regression CRPS is normalized by the training-response standard deviation, and point-only regression models are omitted.
The inset enlarges the outlined TFM region; dashed lines mark aggregate Pareto fronts when defined.
}
\label{fig:absolute-profiles}
\end{figure}

Ranks describe the relative ordering of methods but not the magnitude of their performance differences.
For each dataset--configuration pair, we first average scores across the five splits and then take their median across datasets, giving every dataset equal weight.
Regression CRPS is normalized by the training-response standard deviation within each split before averaging so that its scale is comparable across datasets.
Figure~\ref{fig:absolute-profiles} reports the resulting point and distributional prediction scores together with direct calibration diagnostics.
The separation between methods varies across criteria, and the regression calibration diagnostics do not always give the same ordering.
Proper scores and direct calibration diagnostics therefore provide complementary information about predictive distributions.
Tables~\ref{tab:absolute-profiles-classification} and~\ref{tab:absolute-profiles-regression} provide the numerical summaries and interquartile ranges, with the regression table additionally retaining the point-only models in the $R^2$ column.

\begin{table*}[tbp]
\centering\small
\setlength{\tabcolsep}{4.5pt}
\caption{
Classification absolute profiles over 160 datasets.
Entries are median $[Q_1,Q_3]$ of dataset-level five-split means.
Arrows indicate the preferred direction; bold marks the best displayed median.
}
\label{tab:absolute-profiles-classification}
\begin{tabular}{@{}lcccc@{}}
\toprule
Model & Accuracy $\uparrow$ & Brier $\downarrow$ & Conf. ECE $\downarrow$ & Classwise ECE $\downarrow$\\
\midrule
TabPFNv3.5 & 0.902 [0.785, 0.970] & 0.153 [0.045, 0.301] & 0.0184 [0.0077, 0.0502] & 0.0170 [0.0073, 0.0455]\\
TabPFNv3 & 0.897 [0.781, 0.970] & 0.158 [0.045, 0.310] & 0.0204 [0.0092, 0.0537] & 0.0191 [0.0086, 0.0455]\\
TabPFNv2.5 & 0.896 [0.784, 0.966] & 0.162 [0.048, 0.303] & 0.0206 [0.0112, 0.0502] & 0.0198 [0.0095, 0.0478]\\
TabPFNv2 & 0.899 [0.779, 0.966] & 0.161 [0.056, 0.309] & 0.0241 [0.0130, 0.0548] & 0.0225 [0.0110, 0.0478]\\
TabICLv2 & 0.899 [0.781, 0.970] & 0.158 [0.046, 0.306] & 0.0210 [0.0099, 0.0474] & 0.0189 [0.0087, 0.0451]\\
TabICLv1 & 0.890 [0.776, 0.964] & 0.170 [0.054, 0.312] & 0.0212 [0.0117, 0.0495] & 0.0203 [0.0101, 0.0471]\\
TabDPT1.3 & 0.901 [0.784, 0.972] & 0.152 [0.046, 0.304] & \textbf{0.0178 [0.0088, 0.0461]} & \textbf{0.0161 [0.0076, 0.0451]}\\
TabDPT1.1 & 0.893 [0.771, 0.966] & 0.162 [0.051, 0.316] & 0.0272 [0.0136, 0.0522] & 0.0248 [0.0124, 0.0493]\\
Causilo & 0.901 [0.788, 0.969] & 0.153 [0.045, 0.303] & 0.0185 [0.0076, 0.0510] & 0.0169 [0.0069, 0.0455]\\
LimiX-2 & \textbf{0.905 [0.789, 0.975]} & \textbf{0.146 [0.040, 0.302]} & 0.0210 [0.0088, 0.0499] & 0.0186 [0.0062, 0.0472]\\
TabFM & 0.901 [0.793, 0.975] & 0.151 [0.043, 0.293] & 0.0200 [0.0090, 0.0482] & 0.0193 [0.0070, 0.0433]\\
\midrule
BART & 0.880 [0.750, 0.943] & 0.189 [0.105, 0.352] & 0.0445 [0.0199, 0.0785] & 0.0352 [0.0176, 0.0677]\\
RealMLP & 0.893 [0.748, 0.953] & 0.178 [0.079, 0.351] & 0.0316 [0.0163, 0.0697] & 0.0298 [0.0144, 0.0707]\\
RealMLP-HPO & 0.886 [0.756, 0.954] & 0.180 [0.077, 0.351] & 0.0306 [0.0164, 0.0701] & 0.0280 [0.0136, 0.0638]\\
XGBoost & 0.881 [0.748, 0.949] & 0.190 [0.081, 0.346] & 0.0317 [0.0171, 0.0632] & 0.0276 [0.0161, 0.0598]\\
XGBoost-HPO & 0.883 [0.764, 0.954] & 0.181 [0.072, 0.329] & 0.0289 [0.0147, 0.0601] & 0.0253 [0.0143, 0.0570]\\
\bottomrule
\end{tabular}

\end{table*}

\begin{table*}[tbp]
\centering\small
\setlength{\tabcolsep}{4.5pt}
\caption{
Regression absolute profiles over 81 datasets.
Normalized CRPS uses the population standard deviation of the corresponding training responses; total coverage error uses 90\% nominal coverage.
Entries are median $[Q_1,Q_3]$ of dataset-level five-split means.
Point-only models have no distributional entries.
Arrows indicate the preferred direction; bold marks the best displayed median.
}
\label{tab:absolute-profiles-regression}
\begin{tabular}{@{}lcccc@{}}
\toprule
Model & $R^2$ $\uparrow$ & Norm. CRPS $\downarrow$ & PIT--KS $\downarrow$ & Total error $\downarrow$\\
\midrule
TabPFNv3.5 & 0.881 [0.563, 0.972] & \textbf{0.134 [0.088, 0.299]} & 0.0473 [0.0352, 0.0867] & 0.0559 [0.0356, 0.0761]\\
TabPFNv3 & 0.869 [0.556, 0.972] & 0.135 [0.087, 0.304] & 0.0463 [0.0349, 0.0796] & 0.0610 [0.0437, 0.0808]\\
TabPFNv2.5 & 0.864 [0.465, 0.973] & 0.144 [0.088, 0.306] & 0.0444 [0.0317, 0.0795] & 0.0607 [0.0405, 0.0813]\\
TabPFNv2 & 0.866 [0.544, 0.972] & 0.152 [0.088, 0.309] & \textbf{0.0440 [0.0363, 0.0855]} & 0.0553 [0.0361, 0.0831]\\
TabICLv2 & 0.871 [0.547, 0.969] & 0.145 [0.093, 0.304] & 0.0471 [0.0311, 0.0653] & 0.0541 [0.0315, 0.0799]\\
TabDPT1.3 & 0.888 [0.544, 0.969] & 0.148 [0.092, 0.310] & 0.0541 [0.0407, 0.0822] & 0.0583 [0.0365, 0.0924]\\
Causilo & 0.885 [0.562, 0.972] & 0.143 [0.088, 0.300] & 0.0458 [0.0304, 0.0709] & \textbf{0.0502 [0.0312, 0.0746]}\\
LimiX-2 & \textbf{0.899 [0.570, 0.978]} & -- & -- & --\\
TabFM & 0.891 [0.550, 0.973] & -- & -- & --\\
\midrule
BART & 0.846 [0.491, 0.945] & 0.195 [0.132, 0.357] & 0.0887 [0.0623, 0.1113] & 0.0983 [0.0649, 0.1130]\\
RealMLP & 0.820 [0.451, 0.942] & 0.213 [0.130, 0.351] & 0.0897 [0.0608, 0.1284] & 0.1148 [0.0632, 0.1971]\\
RealMLP-HPO & 0.846 [0.457, 0.954] & 0.168 [0.113, 0.342] & 0.0781 [0.0527, 0.1234] & 0.1128 [0.0735, 0.1579]\\
XGBoost & 0.570 [0.316, 0.722] & 0.363 [0.286, 0.412] & 0.3547 [0.2460, 0.4286] & 0.3490 [0.1804, 0.5146]\\
XGBoost-HPO & 0.736 [0.470, 0.828] & 0.244 [0.191, 0.359] & 0.1896 [0.1258, 0.2337] & 0.2373 [0.1648, 0.3254]\\
\bottomrule
\end{tabular}

\end{table*}

\subsection{Coverage--Width Profiles at Additional Nominal Levels}
\label{sec:support-intervals}
We repeat the coverage--width comparison from Section~\ref{sec:intervals} at 95\%, 85\%, and 80\% nominal coverage.
The aggregation is unchanged: each quantity is averaged across the five splits within a dataset--configuration pair and then summarized by its median across the 81 datasets.
Across the four nominal levels, TabPFNv3.5 has higher marginal and worst-slab coverage than TabPFNv3, with slightly wider intervals.
BART increasingly exceeds nominal marginal coverage as the target level decreases, whereas the XGBoost quantile configurations combine wide intervals with marked undercoverage.
RealMLP intervals can be as narrow as those of TFMs but generally remain further below nominal coverage.
Because coverage and width are summarized by separate dataset medians, these plots do not establish width orderings at equal achieved coverage.

\begin{figure}[tbp]
\centering
\includegraphics[width=\textwidth]{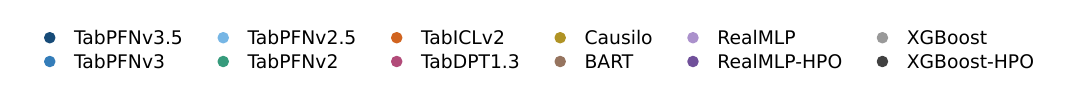}
\includegraphics[width=.32\textwidth]{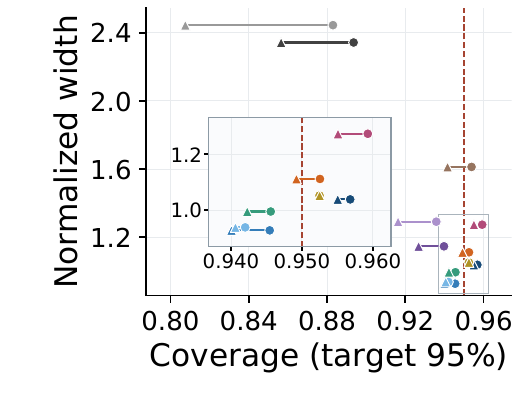}\hfill
\includegraphics[width=.32\textwidth]{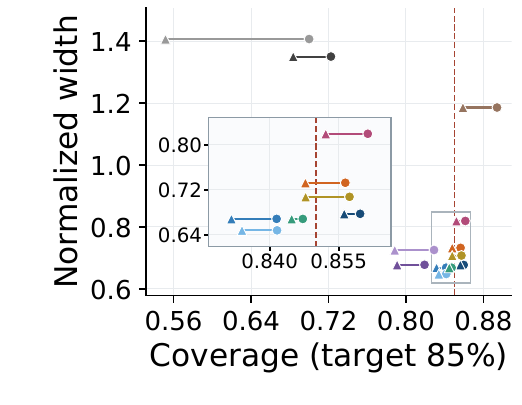}\hfill
\includegraphics[width=.32\textwidth]{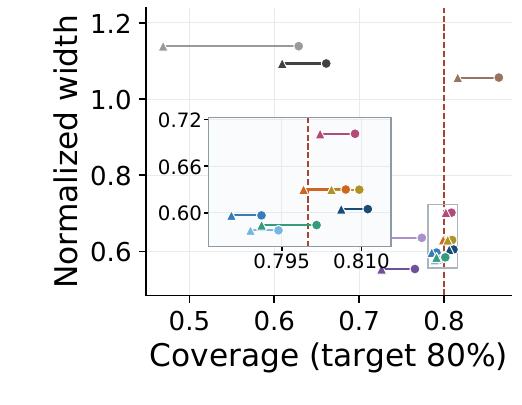}
\caption{
Coverage--width trade-offs across 81 regression datasets at 95\%, 85\%, and 80\% nominal coverage (reading order).
Coordinates are separate medians across datasets of five-split means; circles show marginal coverage and triangles WSC.
Dashed lines mark nominal coverage, and each marker pair shares one normalized interval width.
Connecting segments are visual guides rather than uncertainty intervals; insets enlarge the outlined TFM regions.
}
\label{fig:base-intervals-additional}
\end{figure}

\subsection{Score Decomposition with Conventional Baselines}
\label{sec:support-decomposition}
We repeat the analysis from Section~\ref{sec:variation} after adding the five conventional configurations.
All other choices are unchanged, including equal weighting of datasets and named configurations.

The expanded comparison preserves the classification conclusion but changes the regression picture.
Classification prediction remains dominated by datasets on which the methods generally improve or deteriorate together, whereas the more heterogeneous set of regression methods exhibits substantially larger model and dataset--model components.
The TFM-only regression comparison is comparatively homogeneous, so its persistent differences among methods across datasets are small relative to split variation.
Adding BART, RealMLP, and XGBoost introduces distinct inductive biases and, for distributional predictions, different uncertainty constructions.
The resulting model--dataset differences provide a stronger recurring interaction signal relative to split variation, which raises the split-pair correlations for the regression calibration metrics.

\begin{figure}[H]
\centering
\includegraphics[width=.34\textwidth]{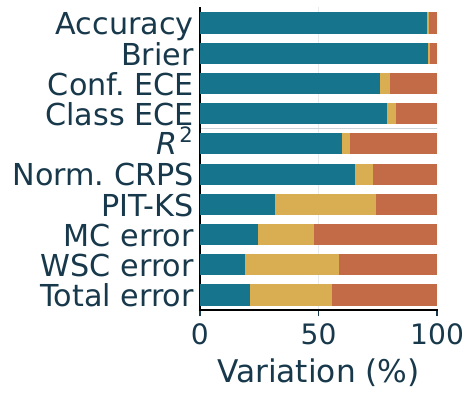}\hspace{.02\textwidth}
\includegraphics[width=.34\textwidth]{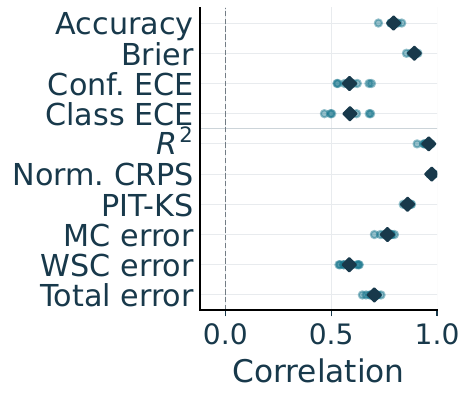}
\caption{
Score-variation decomposition after adding the five conventional configurations.
Definitions and visual encodings match Figure~\ref{fig:decomposition}.
Regression CRPS is normalized by the training-response standard deviation, and coverage errors use 90\% nominal coverage.
}
\label{fig:decomp-expanded}
\end{figure}

\subsection{Persistent Interactions versus Split Variation}
\label{sec:nested-reliability}
The split-averaged analysis cannot distinguish a recurring model--dataset interaction from residual variation due to resampling.
We therefore decompose the individual split-level scores for the same TFMs and metrics used in Section~\ref{sec:variation}.

\textbf{Exact decomposition.}
For model $m$ on split $s$ of dataset $d$, write
\begin{equation}
Y_{dms}=\mu+a_d+b_m+c_{dm}+e_{s(d)}+\varepsilon_{dms}.
\label{eq:nested-score}
\end{equation}
Here $e_{s(d)}$ is a split effect shared by all models within a dataset, and $\varepsilon_{dms}$ is the remaining model--split interaction. Dots denote averages over the corresponding indices. The empirical effects are
\[
\begin{aligned}
\mu&=\bar Y_{\cdot\cdot\cdot}, & a_d&=\bar Y_{d\cdot\cdot}-\mu, & b_m&=\bar Y_{\cdot m\cdot}-\mu,\\
c_{dm}&=\bar Y_{dm\cdot}-\bar Y_{d\cdot\cdot}-\bar Y_{\cdot m\cdot}+\mu,
& e_{s(d)}&=\bar Y_{d\cdot s}-\bar Y_{d\cdot\cdot},\\
\varepsilon_{dms}&=Y_{dms}-\bar Y_{dm\cdot}-\bar Y_{d\cdot s}+\bar Y_{d\cdot\cdot}.
\end{aligned}
\]
Their orthogonality gives
\begin{equation}
\sum_{d,m,s}(Y_{dms}-\mu)^2
=MS\sum_d a_d^2+DS\sum_m b_m^2+S\sum_{d,m}c_{dm}^2+M\sum_{d,s}e_{s(d)}^2+\sum_{d,m,s}\varepsilon_{dms}^2.
\label{eq:nested-ss}
\end{equation}
The respective degrees of freedom are $D-1$, $M-1$, $(D-1)(M-1)$, $D(S-1)$, and $D(M-1)(S-1)$. The first three sums of squares equal $S$ times their split-averaged counterparts. Figure~\ref{fig:nested-five} instead divides by total split-level variation, including both additional terms; its percentages therefore have a different denominator from Figure~\ref{fig:decomposition}.

\textbf{Variance-component estimates.}
The identity above requires no stochastic assumptions. To estimate persistent interaction variance, we additionally treat model effects as fixed and dataset, interaction, split, and residual effects as mean-zero, mutually uncorrelated random effects with homogeneous component variances. Under this working model, the relevant expected mean squares are
\[
\mathbb E[MS_{D\times M}]=\sigma_\varepsilon^2+S\sigma_c^2,\qquad
\mathbb E[MS_{S(D)}]=\sigma_\varepsilon^2+M\sigma_e^2,\qquad
\mathbb E[MS_{M\times S(D)}]=\sigma_\varepsilon^2.
\]
Writing $MS_\varepsilon=MS_{M\times S(D)}$, the method-of-moments estimates are
\[
\hat\sigma_\varepsilon^2=MS_\varepsilon,\qquad
\hat\sigma_c^2=\max\!\left\{0,\frac{MS_{D\times M}-MS_\varepsilon}{S}\right\},\qquad
\hat\sigma_e^2=\max\!\left\{0,\frac{MS_{S(D)}-MS_\varepsilon}{M}\right\}.
\]
Negative component estimates are truncated at zero. Because the splits reuse the observed samples, these estimates describe persistence and resplit variation conditional on those samples, rather than variability over independently collected datasets. Residual variation includes model-specific responses to splitting and any run variability; it is not identified as pure measurement error.

\textbf{Relation to the original interaction share.}
Let $\nu_c=(D-1)(M-1)$ and $T_{\mathrm{avg}}=\sum_{d,m}(\bar Y_{dm\cdot}-\mu)^2$. Under the working model,
\[
\mathbb E\!\left[\sum_{d,m}c_{dm}^2\right]
=\nu_c\left(\sigma_c^2+\frac{\sigma_\varepsilon^2}{S}\right).
\]
Thus, the original interaction share includes residual split variation after averaging. For direct comparison, we retain its denominator and report $\nu_c\hat\sigma_c^2/T_{\mathrm{avg}}$ as the estimated persistent contribution. The remainder is the estimated split-noise contribution (capped at the observed interaction sum of squares when truncation is needed). These are not renormalized variance-component shares. Table~\ref{tab:nested-results} and Figure~\ref{fig:nested-correction} use this convention.

\begin{table}[H]
\centering\small
\caption{TFM interaction size and persistence under the nested model. Original and corrected contributions use the same split-averaged total sum of squares. Persistence estimates refer to a single split and the mean of five splits, respectively.}
\label{tab:nested-results}
\begin{tabular}{@{}llrrrr@{}}
\toprule
Task & Metric & Original (\%) & Corrected (\%) & $\hat\rho_{\rm single}$ & $\hat\rho_{\rm mean}$\\
\midrule
Classification & Accuracy & 2.28 & 2.13 & 0.733 & 0.932 \\
Classification & Brier & 1.63 & 1.60 & 0.906 & 0.980 \\
Classification & Confidence ECE & 11.29 & 8.78 & 0.412 & 0.778 \\
Classification & Classwise ECE & 10.94 & 8.52 & 0.413 & 0.779 \\
Regression & $R^2$ & 4.50 & 3.81 & 0.528 & 0.848 \\
Regression & PIT-KS & 13.20 & 12.26 & 0.722 & 0.929 \\
Regression & Coverage error & 27.82 & 19.90 & 0.334 & 0.715 \\
Regression & WSC error & 25.96 & 4.90 & 0.045 & 0.189 \\
Regression & Total error & 22.78 & 10.48 & 0.146 & 0.460 \\
\bottomrule

\end{tabular}
\end{table}

\textbf{Persistence of individual and averaged interactions.}
We distinguish the variance ratio for a single split from that for the five-split mean:
\[
\hat\rho_{\rm single}=\frac{\hat\sigma_c^2}{\hat\sigma_c^2+\hat\sigma_\varepsilon^2},\qquad
\hat\rho_{\rm mean}=\frac{\hat\sigma_c^2}{\hat\sigma_c^2+\hat\sigma_\varepsilon^2/S}.
\]
Figure~\ref{fig:nested-persistence} compares these estimates with the original median split-pair correlation. Pointwise 95\% percentile intervals use 1,000 dataset bootstrap draws, retaining all models and splits of each sampled dataset and recomputing the estimates. They assess sensitivity to dataset composition under the same analysis, without removing uncertainty from the working assumptions.

\textbf{Results.}
The correction separates recurring departures from effects attributable to the
particular split. Brier and PIT--KS interactions remain largely persistent,
whereas much of the apparent worst-slab and total-coverage interaction is
assigned to split variation (Table~\ref{tab:nested-results}). Large uncorrected
coverage interactions therefore need not represent reproducible
model-specific behavior.

Averaging splits improves persistence, but it does not erase the contrast
between diagnostics: PIT--KS remains highly repeatable, while worst-slab
coverage remains unstable even after averaging. These ratios measure
repeatability, not predictability from meta-features.

The variance-ratio and split-pair summaries are complementary rather than
interchangeable. A median correlation can conceal a single atypical split,
whereas the variance ratio also reflects interaction magnitudes. Agreement
between both diagnostics is therefore stronger evidence of persistent
model--dataset structure than either statistic alone.

\begingroup
\setlength{\intextsep}{3pt}
\setlength{\abovecaptionskip}{3pt}
\setlength{\belowcaptionskip}{0pt}
\begin{figure}[H]
\centering
\includegraphics[width=.67\textwidth]{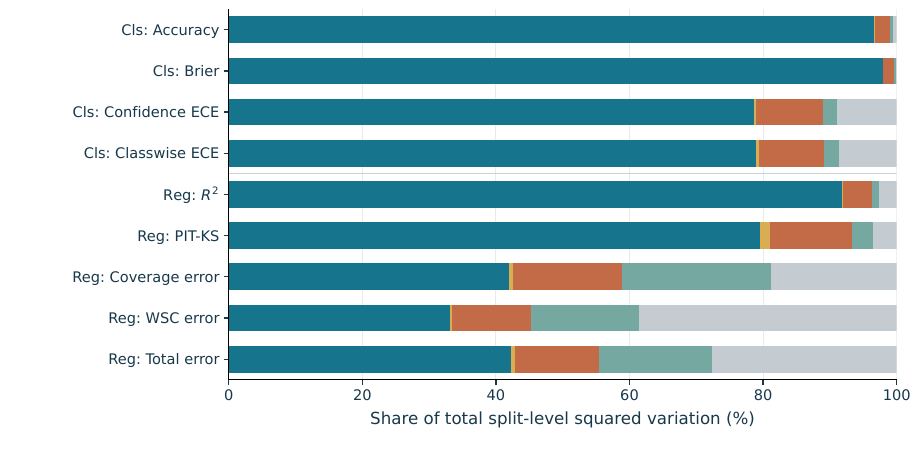}
\caption{Exact decomposition of split-level TFM scores. The five contributions sum to 100\% of total split-level squared variation. The dataset--model term still includes residual variation in the split mean; the five-term identity alone does not correct it. \colorblock{datasetteal}, \colorblock{modelgold}, \colorblock{interactionorange}, \colorblock{splitgreen}, and \colorblock{residualgray} denote dataset effects, model effects, dataset--model interactions, splits within datasets, and model--split residuals, respectively.}
\label{fig:nested-five}
\end{figure}
\begin{figure}[H]
\centering
\includegraphics[width=.67\textwidth]{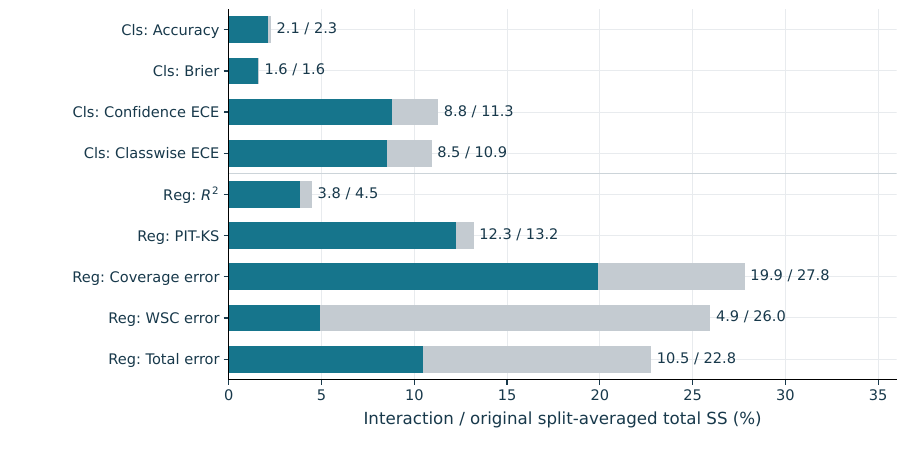}
\caption{Estimated persistent interaction contribution (\colorblock{datasetteal}) and split-noise contribution (\colorblock{residualgray}), expressed relative to the original split-averaged total squared variation. Labels give corrected/original percentages. The correction uses the nested working model; the denominator is held fixed.}
\label{fig:nested-correction}
\end{figure}
\begin{figure}[H]
\centering
\includegraphics[width=.67\textwidth]{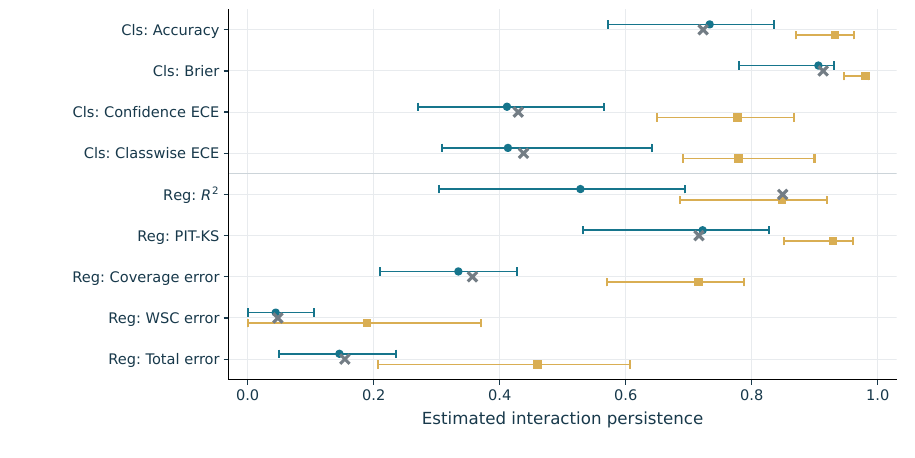}
\caption{
Estimated interaction persistence for one split (\colorblock{datasetteal}) and the five-split mean (\colorblock{medianochre}), with pointwise 95\% dataset-bootstrap intervals.
\colorblock{graycross} crosses show the original median correlation across the ten split pairs.
All estimates are conditional on the observed samples and evaluated models.
}
\label{fig:nested-persistence}
\end{figure}
\endgroup

\section{Reliability Meta-Prediction}
\label{sec:meta-prediction-details}
\subsection{Held-Out Component Targets}
The component-prediction analysis distinguishes the average score shift across models from each model's deviation from that average, but constructs every target using only the corresponding outer training fold.
In particular, the model baselines $\widehat B_m$ and the targets $q_d$ and $h_{dm}$ in Equation~\ref{eq:component-targets} are re-estimated within each fold.
Because $\sum_m h_{dm}=\sum_m\widehat g_m(z_d)=0$, held-out prediction error separates exactly into an average-shift term and a relative-deviation term:
\[
\frac1M\sum_m(A_{dm}-\widehat A_{dm})^2
=(q_d-\widehat f(z_d))^2+\frac1M\sum_m(h_{dm}-\widehat g_m(z_d))^2.
\]
This identity is verified for every held-out dataset.
The gain for the average shift is $1-\sum_d(q_d-\widehat f_d)^2/\sum_dq_d^2$; the gain for the relative deviations replaces the numerator and denominator with the corresponding sums over $h_{dm}$.
Gains are computed from all out-of-fold predictions within a repeat and then summarized across repeats.
Consequently, a positive gain means that meta-features improve on the training-fold baseline for that component, rather than merely fitting the observed datasets.

\subsection{Meta-Feature Groups}
\begin{table}[!ht]
\centering\small
\caption{
Cumulative meta-feature sets for component prediction.
Every meta-feature is computed from benchmark training data.
}
\label{tab:feature-groups}
\begin{tabular}{@{}lp{.72\textwidth}@{}}
\toprule
Group added & Information supplied\\
\midrule
$X$ & Sample size, dimensionality, missingness, numeric feature moments, inter-feature correlations, and categorical cardinality and frequency summaries.\\
$Y$ & Class balance and entropy for classification; response shape, entropy, outlier frequency, and response-bin dispersion for regression.\\
$XY$ & Direct feature--response association, mutual information, neighborhood label mixing or response variation, and class separation.\\
Aux & Scores of auxiliary linear and nonlinear predictors, supervised tree structure, and hierarchical-shrinkage tree summaries.\\
\bottomrule
\end{tabular}
\end{table}

\subsection{Meta-Predictor Configuration and Cross-Validation}
\label{sec:meta-fitting}
\textbf{Cross-validation design.}
All meta-predictors use five shuffled outer folds in each of five repetitions, with datasets as the sampling units.
Fold assignments are shared across metrics, sets of models, information groups, and learners.
Within each outer training fold, we estimate $\widehat B_m$ and construct the average-shift target $q_d$ and relative-deviation targets $h_{dm}$.
No dataset appears in both training and validation/test data.

\textbf{Meta-feature preprocessing.}
For component prediction, nonfinite meta-features are treated as missing, imputed with training-fold medians, and augmented with missingness indicators.
Entirely missing training columns are set to zero.
We then apply a training-fold quantile transform with $\min(100,D_{\mathrm{train}})$ landmarks, followed by centering and unit-variance scaling.
The same fitted transformations are applied to validation and test data and refitted within every inner fold.

\textbf{Ridge predictor.}
Ridge includes an intercept and minimizes squared error plus $\lambda\lVert\beta\rVert_2^2$, with $\lambda\in\{0.1,1,10,100,1000\}$.
Three shuffled inner folds select separate penalties for predicting $q_d$ and $h_{dm}$.
Baseline means and targets are recomputed within each inner training fold; model-specific outputs receive equal weight, and ties select the smallest penalty.

\textbf{Extra-Trees predictor.}
Extra-Trees uses 100 trees, squared-error splits, minimum leaf size four, minimum split size two, 70\% feature subsampling, no bootstrap sampling, and unrestricted depth.
A scalar regressor predicts $q_d$, and a multi-output regressor predicts the model-specific values $h_{dm}$.
We subtract the mean prediction across models from the predicted $h_{dm}$ values to enforce the zero-sum constraint.
Extra-Trees has no hyperparameter search; the expanded set of models is evaluated separately, while incremental information comparisons use Ridge with the TFMs.

\subsection{Additional Meta-Prediction Results}

\subsubsection{Sensitivity to the Models and Meta-Predictor}
Across choices of models and meta-predictors, meta-features predict average score shifts substantially better than differences among models on the predictive metrics.
Adding the classical methods exposes additional calibration signal, while model-specific deviations remain weakly predictable.
\begin{figure}[!ht]
\centering
\includegraphics[width=.45\textwidth]{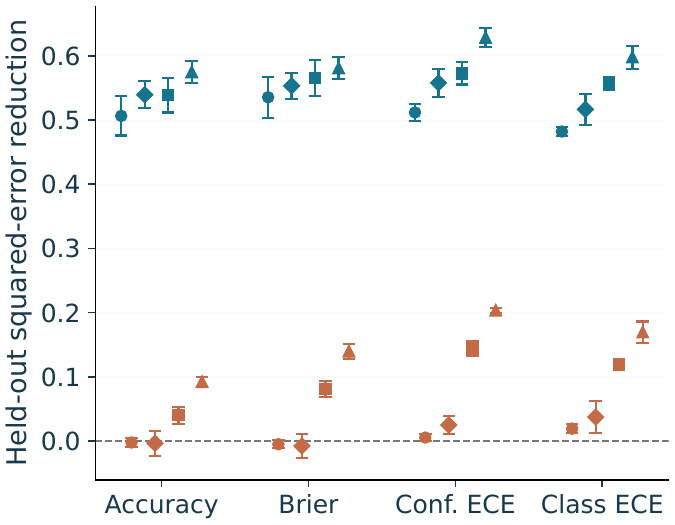}
\includegraphics[width=.45\textwidth]{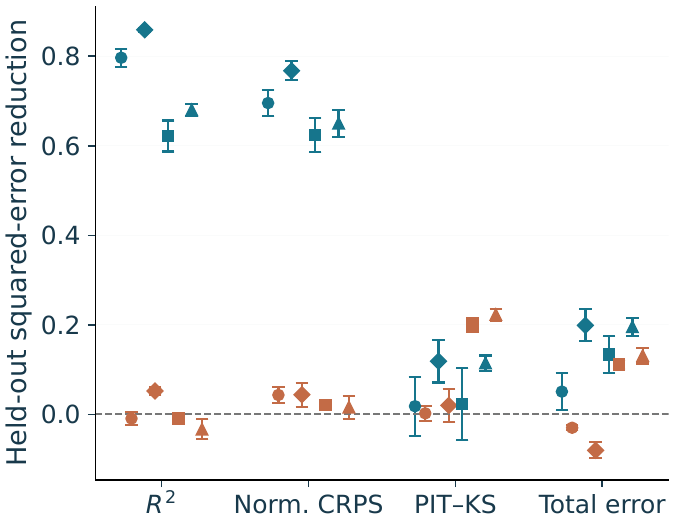}
\caption{
Sensitivity of component prediction to the evaluated models and meta-predictor.
Left: classification; right: regression.
\colorblock{datasetteal} denotes the average shift across models, and \colorblock{interactionorange} denotes model-specific deviations.
$\circ$ TFM Ridge; $\diamond$ TFM Extra-Trees; $\square$ expanded-set Ridge; $\triangle$ expanded-set Extra-Trees.
}
\label{fig:component-sensitivity}
\end{figure}

\subsubsection{Individual-Model Score Prediction}
\label{sec:support-meta}
\begin{figure}[!ht]
\centering
\includegraphics[width=.325\textwidth]{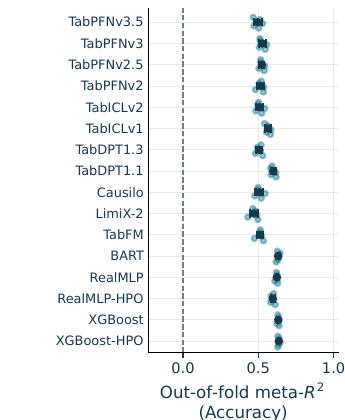}\hfill
\includegraphics[width=.325\textwidth]{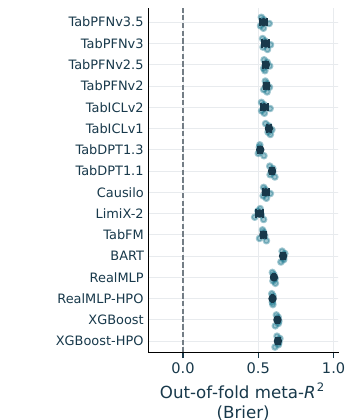}\hfill
\includegraphics[width=.325\textwidth]{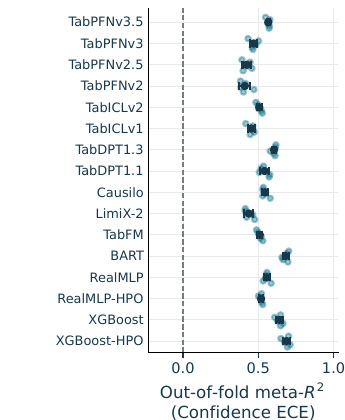}
\par\vspace{1ex}
\includegraphics[width=.325\textwidth]{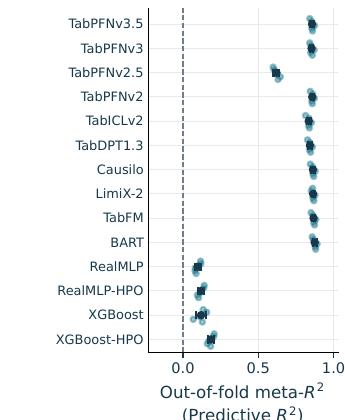}\hfill
\includegraphics[width=.325\textwidth]{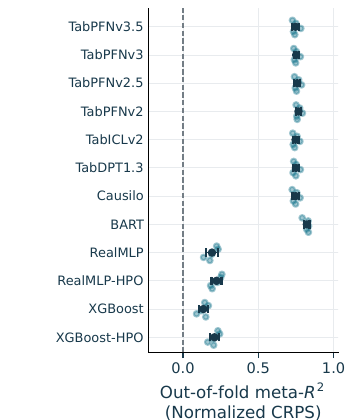}\hfill
\includegraphics[width=.325\textwidth]{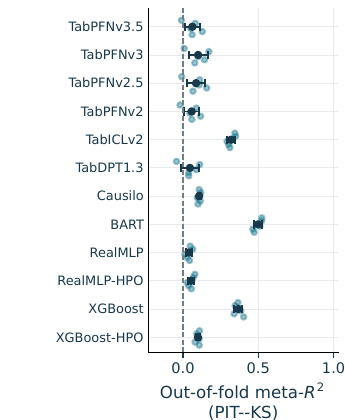}
\caption{
Out-of-fold $R^2$ for predicting each model's score from dataset meta-features.
Dark points and bars show means and one standard deviation across repetitions, while \colorblock{repeatteal} points show individual repetitions.
Panels show accuracy, Brier score, and confidence ECE for classification (top) and $R^2$, normalized CRPS, and PIT--KS for regression (bottom).
}
\label{fig:absolute-predictability}
\end{figure}

Complementary Extra-Trees meta-regressions predict each model's absolute score from the full meta-feature set, using the tree parameters above and the same repeated dataset-level folds.
They use training-fold median imputation with missingness indicators, without the quantile transform or standardization.
Each model--metric response is fitted separately.
For each repetition, meta-prediction $R^2$ is computed from all out-of-fold predictions with the overall response mean in its denominator.
Predictability is strongest for classification scores and regression predictive metrics (Figure~\ref{fig:absolute-predictability}).
Across model-specific fits, out-of-fold $R^2$ ranges from $0.41$ to $0.69$ for classification scores, from $0.10$ to $0.87$ for regression $R^2$, and from $0.14$ to $0.82$ for normalized CRPS.
PIT--KS is less predictable, with out-of-fold $R^2$ ranging from $0.04$ to $0.50$.

\subsection{Meta-Feature Reference Catalog}
\label{sec:feature-definitions}
\subsubsection{Overview and Shared Conventions}
We organize the catalog according to the four cumulative information groups introduced in Table~\ref{tab:feature-groups}: covariate-only ($X$), response-only ($Y$), direct feature--response ($XY$), and auxiliary predictive meta-features (Aux).
The complete set contains 62 meta-features for classification and 59 for regression.
Table~\ref{tab:metafeature-inventory} reports both the number added by each group and the resulting cumulative dimension.
Every meta-feature is computed separately on each benchmark training partition and then averaged across repetitions.

\begin{table}[H]
\centering\small
\caption{
Incremental and cumulative meta-feature counts.
Parentheses give the number added by each information group.
}
\label{tab:metafeature-inventory}
\begin{tabular}{@{}lrrrr@{}}
\toprule
Task & $X$ & $+Y$ & $+XY$ & $+\mathrm{Aux}$\\
\midrule
Classification & 27 & $33\,(+6)$ & $49\,(+16)$ & $62\,(+13)$\\
Regression & 27 & $36\,(+9)$ & $45\,(+9)$ & $59\,(+14)$\\
\bottomrule
\end{tabular}
\end{table}

\textbf{Data handling.}
Size, missingness, categorical summaries, and response-only meta-features use the training frame.
For numeric calculations, we retain numeric columns with at most 50\% missing entries.
Joint calculations use rows complete across these retained columns; distances, PCA, and auxiliary learners use their zero-mean, unit-variance representation.
Marginal feature moments instead drop missing values separately within each column.
A column contributes to moment summaries when it has at least four observations and standard deviation above $10^{-10}$.
Undefined meta-features are recorded as missing and handled within each meta-training fold.

\textbf{Statistical conventions.}
Standard deviations and variances use population normalization unless specified otherwise. Entropy uses natural logarithms, while skewness and excess kurtosis use empirical central moments. For a nonnegative vector $v$, we define Gini concentration as $2\sum_{j=1}^J jv_{(j)}/(J\sum_jv_j)-(J+1)/J$ and set it to zero when the total is below $10^{-10}$.

\subsubsection{Covariate-Only Meta-Features \texorpdfstring{($X$)}{(X)}}
\label{sec:x-metafeatures}
The 27 covariate-only meta-features characterize dataset size, feature types, marginal distributions, dependence, and unsupervised geometry.
They are shared by classification and regression.

\begin{table}[H]
\centering\small
\caption{
The 27 covariate-only meta-features are shared by both tasks.
``Mean/max'' denotes two distinct meta-features.
Let $n$ be training size, $p$ total feature count, and $p_n$ retained numeric feature count.
}
\label{tab:x-features}
\begin{tabular}{@{}p{.27\textwidth}rp{.61\textwidth}@{}}
\toprule
Meta-feature family & Count & Definition\\
\midrule
Size and dimensionality & 4 & $n$, $p$, $n/p$, and the number of principal components needed for 95\% explained variance divided by $p_n$.\\
Feature types / missingness & 3 & Fractions of nonnumeric columns, columns with exactly two observed values, and missing cells.\\
Near-constant features & 1 & Fraction of numeric columns with standard deviation divided by range below 0.01; columns with at most one observation also count. The range denominator is floored at $10^{-10}$.\\
Inter-feature dependence & 2 & Mean/max absolute Pearson correlation over distinct pairs of nonconstant numeric columns.\\
Neighborhood geometry & 1 & Mean distance to five non-self nearest neighbors, divided by its mean under five independent column-wise permutations.\\
Feature skewness & 2 & Mean/max absolute skewness across qualifying numeric columns.\\
Feature kurtosis & 2 & Mean/max excess kurtosis across qualifying numeric columns.\\
Feature outliers & 1 & Mean fraction more than three standard deviations from the column mean.\\
Feature bimodality & 1 & Mean $(\mathrm{skew}^2+1)/(\mathrm{kurtosis}+3)$ across columns.\\
Feature concentration & 1 & Mean frequency of each column's most common value.\\
Robust feature scale & 1 & Mean median absolute deviation divided by standard deviation.\\
Scale dispersion & 1 & Coefficient of variation of numeric-column standard deviations.\\
Categorical cardinality & 3 & Mean, maximum, and population variance of categorical level counts.\\
ID-like categories & 1 & Number of categorical columns with at least 20 observations and unique-value fraction at least 0.9, divided by $p$. Such columns are excluded from the other categorical summaries.\\
Categorical frequencies & 3 & Across-column means of entropy divided by log cardinality, mode frequency, and total probability mass in levels with frequency below 1\%. Missing values are excluded.\\
\bottomrule
\end{tabular}
\end{table}

The neighborhood-geometry meta-feature requires at least ten complete observations.
Its permutation reference preserves the observed feature marginals through five independent column-wise permutations.

\subsubsection{Response-Only Meta-Features \texorpdfstring{($Y$)}{(Y)}}
\label{sec:y-metafeatures}
The response-only group adds six meta-features for classification and nine for regression, increasing the cumulative dimensions to 33 and 36, respectively.

\begin{table}[H]
\centering\small
\caption{
Response-only meta-features.
Class frequencies here use the full training response vector.
}
\label{tab:y-features}
\begin{tabular}{@{}p{.22\textwidth}rp{.66\textwidth}@{}}
\toprule
Task / family & Count & Definition\\
\midrule
Classification & 6 & Number of classes $K$; entropy $-\sum_kp_k\log p_k/\log K$; largest and smallest class frequencies; their ratio; normalized class Gini $(1-\sum_kp_k^2)/(1-1/K)$.\\
Regression shape & 4 & Signed skewness, excess kurtosis, standard deviation divided by absolute mean, and $(\mathrm{skew}^2+1)/(\mathrm{kurtosis}+3)$.\\
Regression normality / tails & 3 & KS distance between standardized responses and $\mathcal N(0,1)$, and fractions with absolute standardized response above two and three.\\
Regression entropy & 1 & Entropy of the response histogram using ten equal-width bins from its minimum to maximum.\\
Response-bin dispersion & 1 & Partition responses at their 20th, 40th, 60th, and 80th percentiles. Among bins with more than five observations, divide the largest standard deviation by the smallest, flooring the denominator at $10^{-10}$.\\
\bottomrule
\end{tabular}
\end{table}

\subsubsection{Feature--Response Meta-Features \texorpdfstring{($XY$)}{(XY)}}
\label{sec:xy-metafeatures}
The direct feature--response group adds 16 meta-features for classification and nine for regression, increasing the cumulative dimensions to 49 and 45, respectively.
Table~\ref{tab:xy-features} summarizes the constituent families.

\begin{table}[H]
\centering\small
\caption{
Direct feature--response meta-features: 16 for classification and nine for regression.
Distances use standardized numeric features.
}
\label{tab:xy-features}
\begin{tabular}{@{}p{.23\textwidth}rp{.65\textwidth}@{}}
\toprule
Task / family & Count & Definition\\
\midrule
Both: MI summaries & 5 & Mean and maximum MI (entropy-normalized for classification), MI entropy, MI Gini concentration, and weak-signal fraction.\\
Classification: ANOVA & 2 & Mean/max $\eta_j^2=F_j(K-1)/[F_j(K-1)+n_c-K]$, where $F_j$ is the finite one-way ANOVA statistic and $n_c$ is complete-case sample size.\\
Classification: neighbors & 2 & Mean fraction of the five nearest neighbors having another label; fraction of observations with at least one such neighbor.\\
Classification: N2 & 1 & Sum of nearest same-class distances divided by sum of nearest other-class distances, over observations with both distances defined.\\
Classification: range overlap & 2 & For each class pair, form each feature's 5--95\% ranges. F2 averages intersection length divided by total span over features and pairs. F3 averages over pairs the maximum fraction of their observations outside a feature's intersection.\\
Classification: clustering & 4 & Silhouette, Davies--Bouldin, Calinski--Harabasz, and mean within-class radius divided by mean inter-centroid distance, each divided by its mean under five label permutations. Silhouette uses $(s+1)/(s_{\mathrm{null}}+1)$.\\
Regression: correlations & 3 & Mean/max absolute Pearson feature--response correlation; mean absolute Spearman correlation minus mean absolute Pearson correlation.\\
Regression: local variation & 1 & Mean response variance among each observation's five nearest neighbors, divided by overall response variance.\\
\bottomrule
\end{tabular}
\end{table}

\textbf{Mutual-information summaries.}
For each feature, we estimate its mutual information (MI) with the response using the scikit-learn estimator with three nearest neighbors. Numeric features are treated as continuous and use complete numeric rows. Categorical features use the full training sample, a separate missing-value category, and the estimator's discrete-feature option. Write $I_j$ for the resulting MI values, $J$ for their count, and $w_j=I_j/(\sum_l I_l+10^{-10})$. MI concentration uses the Gini coefficient of $(I_j)$; MI entropy is $-\sum_jw_j\log w_j/\log J$. The weak-signal fraction is $J^{-1}\sum_j\mathbf1\{I_j<0.05\max(\max_lI_l,10^{-10})\}$. Classification divides the mean and maximum MI by response entropy computed on the complete numeric rows.

\textbf{Neighborhood and clustering conventions.}
The neighborhood and clustering meta-features require at least ten complete observations.
Clustering references use label permutations, preserving the observed label distribution.
The observed silhouette uses all complete observations; each permuted silhouette uses at most 1,000 uniformly sampled observations.
Other clustering indices use all complete observations.
Class radii and centroid-pair distances are averaged with equal weight per class and per pair, respectively.

\subsubsection{Auxiliary Meta-Features (Aux)}
\label{sec:aux-metafeatures}
The auxiliary group contributes 13 meta-features for classification and 14 for regression.
Auxiliary prediction scores contribute three and four meta-features, respectively; the threshold-tree, tree-shape, and hierarchical-shrinkage families contribute two, five, and three meta-features for both tasks.

\textbf{Auxiliary prediction scores.}
Linear classification uses logistic regression with $C=1$, an $L_2$ penalty, the SAGA solver, and at most 500 iterations; linear regression uses ridge with penalty 1.
Nonlinear prediction uses gradient boosting with 50 trees, maximum depth 3, and learning rate 0.1.
The auxiliary response is balanced accuracy for classification and $R^2$ for regression.
We use five-fold cross-validation when at least 50 complete observations are available and three folds for 20--49 observations.
Classification folds are stratified; regression folds are ordinary K-fold partitions.
These scores contribute two meta-features.
Classification scores are chance-adjusted as $a^*=(a-1/K)/(1-1/K)$.
A third meta-feature is nonlinear score divided by $\max(\text{linear score},0.01)$, using adjusted classification scores.
Regression additionally includes $r/\max(1-r,0.01)$, where $r=\max(R^2_{\mathrm{nonlinear}},0)$.

\textbf{Threshold-tree meta-features.}
To measure attainable tree prediction, we evaluate CART maximum depths $1,\ldots,30$ using five shuffled folds (or $\min(5,\lfloor n_c/2\rfloor)$ folds for smaller samples).
We stop at the first depth attaining mean balanced accuracy or $R^2$ of at least 0.8.
The two retained meta-features are whether this target was attained and the score at the stopping depth (depth 30 otherwise); classification again uses chance adjustment.
This procedure requires at least ten complete observations.

\textbf{Tree-shape meta-features.}
We fit one CART with at most $L=\max(5,\lfloor\sqrt{n_c}\rfloor)$ leaves, using Gini impurity for classification and squared error for regression.
Its five meta-features are $1-H(p_{\mathrm{leaf}})/\log L_{\mathrm{fit}}$, coefficient of variation of leaf depths, Gini concentration and maximum of impurity-based feature importances, and fraction of numeric features used in splits.
Here $L_{\mathrm{fit}}$ is its actual leaf count.
This procedure also requires at least ten complete observations.

\textbf{Hierarchical-shrinkage meta-features.}
We use the same leaf budget $L$ and select $\lambda$ from
\[
\lambda/n_c\in\{0,10^{-3},10^{-2},0.1,0.5,1,10\}.
\]
Classification minimizes cross-validated log loss on shuffled stratified folds. Regression selects $\lambda$ by mean squared error on unshuffled K-fold partitions. The fold count is five for $n_c\ge50$ and three for $20\le n_c<50$; classification requires every class to support that many folds. Shrinkage recursively updates node predictions by
\[
v_j^{\mathrm{HS}}=v_{\mathrm{parent}(j)}^{\mathrm{HS}}
+\frac{v_j-v_{\mathrm{parent}(j)}}{1+\lambda/N_{\mathrm{parent}(j)}},
\]
with the root prediction unchanged.
The three meta-features are $\sum_{j\in\mathrm{internal}}N_j/(N_j+\lambda)$ divided by $\max(1,L_{\mathrm{fit}}-1)$, the selected ratio $\lambda/n_c$, and predictive quality.
Classification quality is $1-\mathrm{CV\ logloss}/H(Y)$ at the selected penalty; regression quality is $R^2$ from shuffled cross-validation at that penalty.

\section{Dataset Associations}
\label{sec:dataset-associations}

\subsection{Association Protocol}
\label{sec:association-inference}
\paragraph{Responses.}
For metric $r$, let $s_{dm}^{(r)}$ denote configuration $m$'s score on dataset $d$, averaged over the five benchmark splits.
The pooled response is $u_d^{(r)}=\operatorname{median}_{m\in\mathcal M_r}s_{dm}^{(r)}$, with equal weight for every configuration that reports the metric: 16 for classification, 14 for regression $R^2$, and 12 for regression distributional metrics.
We average meta-features over the same splits and compute $\rho_{fr}=\operatorname{Spearman}(z_{df},u_d^{(r)})$ across datasets.
We analyze four classification metrics and six regression metrics: $R^2$, normalized CRPS, PIT--KS, and marginal-coverage (MC), worst-slab-coverage (WSC), and total coverage errors at $\alpha=0.10$.
Normalized CRPS uses each split's training-response standard deviation.
Thus, the pooled analyses correlate meta-features with median scores, rather than aggregating model-specific correlations.

\paragraph{Candidate meta-features.}
We restrict the analysis to properties measured directly from the training data.
Specifically, we exclude the prespecified Aux group in Table~\ref{tab:feature-groups}, which comprises outputs of auxiliary predictors and supervised tree summaries.
This leaves 49 classification and 45 regression meta-features; direct feature--response and neighborhood or class-geometry statistics remain eligible.
Aux is retained in the component-prediction analysis of Section~\ref{sec:meta}.
Appendix~\ref{sec:feature-definitions} defines all meta-features.

\paragraph{Selection and inference.}
We require finite complete pairs from at least 30 classification or 20 regression datasets, at least three distinct meta-feature values, and a nonconstant response.
For each metric, we select the five eligible meta-features with largest $|\rho_{fr}|$ and display their union across metrics within each task, resolving exact ties by meta-feature identifier.
We apply neither significance filtering nor redundancy pruning; the resulting displays contain eight classification and 16 regression meta-features and report the available sample size for each row.

We compute two-sided Spearman $p$-values and apply Benjamini--Hochberg correction over all eligible tests before display selection.
Families are defined separately by task, metric, and response level; individual-model families additionally pool models.
Stars mark $q<0.05$, while all selected coefficients remain visible.
These in-sample associations are exploratory: they do not estimate independent effects, establish causality, or measure held-out predictability, and they do not account for dependence among related datasets.

\paragraph{Abbreviations.}
Table~\ref{tab:association-abbreviations} lists the abbreviations used for the meta-features in Figures~\ref{fig:classification-topk-associations} and~\ref{fig:regression-topk-associations}; Appendix~\ref{sec:feature-definitions} gives their definitions.

\begin{table}[tbp]
\centering\small
\caption{Meta-feature abbreviations in Figures~\ref{fig:classification-topk-associations} and~\ref{fig:regression-topk-associations}.}
\label{tab:association-abbreviations}
\begin{tabular}{@{}ll@{\hspace{2.5em}}ll@{}}
\toprule
\multicolumn{2}{@{}l}{Classification} & \multicolumn{2}{l@{}}{Regression}\\
\cmidrule(r{2.5em}){1-2}\cmidrule{3-4}
Abbreviation & Meta-feature & Abbreviation & Meta-feature\\
\midrule
NLD & Neighbor label disagreement & $\mathrm{Out}_{\mathrm{mean}}$ & Mean feature-outlier fraction\\
NBF & Neighborhood boundary fraction & $\mathrm{MI}_{\max}$ & Maximum feature--target MI\\
N2 & Within/between-class distance & $\mathrm{MI}_{\mathrm{mean}}$ & Mean feature--target MI\\
$\mathrm{NMI}_{\max}$ & Maximum normalized MI & $\mathrm{Skew}_{\mathrm{mean}}$ & Mean feature skewness\\
$\mathrm{Sil}_{\mathrm{perm}}$ & Permutation-normalized silhouette & $\mathrm{Kurt}_{\mathrm{mean}}$ & Mean feature kurtosis\\
F3 & Maximum feature efficiency & $(\mathrm{MAD/SD})_{\mathrm{mean}}$ & Mean feature MAD / SD\\
$\mathrm{CH}_{\mathrm{perm}}$ & Permutation-normalized CH index & $\mathrm{CV}_y$ & Target coefficient of variation\\
$n$ & Training sample size & $\mathrm{KS}_y$ & Target non-normality (KS)\\
 & & $n$ & Training sample size\\
 & & $n/p$ & Training samples / features\\
 & & $\mathrm{Spars}_{\mathrm{mean}}$ & Mean feature sparsity\\
 & & $\mathrm{CatMode}_{\mathrm{mean}}$ & Mean modal-category fraction\\
 & & $\mathrm{CatEnt}_{\mathrm{mean}}$ & Mean normalized category entropy\\
 & & IDFrac & ID-like categorical fraction\\
 & & $|r|_{xy}$ & Mean absolute feature--target correlation\\
 & & $|r|_{xx}$ & Mean absolute inter-feature correlation\\
\bottomrule
\end{tabular}
\end{table}

\subsection{Individual-Model Association Atlas}
\label{sec:individual-model-associations}
For each task and metric, we replace the across-model median response with each model's score averaged over the five benchmark splits.
For every model, we rank the eligible non-auxiliary meta-features by absolute Spearman correlation, retain the top five, and display the union of the selected meta-features against all available models.
The eligibility criteria and task--metric-level Benjamini--Hochberg correction are otherwise identical to Section~\ref{sec:association-inference}.
This construction allows a meta-feature to enter even when its association is specific to one model; it is therefore a descriptive comparison of model-specific patterns, not a cross-validated meta-prediction experiment or a test of between-model differences.
Column abbreviations V, ICL, DPT, MLP, and XGB denote TabPFN, TabICL, TabDPT, RealMLP, and XGBoost; H denotes the HPO configuration.

\begin{figure}[!htbp]\centering
\includegraphics[width=\textwidth,height=.81\textheight,keepaspectratio]{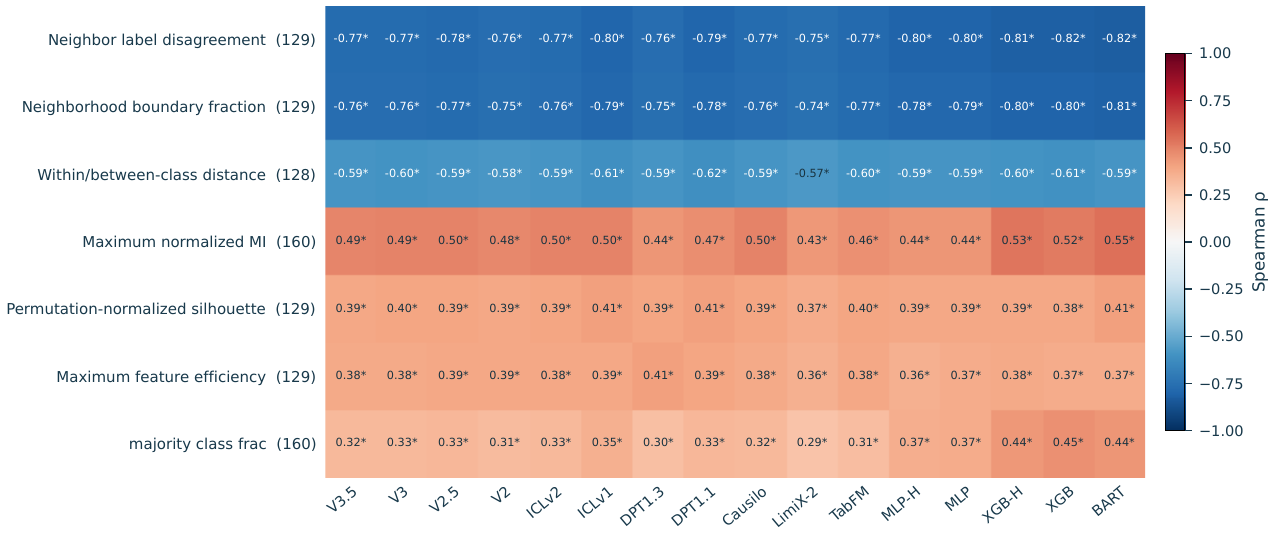}
\caption{Classification: individual-model associations with Accuracy. Rows are the union of each model’s top five non-auxiliary descriptors; parentheses give complete-pair dataset counts. Stars indicate BH $q<0.05$ over all eligible descriptor--model tests for this task and metric; unstarred cells retain their signed estimates and colors but are not significant after correction. Gray cells with em dashes denote unavailable or ineligible coefficients, not zero associations. Positive correlations increase the raw metric. Model abbreviations and selection rules are defined in Appendix~\ref{sec:dataset-associations}.}
\label{fig:top5_models_classification_Accuracy}
\end{figure}

\begin{figure}[!htbp]\centering
\includegraphics[width=\textwidth,height=.81\textheight,keepaspectratio]{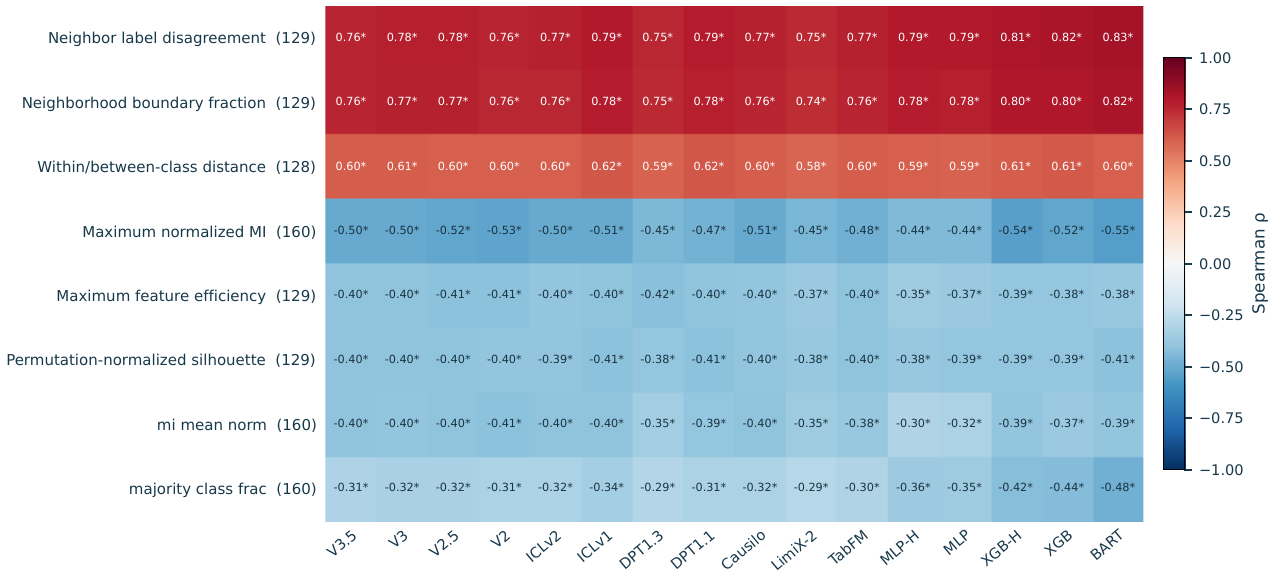}
\caption{Classification: individual-model associations with Brier. Rows are the union of each model’s top five non-auxiliary descriptors; parentheses give complete-pair dataset counts. Stars indicate BH $q<0.05$ over all eligible descriptor--model tests for this task and metric; unstarred cells retain their signed estimates and colors but are not significant after correction. Gray cells with em dashes denote unavailable or ineligible coefficients, not zero associations. Positive correlations increase the raw metric. Model abbreviations and selection rules are defined in Appendix~\ref{sec:dataset-associations}.}
\label{fig:top5_models_classification_Brier}
\end{figure}

\begin{figure}[!htbp]
\centering
\includegraphics[width=\textwidth,height=.81\textheight,keepaspectratio]{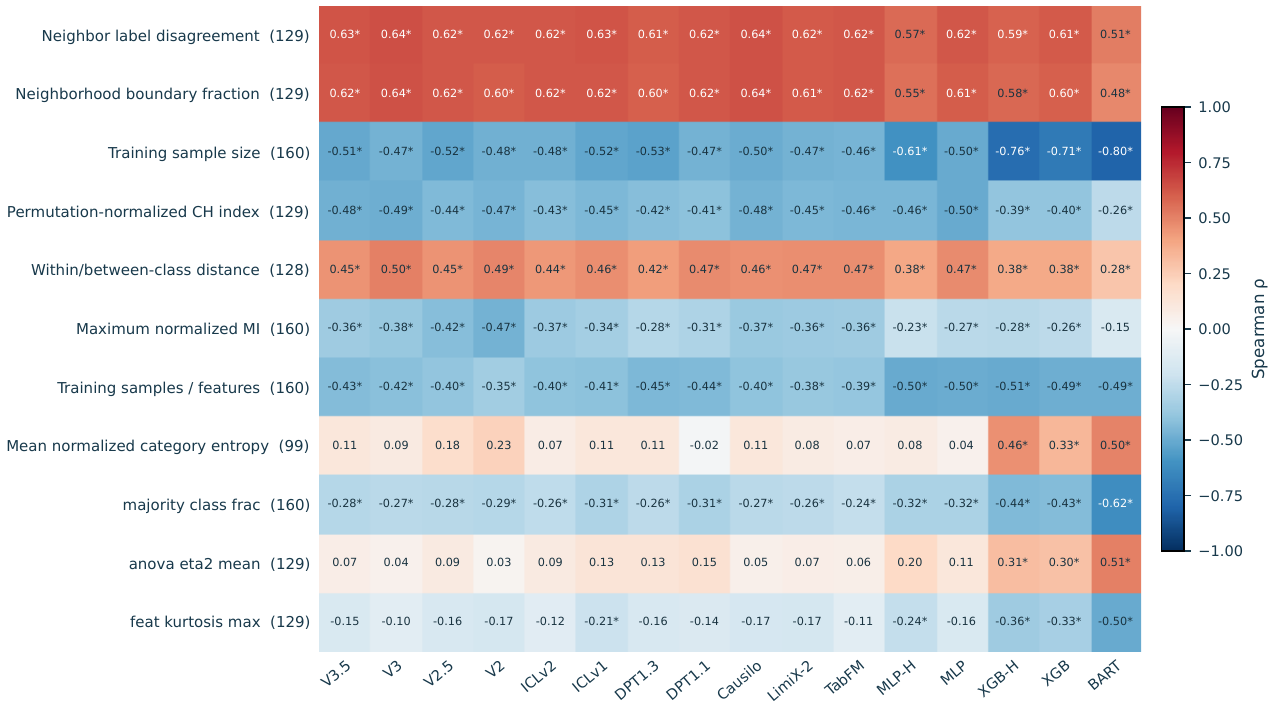}
\caption{Classification: individual-model associations with Confidence ECE. Rows are the union of each model’s top five non-auxiliary descriptors; parentheses give complete-pair dataset counts. Stars indicate BH $q<0.05$ over all eligible descriptor--model tests for this task and metric; unstarred cells retain their signed estimates and colors but are not significant after correction. Gray cells with em dashes denote unavailable or ineligible coefficients, not zero associations. Positive correlations increase the raw metric. Model abbreviations and selection rules are defined in Appendix~\ref{sec:dataset-associations}.}
\label{fig:top5_models_classification_Confidence_ECE}
\end{figure}

\begin{figure}[!htbp]\centering
\includegraphics[width=\textwidth,height=.81\textheight,keepaspectratio]{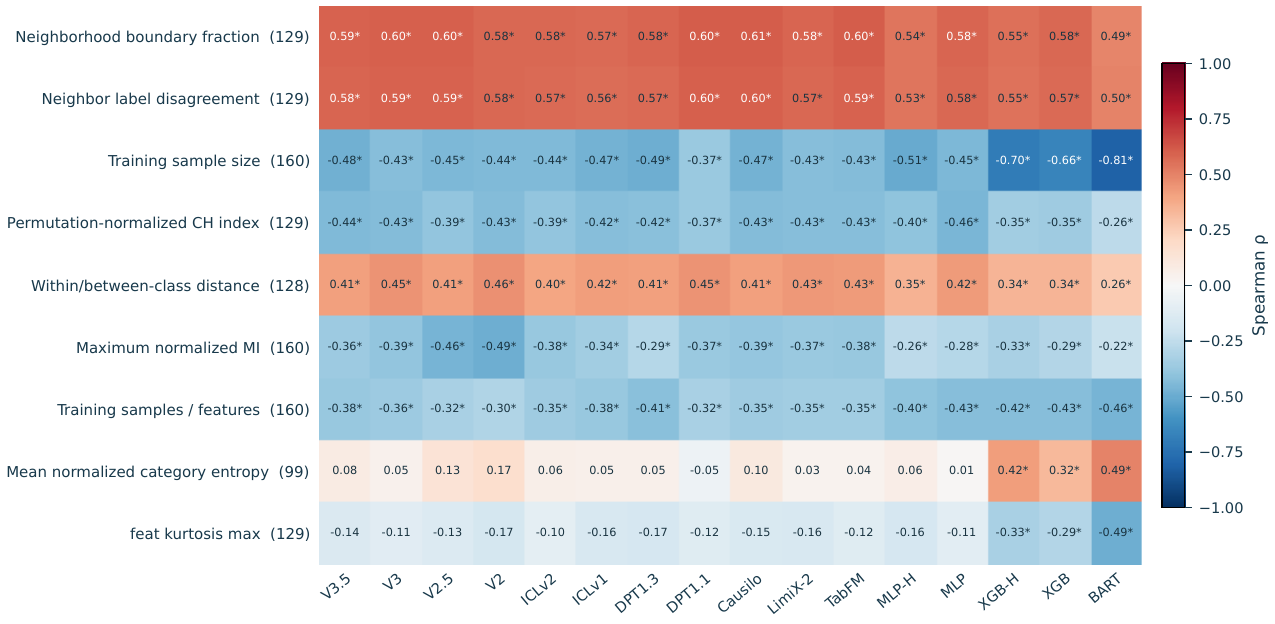}
\caption{Classification: individual-model associations with Classwise ECE. Rows are the union of each model’s top five non-auxiliary descriptors; parentheses give complete-pair dataset counts. Stars indicate BH $q<0.05$ over all eligible descriptor--model tests for this task and metric; unstarred cells retain their signed estimates and colors but are not significant after correction. Gray cells with em dashes denote unavailable or ineligible coefficients, not zero associations. Positive correlations increase the raw metric. Model abbreviations and selection rules are defined in Appendix~\ref{sec:dataset-associations}.}
\label{fig:top5_models_classification_Classwise_ECE}
\end{figure}

\begin{figure}[!htbp]\centering
\includegraphics[width=\textwidth,height=.81\textheight,keepaspectratio]{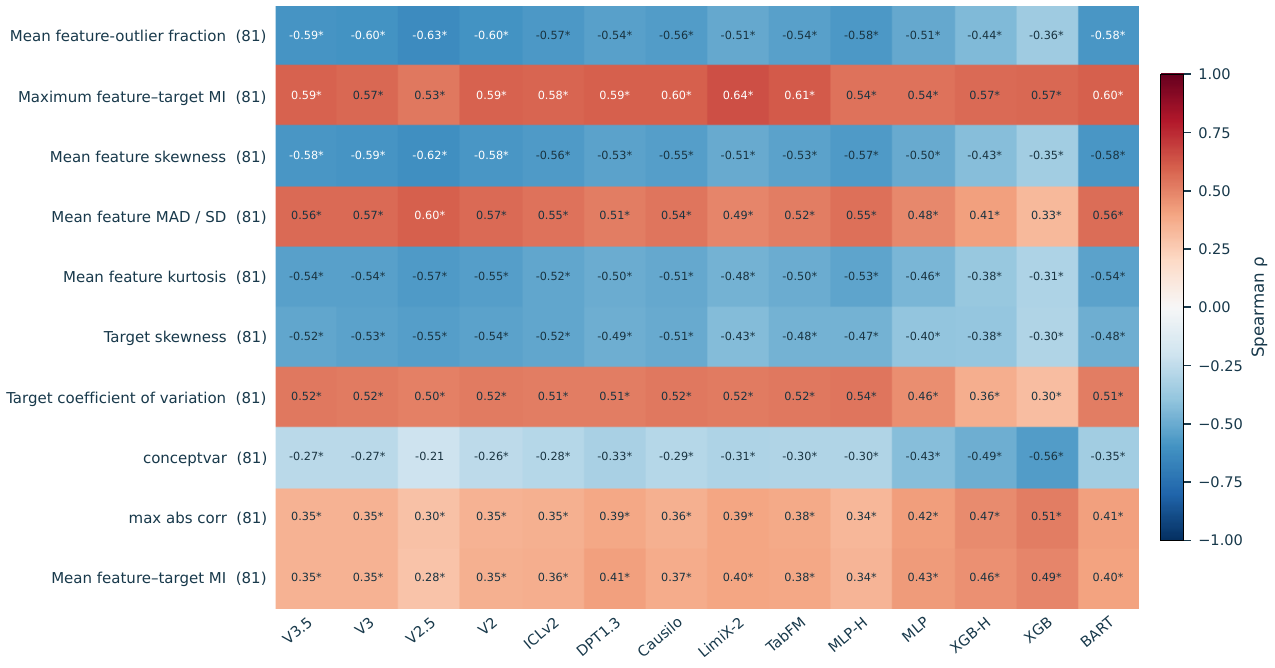}
\caption{Regression: individual-model associations with $R^2$. Rows are the union of each model’s top five non-auxiliary descriptors; parentheses give complete-pair dataset counts. Stars indicate BH $q<0.05$ over all eligible descriptor--model tests for this task and metric; unstarred cells retain their signed estimates and colors but are not significant after correction. Gray cells with em dashes denote unavailable or ineligible coefficients, not zero associations. Positive correlations increase the raw metric. Model abbreviations and selection rules are defined in Appendix~\ref{sec:dataset-associations}.}
\label{fig:top5_models_regression_R2}
\end{figure}

\begin{figure}[!htbp]\centering
\includegraphics[width=\textwidth,height=.81\textheight,keepaspectratio]{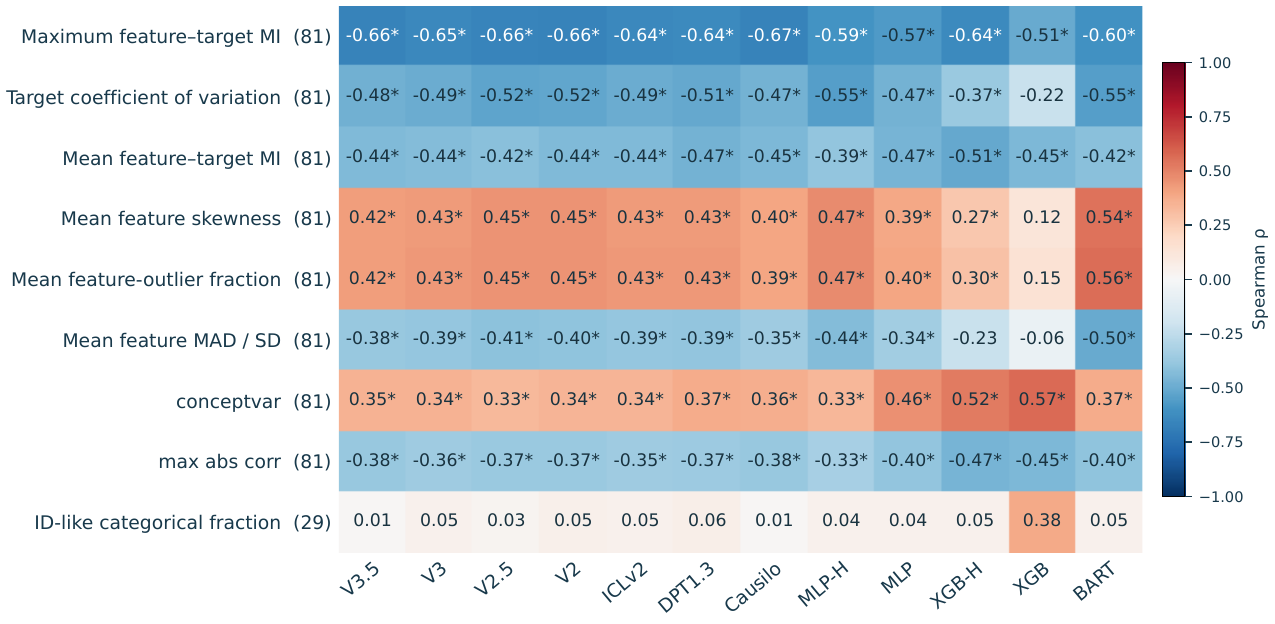}
\caption{Regression: individual-model associations with Normalized CRPS. Rows are the union of each model’s top five non-auxiliary descriptors; parentheses give complete-pair dataset counts. Stars indicate BH $q<0.05$ over all eligible descriptor--model tests for this task and metric; unstarred cells retain their signed estimates and colors but are not significant after correction. Gray cells with em dashes denote unavailable or ineligible coefficients, not zero associations. Positive correlations increase the raw metric. Model abbreviations and selection rules are defined in Appendix~\ref{sec:dataset-associations}.}
\label{fig:top5_models_regression_Normalized_CRPS}
\end{figure}

\begin{figure}[!htbp]\centering
\includegraphics[width=\textwidth,height=.81\textheight,keepaspectratio]{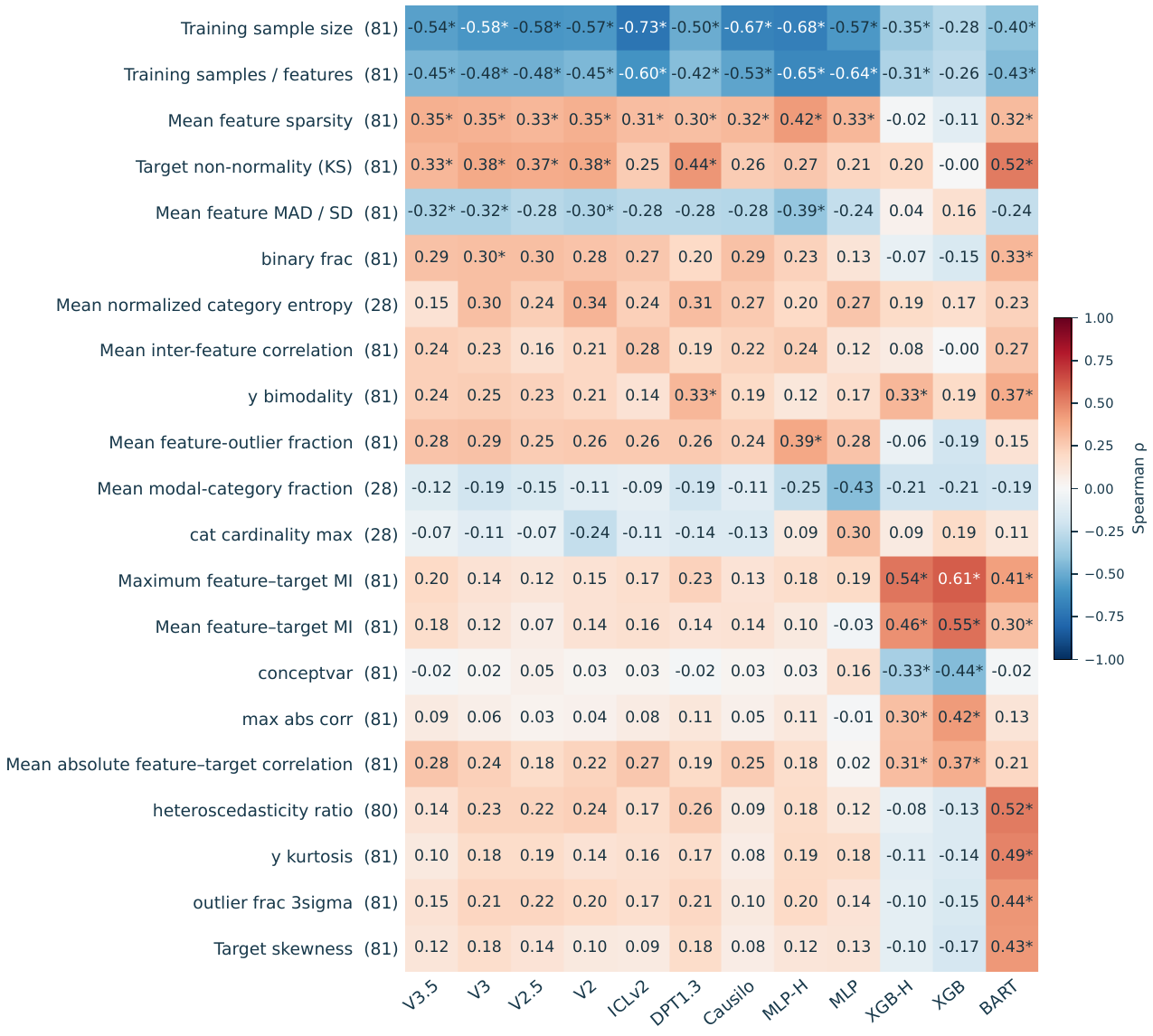}
\caption{Regression: individual-model associations with PIT--KS. Rows are the union of each model’s top five non-auxiliary descriptors; parentheses give complete-pair dataset counts. Stars indicate BH $q<0.05$ over all eligible descriptor--model tests for this task and metric; unstarred cells retain their signed estimates and colors but are not significant after correction. Gray cells with em dashes denote unavailable or ineligible coefficients, not zero associations. Positive correlations increase the raw metric. Model abbreviations and selection rules are defined in Appendix~\ref{sec:dataset-associations}.}
\label{fig:top5_models_regression_PIT-KS}
\end{figure}

\begin{figure}[!htbp]\centering
\includegraphics[width=\textwidth,height=.81\textheight,keepaspectratio]{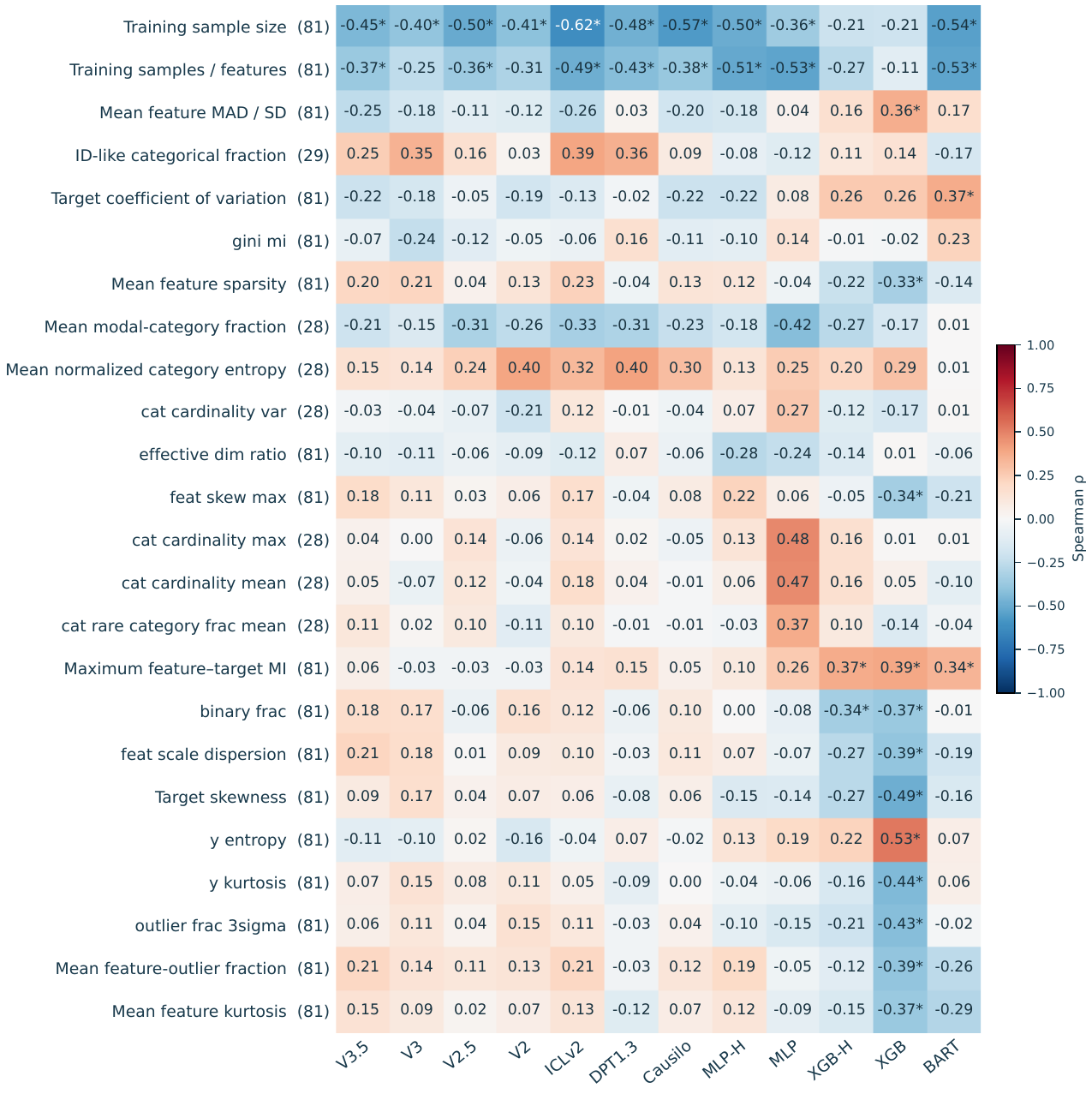}
\caption{Regression: individual-model associations with Coverage error. Rows are the union of each model’s top five non-auxiliary descriptors; parentheses give complete-pair dataset counts. Stars indicate BH $q<0.05$ over all eligible descriptor--model tests for this task and metric; unstarred cells retain their signed estimates and colors but are not significant after correction. Gray cells with em dashes denote unavailable or ineligible coefficients, not zero associations. Positive correlations increase the raw metric. Model abbreviations and selection rules are defined in Appendix~\ref{sec:dataset-associations}.}
\label{fig:top5_models_regression_Coverage_error}
\end{figure}

\begin{figure}[!htbp]\centering
\includegraphics[width=\textwidth,height=.81\textheight,keepaspectratio]{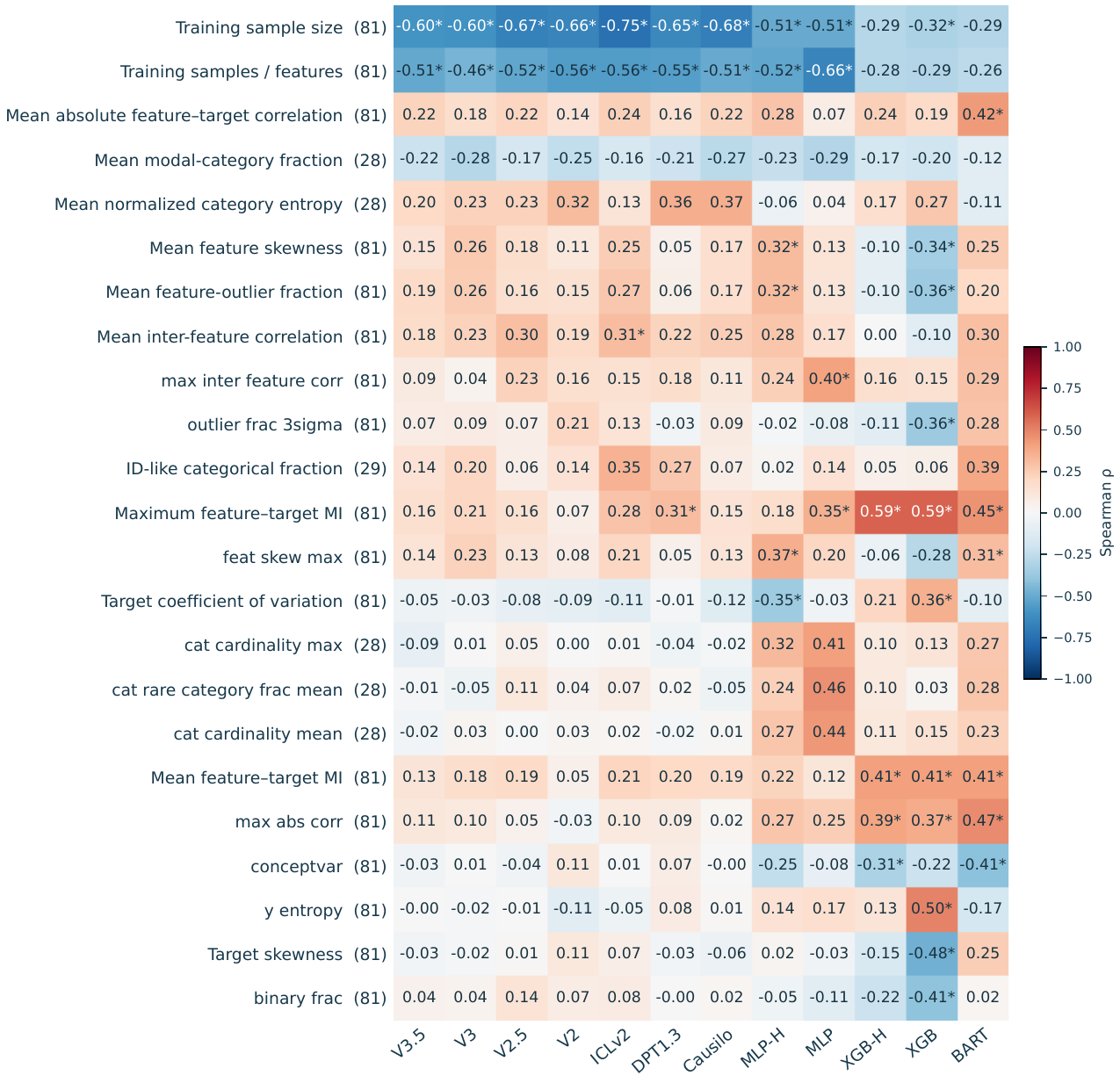}
\caption{Regression: individual-model associations with WSC error. Rows are the union of each model’s top five non-auxiliary descriptors; parentheses give complete-pair dataset counts. Stars indicate BH $q<0.05$ over all eligible descriptor--model tests for this task and metric; unstarred cells retain their signed estimates and colors but are not significant after correction. Gray cells with em dashes denote unavailable or ineligible coefficients, not zero associations. Positive correlations increase the raw metric. Model abbreviations and selection rules are defined in Appendix~\ref{sec:dataset-associations}.}
\label{fig:top5_models_regression_WSC_error}
\end{figure}

\begin{figure}[!htbp]\centering
\includegraphics[width=\textwidth,height=.81\textheight,keepaspectratio]{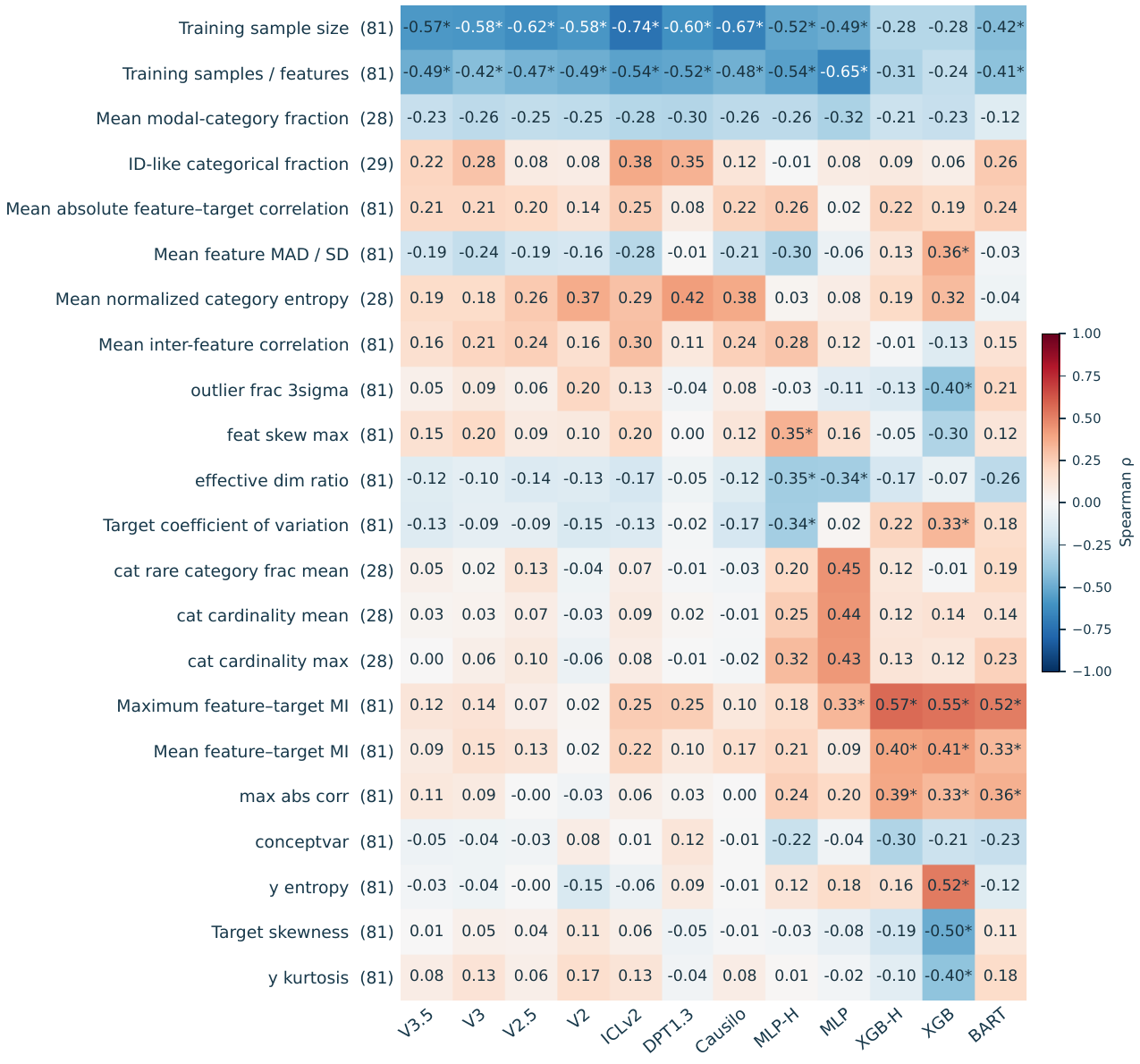}
\caption{Regression: individual-model associations with Total error. Rows are the union of each model’s top five non-auxiliary descriptors; parentheses give complete-pair dataset counts. Stars indicate BH $q<0.05$ over all eligible descriptor--model tests for this task and metric; unstarred cells retain their signed estimates and colors but are not significant after correction. Gray cells with em dashes denote unavailable or ineligible coefficients, not zero associations. Positive correlations increase the raw metric. Model abbreviations and selection rules are defined in Appendix~\ref{sec:dataset-associations}.}
\label{fig:top5_models_regression_Total_error}
\end{figure}

Figures~\ref{fig:top5_models_classification_Accuracy}--\ref{fig:top5_models_classification_Classwise_ECE}
report the classification results, and
Figures~\ref{fig:top5_models_regression_R2}--\ref{fig:top5_models_regression_Total_error}
report the regression results.

Across classification models, class-overlap meta-features have consistent signs and the strongest associations with accuracy, Brier score, and both ECE variants; training size is more specifically associated with calibration.
Across regression models, feature--target mutual information and feature outliers or skewness consistently track predictive performance, whereas training size and samples per feature are the most stable correlates of PIT--KS and coverage errors.
Association magnitudes still vary across models, especially for the regression calibration metrics, so the pooled patterns do not imply identical model behavior.

\section{Conformal Methods and Additional Results}
\label{sec:support-conformal}
This appendix documents the experimental protocol and complete set of post-processing methods underlying Section~\ref{sec:conformal}.
It then reports the full results across nominal coverage levels, separating coverage reliability from interval efficiency and degenerate outputs.

\subsection{Experimental Protocol}
The conformal experiment uses the seven distributional TFMs and BART, with
base-model training and predictions identical to the vanilla benchmark. From
an original test set of size $n$, a shared random permutation assigns
$\lfloor n/4\rfloor$ observations to an auxiliary block $A$, the next
$\lfloor n/4\rfloor$ to a calibration block $C$, and the remainder to a
held-out block $T$. The allocation is shared across base models, methods, and
nominal levels, and base predictions remain frozen throughout post-processing.
Each method receives the same labeled development pool $P=A\cup C$, although
the number of labels used for final calibration and the computational cost
need not be identical. Here $A$ is the auxiliary fitting block, $C$ is the
final calibration block, and $T$ is the held-out evaluation block; $P$ is the
development pool available before evaluation. The method-specific allocations
are described below. We fit each post-processor separately at
$\alpha\in\{0.05,0.10,0.15,0.20\}$.

Within this common allocation, SCP-ABS, SCP-NORM, SCP-CQR, and SCP-PIT
perform global split-conformal calibration on the pooled development set $P$;
SCP-CQR uses the conformalized quantile-regression construction of
\citet{romano2019cqr}. CalLCP and RLCP likewise use $P$ for localized
calibration, following \citet{guan2021localized} and \citet{hore2023local},
respectively. SLCP fits its local score model on $A$ and calibrates the
resulting score correction on $C$ \citep{han2022slcp}. PCP and RCP also use
$A$ for auxiliary fitting and $C$ for final calibration, following
\citet{zhang2024posterior} and \citet{plassier2025rectifying}, respectively.

\paragraph{Evaluation and aggregation.}
We divide $T$ into search and evaluation halves using the WSC procedure in Appendix~\ref{sec:metric-definitions}.
All methods share the evaluation observations, while each method selects its own worst slab on the search half.
The comparison retains dataset--repetition pairs for which every method--model combination is present and WSC is defined at every nominal level.
We first average retained repetitions within each dataset and then weight datasets equally.
Uncertainty intervals use paired bootstrap resampling of whole datasets.
Width summaries additionally exclude evaluation blocks with response standard deviation at most $10^{-10}$; this exclusion is applied to all configurations on the same dataset--repetition pair.
The main-text Figure~\ref{fig:conformal-intervals} compares Vanilla and SCP-CQR at 90\% nominal coverage on this positive-variance subset.

\paragraph{Output conventions.}
Infinite intervals are retained and therefore imply infinite arithmetic mean
width. For signed-score methods, adjusted lower and upper endpoints can cross;
we treat the resulting prediction set as empty, with zero coverage and zero
width. Interval and endpoint-pinball losses are undefined for empty sets and
are not replaced by finite surrogates. Infinite-interval and empty-set rates
are computed on the evaluation half of $T$.

\subsection{Post-processing Methods}
\subsubsection{Global Split Conformal Methods}
Let $\mu(x)$ be the stored point prediction, $Q_x$ the monotone quantile-grid
interpolant, and $F_x$ its CDF interpolant. We use four split-conformal scores,
including the signed quantile score of \citet{romano2019cqr}:
\begin{align*}
 s_{\mathrm{ABS}}(x,y)&=|y-\mu(x)|, &
 s_{\mathrm{NORM}}(x,y)&=\frac{|y-\mu(x)|}{\max\{\sigma(x),10^{-8}\}},\\
 s_{\mathrm{CQR}}(x,y)&=\max\{Q_x(\alpha/2)-y,\ y-Q_x(1-\alpha/2)\}, &
 s_{\mathrm{PIT}}(x,y)&=\max\{F_x(y),1-F_x(y)\}.
\end{align*}
For NORM, $\sigma^2(x)$ is the quantile-grid second moment minus its squared
first moment, floored at $10^{-16}$. Both moments use trapezoidal integration
and constant quantile extension to 0 and 1, so every base model uses the same
construction.

All four variants use the same finite-sample calibration quantile. With
$n_{\mathrm{cal}}$ calibration scores, the threshold $t$ is their $k$th order
statistic, where
$k=\lceil(n_{\mathrm{cal}}+1)(1-\alpha)\rceil$; we set $t=+\infty$ when
$k>n_{\mathrm{cal}}$. ABS returns $[\mu-t,\mu+t]$, NORM returns
$[\mu-t\sigma,\mu+t\sigma]$, and CQR returns
\[
[Q_x(\alpha/2)-t,\ Q_x(1-\alpha/2)+t].
\]
The CQR threshold remains signed, so a negative threshold shrinks the
interval. PIT returns $[Q_x(1-t),Q_x(t)]$ for $t<1$ and the real line for
$t\ge1$. Its inversion uses interpolation on 100 equally spaced probabilities
from 0.001 to 0.999; CDF evaluation uses the original 199-point grid and the
repeated-quantile convention in Appendix~\ref{sec:metric-definitions}.

\subsubsection{Localized Conformal Methods}
\label{sec:local-conformal-settings}
For distance calculations, numeric features are median-imputed and standardized, and categorical features are one-hot encoded with a separate missing category and all-zero blocks for unseen categories. This transformation is fitted on $A$. Write $z(x)$ for the resulting vector and use
\[
K_h(z,z')=\exp\{-\lVert z-z'\rVert_2^2/(2h^2)\}.
\]
For CalLCP and RLCP, bandwidth selection uses only the covariates in $P$. Let $K$ be the bandwidth-selection matrix and $H_{ij}=K_{ij}/\sum_lK_{il}$. The implemented effective-sample-size criterion is
\[
 n_{\mathrm{eff}}(h)=\frac{|P|}{\sum_{i,j}H_{ij}^2}-1.
\]
CalLCP uses $K_{ij}=K_h(z_i,z_j)$, whereas RLCP uses
$K_{ij}=K_h(z_i,z_j+h\epsilon_j)$ with fixed independent
$\epsilon_j\sim\mathcal N(0,I)$ during bandwidth selection. Both choose $h$ by
bisection to target $n_{\mathrm{eff}}=100$; if the target lies outside the
search bracket, we use the corresponding endpoint.

\paragraph{CalLCP.}
We implement the localized conformal procedure of \citet{guan2021localized}.
For a query $x$, write $K_{ij}=K_h(z_i,z_j)$ and $K_{ix}=K_h(z_i,z(x))$ for the calibration observations. Let $a_i=\sum_jK_{ij}\mathbf1\{s_j<s_i\}$. At a candidate score breakpoint $t$, the calibrated local-rank statistics are
\[
T_i(t)=\frac{a_i+K_{ix}\mathbf1\{t\le s_i\}}{\sum_jK_{ij}+K_{ix}},
\qquad
T_x(t)=\frac{\sum_jK_{xj}\mathbf1\{s_j<t\}}{\sum_jK_{xj}+1}.
\]
The calibration expression uses the query-score left limit at a breakpoint. With $n=|P|$ and $k=\min\{n,\lceil(n+1)(1-\alpha)\rceil\}$, we choose the largest breakpoint satisfying $T_x(t)\le T_{(k)}(t)$, where $T_{(k)}$ is the $k$th order statistic of the $n$ calibration statistics. An infinite threshold is returned if this condition also holds beyond all finite scores. Self-weights are one and tied calibration scores use the strict inequality in $a_i$.

\paragraph{RLCP.}
Following \citet{hore2023local}, for each query we sample a localization center $\widetilde z=z(x)+h\epsilon$, $\epsilon\sim\mathcal N(0,I)$. Calibration scores receive weights $K_h(\widetilde z,z_i)$, and an atom at $+\infty$ receives weight $K_h(\widetilde z,z(x))$. After normalization, the weighted $(1-\alpha)$ quantile defines the signed CQR threshold.

\paragraph{SLCP.}
Following \citet{han2022slcp}, set $h$ to the median Euclidean distance over distinct observation pairs in $A$, using $h=1$ if the median is zero. The local estimate $\widehat q(x)$ is the weighted $(1-\alpha)$ quantile of auxiliary signed CQR scores, with weights proportional to $K_h(z(x),z_i)$ for $i\in A$. Calibrate $s_i-\widehat q(x_i)$ on $C$ using the split-conformal order statistic above, obtaining $u$. The final CQR threshold is $\widehat q(x)+u$.

\subsubsection{Learned Conditional Adaptation}
\paragraph{RCP.}
Following \citet{plassier2025rectifying}, RCP fits a distributional
regressor to covariates and signed CQR scores. Its estimated $(1-\alpha)$ score
quantile is $\widehat\tau(x)$. On a disjoint calibration block we compute
$r_i=s_i-\widehat\tau(x_i)$ and obtain its split-conformal threshold $u$. The
resulting CQR threshold is $\widehat\tau(x)+u$; no division or nonnegative
clipping is applied.

The score regressor uses the same model
family, checkpoint, preprocessing, and
eight-member inference configuration as the corresponding base TFM, with the
scores replacing the response. The BART score regressor uses the configuration
in Appendix~\ref{sec:baseline-parameters}. RCP uses $A$ as
the score-model context and evaluates the learned score quantile on $C$ and
$T$; conformal correction is calibrated on $C$. Here
$A$ is separate post-hoc fitting data, not the base model's training set.
The original response predictor remains frozen.

\paragraph{PCP.}
PCP \citep{zhang2024posterior} is applied to absolute residuals from the frozen predictor, using $A$ for
auxiliary fitting and $C$ for calibration. We retain this score because the
published regression theory and the authors' official implementation are
formulated for the absolute-residual PCP construction. Replacing it with a
CQR score would define a separate PCP--CQR variant, requiring its own score
model and prediction-set inversion. We therefore reuse the authors' official
implementation, available at \url{https://github.com/yaozhang24/pcp}, and
retain its prescribed fitting and tuning procedure. Inputs use the
auxiliary-fitted imputed, standardized, and one-hot representation defined
above; the official implementation then applies its own standardization and
intercept. We fit a fresh PCP instance at each nominal level.

\subsection{Coverage Reliability}
\label{sec:conformal-levels}
\begin{figure}[H]
\centering
\includegraphics[width=\textwidth]{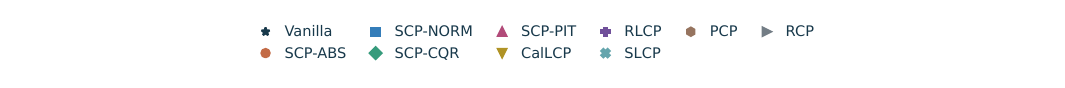}
\par\vspace{.4ex}
\includegraphics[width=.24\textwidth]{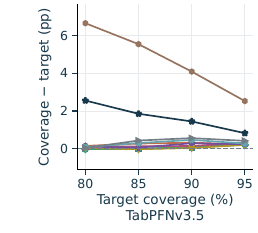}\hfill
\includegraphics[width=.24\textwidth]{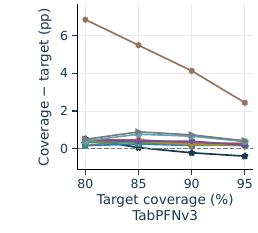}\hfill
\includegraphics[width=.24\textwidth]{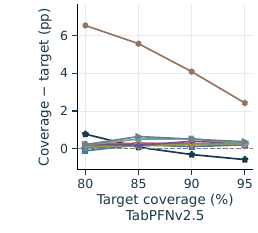}\hfill
\includegraphics[width=.24\textwidth]{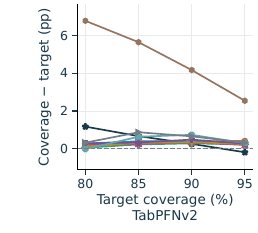}
\par\vspace{1ex}
\includegraphics[width=.24\textwidth]{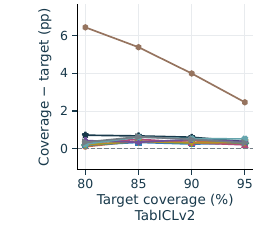}\hfill
\includegraphics[width=.24\textwidth]{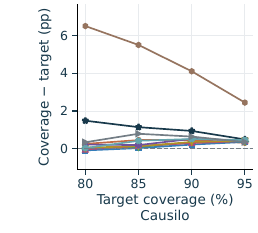}\hfill
\includegraphics[width=.24\textwidth]{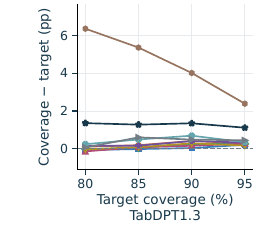}\hfill
\includegraphics[width=.24\textwidth]{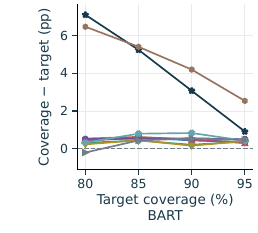}
\caption{
Mean marginal coverage minus nominal coverage, in percentage points.
Zero indicates the target; stars mark vanilla.
Plots in reading order: TabPFNv3.5, TabPFNv3, TabPFNv2.5, TabPFNv2, TabICLv2, Causilo, TabDPT1.3, and BART.
}
\label{fig:conformal-marginal-levels}
\end{figure}

\begin{figure}[H]
\centering
\includegraphics[width=\textwidth]{figures/conformal/methods_legend.pdf}
\par\vspace{.4ex}
\includegraphics[width=.24\textwidth]{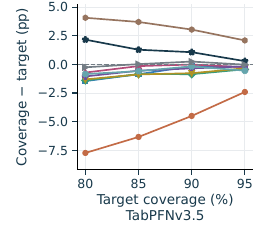}\hfill
\includegraphics[width=.24\textwidth]{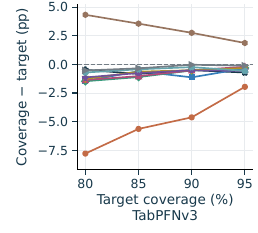}\hfill
\includegraphics[width=.24\textwidth]{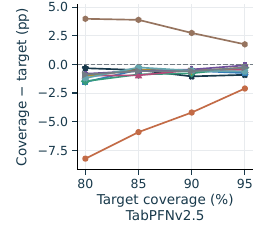}\hfill
\includegraphics[width=.24\textwidth]{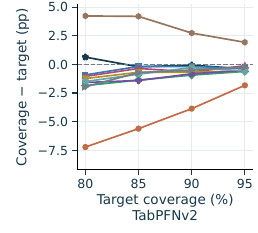}
\par\vspace{1ex}
\includegraphics[width=.24\textwidth]{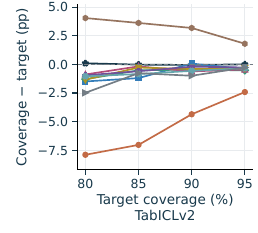}\hfill
\includegraphics[width=.24\textwidth]{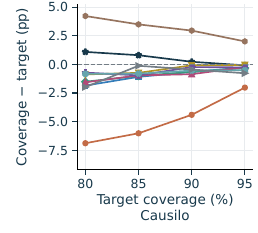}\hfill
\includegraphics[width=.24\textwidth]{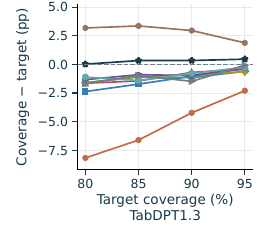}\hfill
\includegraphics[width=.24\textwidth]{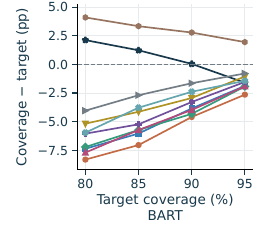}
\caption{
Mean held-out worst-slab coverage minus nominal coverage.
Each method selects its subgroup on the search half and evaluates it on the other half.
Plots in reading order: TabPFNv3.5, TabPFNv3, TabPFNv2.5, TabPFNv2, TabICLv2, Causilo, TabDPT1.3, and BART.
}
\label{fig:conformal-subgroup-levels}
\end{figure}

\paragraph{Coverage across nominal levels.}
SCP-CQR tracks marginal coverage targets across base models (Figure~\ref{fig:conformal-marginal-levels}). The corresponding worst-slab curves remain more heterogeneous (Figure~\ref{fig:conformal-subgroup-levels}). BART's subgroup undercoverage after marginal correction is particularly visible. PCP's conservative coverage accompanies a large rate of infinite intervals, quantified below.

\begin{figure}[H]
\centering
\includegraphics[width=\textwidth]{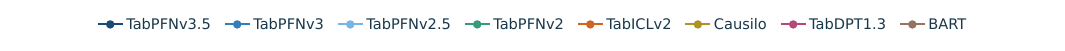}
\par\vspace{.4ex}
\includegraphics[width=.48\textwidth]{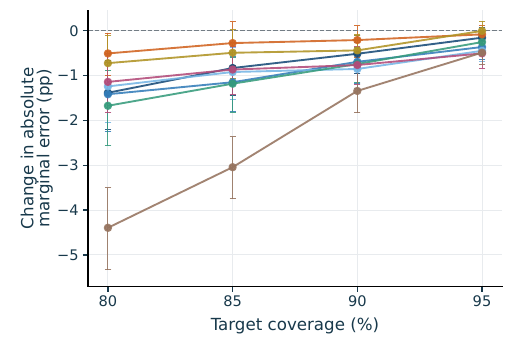}\hfill
\includegraphics[width=.48\textwidth]{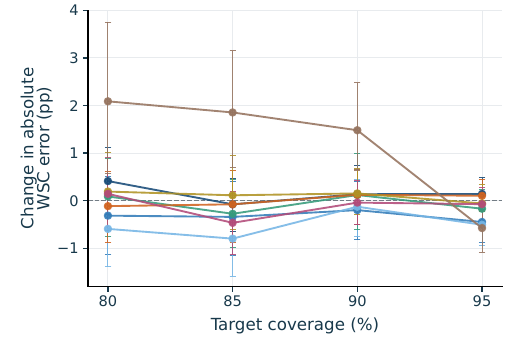}
\caption{
Changes in absolute marginal (left) and worst-slab (right) coverage error after SCP-CQR calibration, relative to the vanilla intervals, in percentage points.
Each curve is one base model, averaged with equal dataset weights; negative values favor calibration.
Error bars are paired 95\% dataset-bootstrap intervals.
}
\label{fig:conformal-cqr}
\end{figure}

\paragraph{Paired improvements over Vanilla.}
For dataset $d$, base model $m$, target level $1-\alpha$, and coverage
metric $q\in\{\mathrm{MC},\mathrm{WSC}\}$, define the trial-level absolute
error of method $k$ as
$e_{d,s}^{(q)}(k)=\lvert C_{d,s}^{(q)}(k)-(1-\alpha)\rvert$.
We compare each post-processor $k$ with Vanilla on the same retained
dataset--trial pairs, using the within-dataset contrast
\[
\Delta_{d}^{(q)}(k)=\overline{e_{d,s}^{(q)}(k)}_s-
\overline{e_{d,s}^{(q)}(\mathrm{Vanilla})}_s.
\]
The reported change is the equal-dataset mean of $\Delta_d^{(q)}(k)$;
uncertainty intervals use paired bootstrap resampling of datasets.
Thus, negative values indicate a reduction in absolute coverage error relative to Vanilla.
Figure~\ref{fig:conformal-cqr} isolates the SCP-CQR comparison across nominal levels, while Figures~\ref{fig:errors-90}--\ref{fig:errors-80} summarize every post-processor--base-model pair, together with the corresponding change in interval width.
SCP-CQR reduces marginal error throughout, most for BART at lower targets, but increases BART's WSC error at 80--90\% coverage.
For width, we divide each dataset's seed-averaged normalized width by that of Vanilla and report the median ratio across datasets minus one, in percent; the median rather than the mean keeps the summary defined when a minority of datasets has infinite mean width.
Marginal-error reductions are more consistent than subgroup-error reductions, and both vary with the target level.
Across the 32 base-model--level combinations, the four global split-conformal variants, CalLCP, and RLCP reduce marginal error with intervals excluding zero in 24--27 cases, whereas SLCP and RCP do so in 12 and PCP increases marginal error in 29.
No post-processor reduces WSC error with an interval excluding zero in more than 10 of the 32 combinations, and SCP-ABS increases it in 30.
\newcommand{\overviewfigure}[3]{%
\begin{figure}[H]
\centering
\includegraphics[width=.32\textwidth]{figures/conformal/overview_#1_marginal_error.pdf}\hfill
\includegraphics[width=.32\textwidth]{figures/conformal/overview_#1_wsc_error.pdf}\hfill
\includegraphics[width=.32\textwidth]{figures/conformal/overview_#1_width.pdf}
\caption{All post-processors and base models at #2\% nominal coverage.
Left and center: paired changes in marginal and worst-slab coverage error relative to Vanilla, in percentage points; blue cells reduce error and red cells increase it.
Bold values have paired 95\% dataset-bootstrap intervals excluding zero.
Right: median across datasets of the percentage change in normalized width relative to Vanilla; gray $\infty$ cells have infinite median width.
Rows are grouped into global split-conformal, localized, and learned-adjustment methods; colors share a common scale across nominal levels.}
\label{#3}
\end{figure}}

\overviewfigure{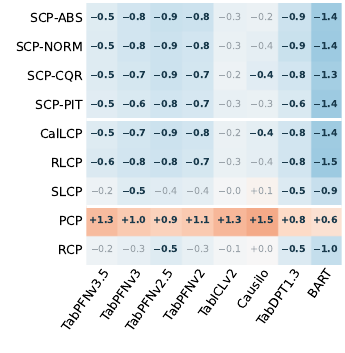}{90}{fig:errors-90}
\overviewfigure{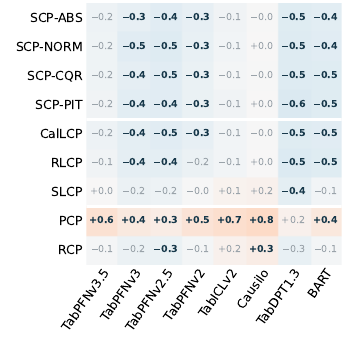}{95}{fig:errors-95}
\overviewfigure{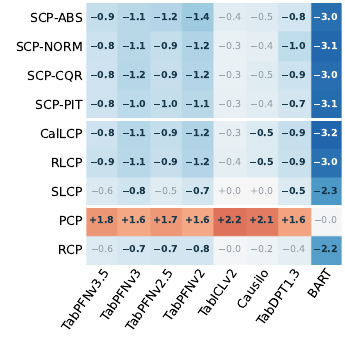}{85}{fig:errors-85}
\overviewfigure{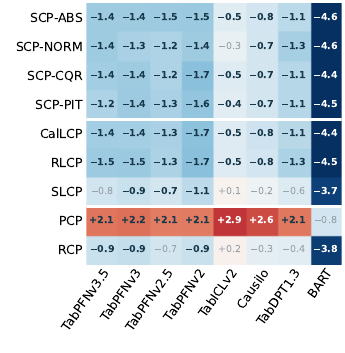}{80}{fig:errors-80}
\subsection{Efficiency and Degenerate Outputs}
\paragraph{Coverage--width profiles.}
Figures~\ref{fig:conformal-width-80}--\ref{fig:conformal-width-95} show the full method comparison at 80\%, 85\%, 90\%, and 95\% target coverage on evaluation trials with positive response standard deviation. Finite widths are comparable on each common scale; methods producing infinite intervals are shown separately. In every figure, the first two rows are marginal coverage and the last two rows are WSC, with a divider between them. Across these profiles, BART generally benefits more from conformal post-processing than the TFMs, especially at lower target coverage.

\newcommand{\tradeofffigure}[3]{%
\begin{figure}[H]
\centering
\includegraphics[width=\textwidth]{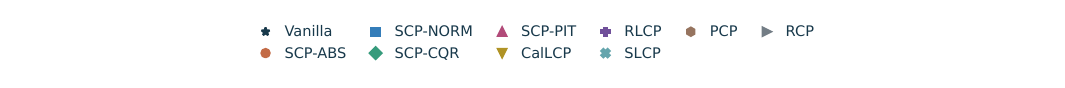}
\par\vspace{.4ex}
\includegraphics[width=.24\textwidth]{figures/conformal/tradeoff_#1_g0_marginal_coverage_tabpfnv3.5.pdf}\hfill
\includegraphics[width=.24\textwidth]{figures/conformal/tradeoff_#1_g0_marginal_coverage_tabpfnv3.pdf}\hfill
\includegraphics[width=.24\textwidth]{figures/conformal/tradeoff_#1_g0_marginal_coverage_tabpfnv2.5.pdf}\hfill
\includegraphics[width=.24\textwidth]{figures/conformal/tradeoff_#1_g0_marginal_coverage_tabpfnv2.pdf}
\par\vspace{1ex}
\includegraphics[width=.24\textwidth]{figures/conformal/tradeoff_#1_g1_marginal_coverage_tabiclv2.pdf}\hfill
\includegraphics[width=.24\textwidth]{figures/conformal/tradeoff_#1_g1_marginal_coverage_causilo.pdf}\hfill
\includegraphics[width=.24\textwidth]{figures/conformal/tradeoff_#1_g1_marginal_coverage_tabdpt1.3.pdf}\hfill
\includegraphics[width=.24\textwidth]{figures/conformal/tradeoff_#1_g1_marginal_coverage_bart.pdf}
\par\vspace{.6ex}\noindent{\color{black}\hrule height .6pt}\par\vspace{.6ex}
\includegraphics[width=.24\textwidth]{figures/conformal/tradeoff_#1_g0_worst_slab_coverage_tabpfnv3.5.pdf}\hfill
\includegraphics[width=.24\textwidth]{figures/conformal/tradeoff_#1_g0_worst_slab_coverage_tabpfnv3.pdf}\hfill
\includegraphics[width=.24\textwidth]{figures/conformal/tradeoff_#1_g0_worst_slab_coverage_tabpfnv2.5.pdf}\hfill
\includegraphics[width=.24\textwidth]{figures/conformal/tradeoff_#1_g0_worst_slab_coverage_tabpfnv2.pdf}
\par\vspace{1ex}
\includegraphics[width=.24\textwidth]{figures/conformal/tradeoff_#1_g1_worst_slab_coverage_tabiclv2.pdf}\hfill
\includegraphics[width=.24\textwidth]{figures/conformal/tradeoff_#1_g1_worst_slab_coverage_causilo.pdf}\hfill
\includegraphics[width=.24\textwidth]{figures/conformal/tradeoff_#1_g1_worst_slab_coverage_tabdpt1.3.pdf}\hfill
\includegraphics[width=.24\textwidth]{figures/conformal/tradeoff_#1_g1_worst_slab_coverage_bart.pdf}
\caption{Coverage and normalized width at #2\% for all eight base models. Stars mark Vanilla and arrows show post-processing changes. The \colorblock{residualgray} strip denotes infinite mean width. The upper two rows show marginal coverage; the lower two rows show held-out worst-slab coverage (WSC). Among methods reaching the target (on or to the right of the dashed line), lower normalized width is preferable.}
\label{#3}
\end{figure}}

\tradeofffigure{0.20}{80}{fig:conformal-width-80}
\tradeofffigure{0.15}{85}{fig:conformal-width-85}
\tradeofffigure{0.10}{90}{fig:conformal-width-all}
\tradeofffigure{0.05}{95}{fig:conformal-width-95}

\begin{figure}[H]
\centering
\includegraphics[width=.48\textwidth]{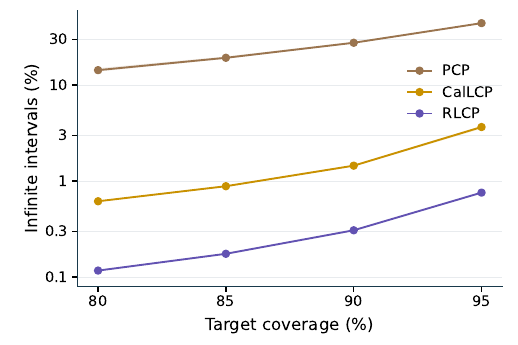}
\caption{
Infinite-interval rates across target coverage levels.
Lines average the eight base regressors, bands span their model-specific means, and the vertical axis is logarithmic.
Methods with zero infinite-interval rates are omitted.
}
\label{fig:conformal-infinite-main}
\end{figure}

\begin{figure}[H]
\centering
\includegraphics[width=\textwidth]{figures/conformal/methods_legend.pdf}
\par\vspace{.4ex}
\includegraphics[width=.24\textwidth]{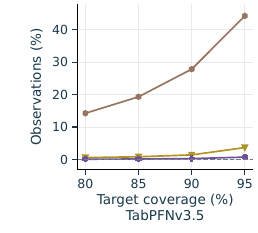}\hfill
\includegraphics[width=.24\textwidth]{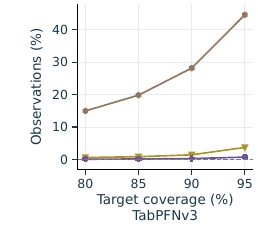}\hfill
\includegraphics[width=.24\textwidth]{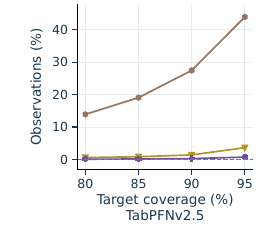}\hfill
\includegraphics[width=.24\textwidth]{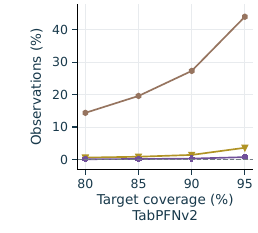}
\par\vspace{1ex}
\includegraphics[width=.24\textwidth]{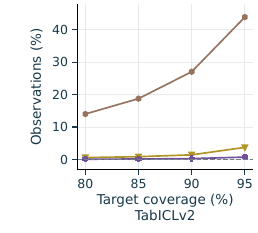}\hfill
\includegraphics[width=.24\textwidth]{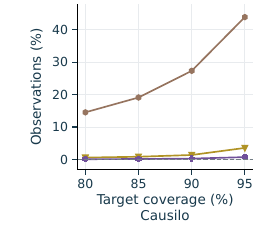}\hfill
\includegraphics[width=.24\textwidth]{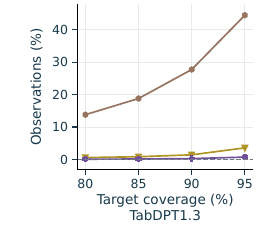}\hfill
\includegraphics[width=.24\textwidth]{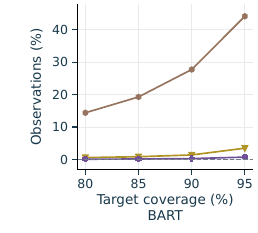}
\caption{
Infinite-interval rates across nominal levels, averaged within datasets and then across datasets.
Methods with zero rates throughout are omitted.
Plots in reading order: TabPFNv3.5, TabPFNv3, TabPFNv2.5, TabPFNv2, TabICLv2, Causilo, TabDPT1.3, and BART.
}
\label{fig:infinite-sets}
\end{figure}

\begin{figure}[H]
\centering
\includegraphics[width=\textwidth]{figures/conformal/methods_legend.pdf}
\par\vspace{.4ex}
\includegraphics[width=.24\textwidth]{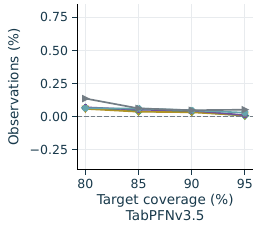}\hfill
\includegraphics[width=.24\textwidth]{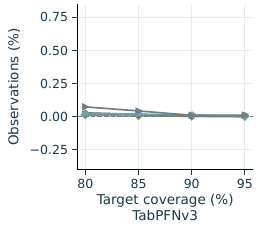}\hfill
\includegraphics[width=.24\textwidth]{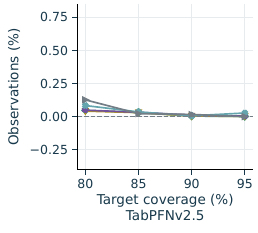}\hfill
\includegraphics[width=.24\textwidth]{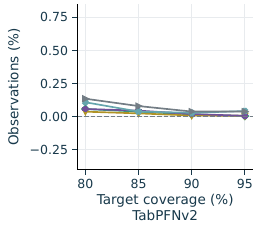}
\par\vspace{1ex}
\includegraphics[width=.24\textwidth]{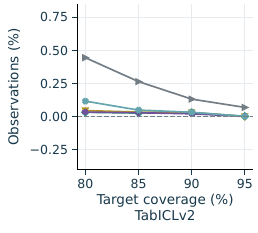}\hfill
\includegraphics[width=.24\textwidth]{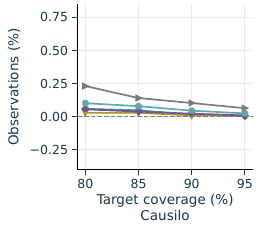}\hfill
\includegraphics[width=.24\textwidth]{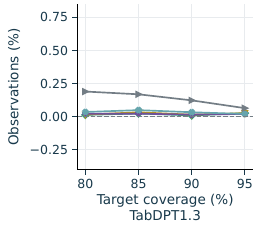}\hfill
\includegraphics[width=.24\textwidth]{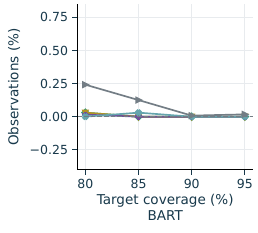}
\caption{
Empty-set rates for signed-score methods.
Empty sets are retained as uncovered observations with zero width.
Plots in reading order: TabPFNv3.5, TabPFNv3, TabPFNv2.5, TabPFNv2, TabICLv2, Causilo, TabDPT1.3, and BART.
}
\label{fig:empty-sets}
\end{figure}

\paragraph{Infinite intervals.}
Figure~\ref{fig:conformal-infinite-main} summarizes infinite-interval rates across base models and target coverage levels.
From 80\% to 95\% target coverage, the base-model-averaged rate rises from 14.3\% to 44.1\% for PCP, from 0.62\% to 3.65\% for CalLCP, and from 0.12\% to 0.76\% for RLCP.
All other methods, including SCP-CQR, remain at zero.
Because any positive rate makes the arithmetic mean width infinite, coverage improvements from these methods must be interpreted together with this loss of informativeness.
An infinite interval occurs when conformal inversion remains feasible as the score threshold tends to $+\infty$; under localized weighting, this can arise when a query has insufficient effective calibration mass for a finite threshold.
Figure~\ref{fig:infinite-sets} reports the corresponding rates separately for each base model.
\paragraph{Empty prediction sets.}
A negative signed-CQR threshold can contract a base quantile interval until
its adjusted lower endpoint exceeds its upper endpoint. Figure~\ref{fig:empty-sets}
reports the frequency of these empty sets. They are retained as uncovered
observations with zero width rather than dropped from the evaluation.

\section{Complete Ensemble-Size Results}
\label{sec:ensemble-appendix}

\subsection{Evaluation Protocol}
\label{sec:ensemble-protocol}

We vary the estimator count over $E\in\{1,4,8,16,32\}$ while holding the dataset splits and all other model settings fixed.
The classification sweep contains all 11 TFMs on 160 datasets and the regression sweep contains all nine regression TFMs on 81 datasets, with five seeds per configuration.
For each metric, we first average the five seeds within each dataset and model--ensemble configuration, then take the median metric across the common set of datasets.
CRPS is normalized at the trial level by the population standard deviation of the training response from the same dataset--seed split before aggregation.
Higher values are better for accuracy and $R^2$; lower values are better for all other reported metrics.

\subsection{Additional Results}
\label{sec:ensemble-additional}

\subsubsection{Median-Metric Results}
\label{sec:ensemble-median}

The median-metric analysis separates predictive metrics from calibration, but
also shows that predictive behavior is metric-dependent. Accuracy and both
proper scores generally improve, with most gains occurring at small to
moderate ensemble sizes. Median $R^2$ is mixed across models. Calibration
metrics remain less monotonic: the preferred ensemble size depends on both
the model and the metric, so larger ensembles do not uniformly improve
calibration.

\begin{figure}[H]
\centering
\includegraphics[width=.70\textwidth]{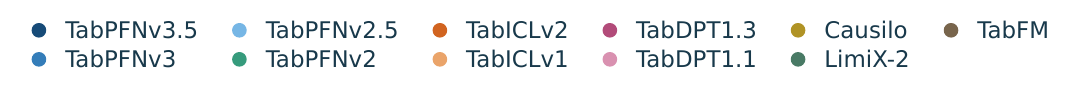}
\par\vspace{.4ex}
\includegraphics[width=.34\textwidth]{figures/ensemble/ensemble_classification_accuracy.pdf}\hspace{.02\textwidth}
\includegraphics[width=.34\textwidth]{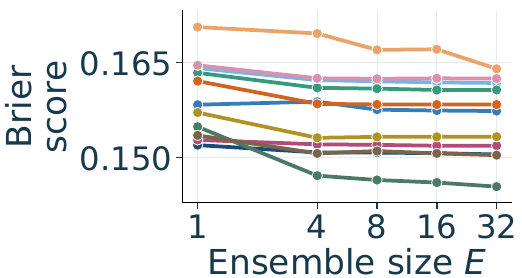}
\par\vspace{1ex}
\includegraphics[width=.34\textwidth]{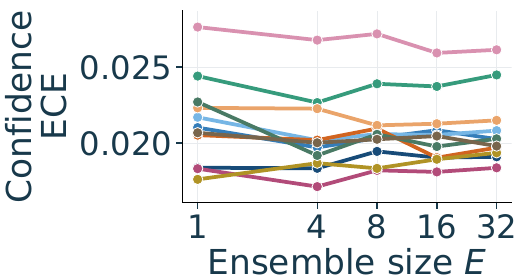}\hspace{.02\textwidth}
\includegraphics[width=.34\textwidth]{figures/ensemble/ensemble_classification_classwise_ece.pdf}
\caption{
Median metrics for the complete classification ensemble-size sweep.
Plots in reading order show accuracy, Brier score, confidence ECE, and classwise ECE.
Each curve is one model, and every point uses the same 160 datasets and five seeds.
Higher accuracy is better; lower Brier score and ECE are better.
}
\label{fig:ensemble-classification-full}
\end{figure}

\begin{figure}[H]
\centering
\includegraphics[width=.70\textwidth]{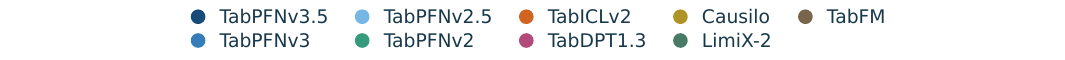}
\par\vspace{.4ex}
\includegraphics[width=.34\textwidth]{figures/ensemble/ensemble_regression_r2.pdf}\hspace{.02\textwidth}
\includegraphics[width=.34\textwidth]{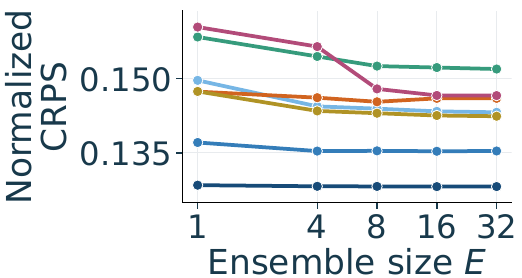}
\par\vspace{1ex}
\includegraphics[width=.34\textwidth]{figures/ensemble/ensemble_regression_pit_ks.pdf}\hspace{.02\textwidth}
\includegraphics[width=.34\textwidth]{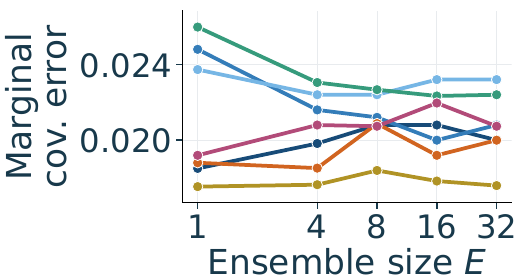}
\par\vspace{1ex}
\includegraphics[width=.34\textwidth]{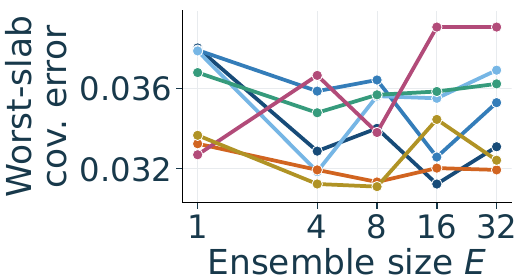}\hspace{.02\textwidth}
\includegraphics[width=.34\textwidth]{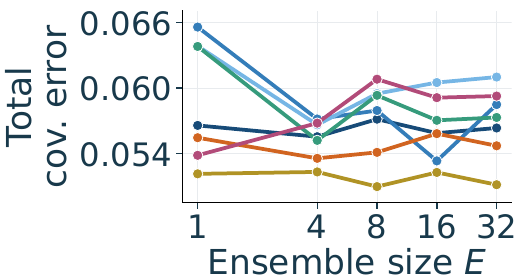}
\caption{
Median metrics for the complete regression ensemble-size sweep.
Plots in reading order show $R^2$, normalized CRPS, PIT--KS, marginal coverage error, worst-slab coverage error, and their sum.
Each curve is one model, and every point uses the same 81 datasets and five seeds.
$R^2$ includes all nine regression TFMs, whereas the distributional plots use the seven models that provide predictive distributions.
Coverage quantities use 90\% nominal intervals.
Higher $R^2$ is better; lower values are better for all other panels.
}
\label{fig:ensemble-regression-full}
\end{figure}

\subsubsection{Coverage--Width Results}
\label{sec:ensemble-absolute}

The median metric plots summarize marginal and subgroup coverage errors, but
do not show the corresponding interval widths. We therefore complement them
with coverage--width summaries for regression intervals. Increasing $E$ changes the coverage--sharpness
operating point: movements toward nominal coverage can be accompanied by
wider intervals, and marginal and worst-slab coverage need not change in the
same way.  Consequently, a better average coverage value alone does not
establish better uncertainty quality or subgroup robustness.

\begin{figure}[H]
\centering
\includegraphics[width=.70\textwidth]{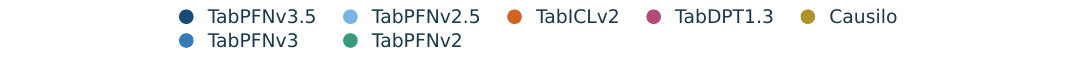}
\par\vspace{.4ex}
\includegraphics[width=.34\textwidth]{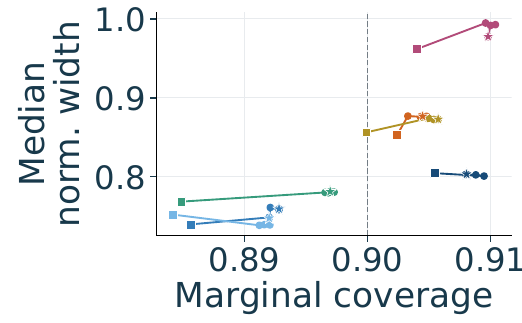}\hspace{.02\textwidth}
\includegraphics[width=.34\textwidth]{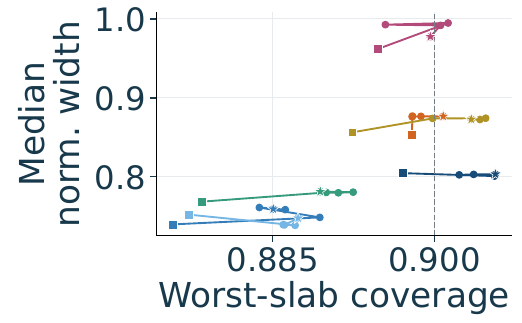}
\caption{
Absolute coverage--width trajectories as ensemble size changes at 90\% nominal coverage.
The left plot uses marginal coverage and the right plot uses worst-slab coverage; the vertical axis is normalized interval width.
Squares mark $E=1$, stars mark $E=32$, and circles mark intermediate ensemble sizes.
Each coordinate is the median across the same 81 datasets after averaging five seeds.
}
\label{fig:ensemble-intervals-full}
\end{figure}

\end{document}